\documentclass[10pt,twocolumn,letterpaper]{article}

\usepackage{cvpr}              

\usepackage[dvipsnames]{xcolor}

\usepackage{adjustbox}
\usepackage{nicematrix}
\usepackage{amssymb}
\usepackage[accsupp]{axessibility}
\definecolor{cvprblue}{rgb}{0.21,0.49,0.74}
\usepackage[pagebackref,breaklinks,colorlinks,citecolor=cvprblue]{hyperref}
\usepackage{multirow}
\usepackage{colortbl}
\usepackage{comment}
\usepackage{tablefootnote} 
\usepackage{xcolor}
\usepackage{supertabular}
\usepackage[dvipsnames]{xcolor}
\usepackage{makecell}
\usepackage{bm}
\usepackage{float}
\usepackage{indentfirst}
\usepackage{setspace}
\usepackage{enumitem}
\usepackage{nicematrix}
\usepackage[linesnumbered,ruled,vlined]{algorithm2e}

\SetCommentSty{mycommfont}

\SetKwInput{KwInput}{Input}                
\SetKwInput{KwOutput}{Output}              

\def\paperID{573} 
\def\confName{CVPR}
\def\confYear{2024}

\newcommand{\rpl}{REAL-Prompt}

\newcommand{\Skip}[1]{}

\newcommand{\reall}{REAL}

\title{The Neglected Tails in Vision-Language Models}

\author{%
  Shubham Parashar\thanks{Co-first authors; $^{\dag}$corresponding author. Project page at \href{https://shubhamprshr27.github.io/neglected-tails-of-vlms/}{link}.} $^{1}$ \quad Zhiqiu Lin$^{*2}$ \quad Tian Liu$^{*1}$ \quad Xiangjue Dong$^1$ 
  \\
   Yanan Li$^{3}$ \quad Deva Ramanan$^{2}$ \quad James Caverlee$^{1}$ \quad Shu Kong$^{14\dag}$ \\ 
  {\small 
  $^1$Texas A\&M University \quad $^2$Carnegie Mellon University \quad $^3$Zhejiang Lab \quad $^4$University of Macau}
  \vspace{-4mm}
}

\begin{document}
\maketitle
\vspace{-3mm}
\begin{abstract}
Vision-language models (VLMs) excel in zero-shot recognition but their performance varies greatly across different visual concepts. 
For example, although CLIP achieves impressive accuracy on ImageNet (60-80\%), its performance drops below 10\% for more than ten concepts like {\tt night} {\tt snake}, presumably due to their limited presence in the pretraining data. 
However, measuring the frequency of concepts in VLMs' large-scale datasets is challenging. We address this by using large language models (LLMs) to count the number of pretraining texts that contain synonyms of these concepts.
Our analysis confirms that popular datasets, such as LAION, exhibit a long-tailed concept distribution, yielding biased performance in VLMs. We also find that downstream applications of VLMs, including visual chatbots (e.g., GPT-4V) and text-to-image models (e.g., Stable Diffusion), often fail to recognize or generate images of rare concepts identified by our method.
To mitigate the imbalanced performance of zero-shot VLMs,
we propose {\bf RE}trieval-{\bf A}ugmented {\bf L}earning (REAL).
First, instead of prompting VLMs using the original class names, REAL uses their most frequent synonyms found in pretraining texts. This simple change already outperforms costly human-engineered and LLM-enriched prompts over nine benchmark datasets. Second, REAL trains a linear classifier on a small yet balanced set of pretraining data retrieved using concept synonyms. REAL surpasses the previous zero-shot SOTA, using 400$\times$ less storage and 10,000$\times$ less training time! 
\end{abstract}

\vspace{-2mm}
\section{Introduction}
\label{sec:intro}

\begin{figure}[h]
    \centering
    \ \hspace{0.5mm} {\footnotesize {\bf (a)} freq. of ImageNet concepts} \hspace{4.5mm} {\footnotesize {\bf (b)} freq. vs. zero-shot accuracy} \  \\
    \vspace{-0.3mm}
    \includegraphics[width=.49\linewidth, clip=true,trim = 0mm 0mm 0mm 0mm]{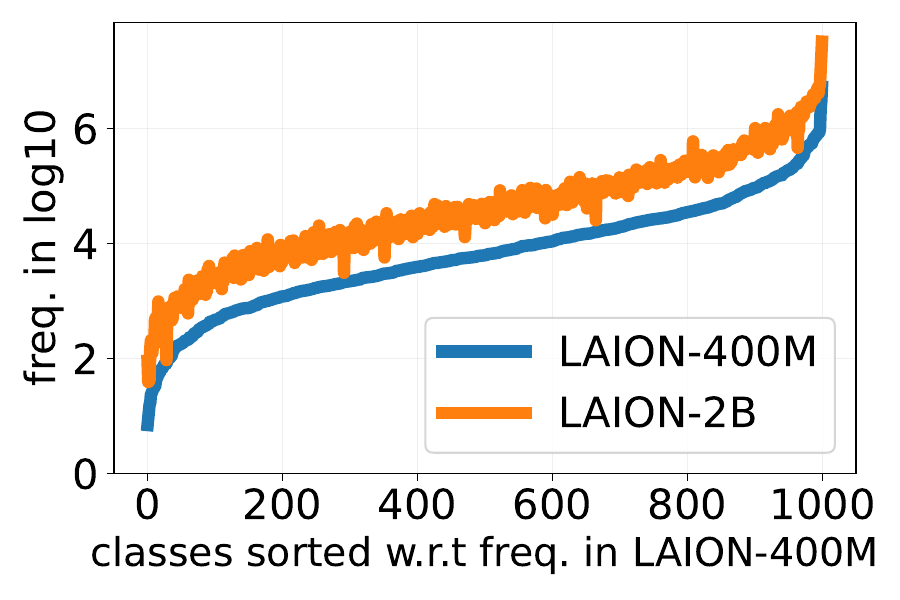} 
    \includegraphics[width=.49\linewidth, clip=true,trim = 0mm 0mm 0mm 0mm]{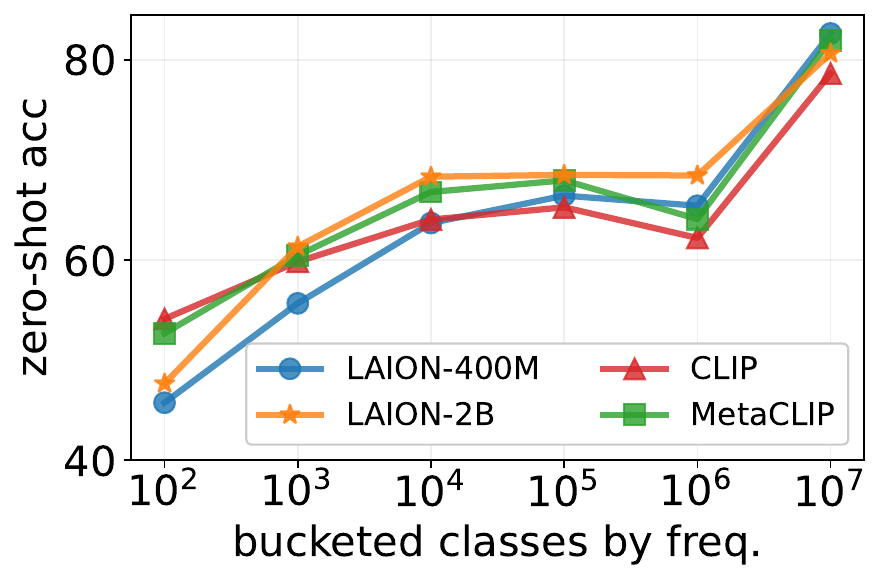} 
    \\
    \begin{enumerate}[]
    \item[\bf {\footnotesize (c)}]\setstretch{0.7}{\footnotesize Popular multimodal systems like visual chatbots and generative models struggle with tailed concepts. An example is the {\tt night snake}, a rare class found by our method on ImageNet.}
    \end{enumerate}
    \includegraphics[width=0.96\linewidth, clip=true,trim = 0mm 0mm 0mm 0mm]{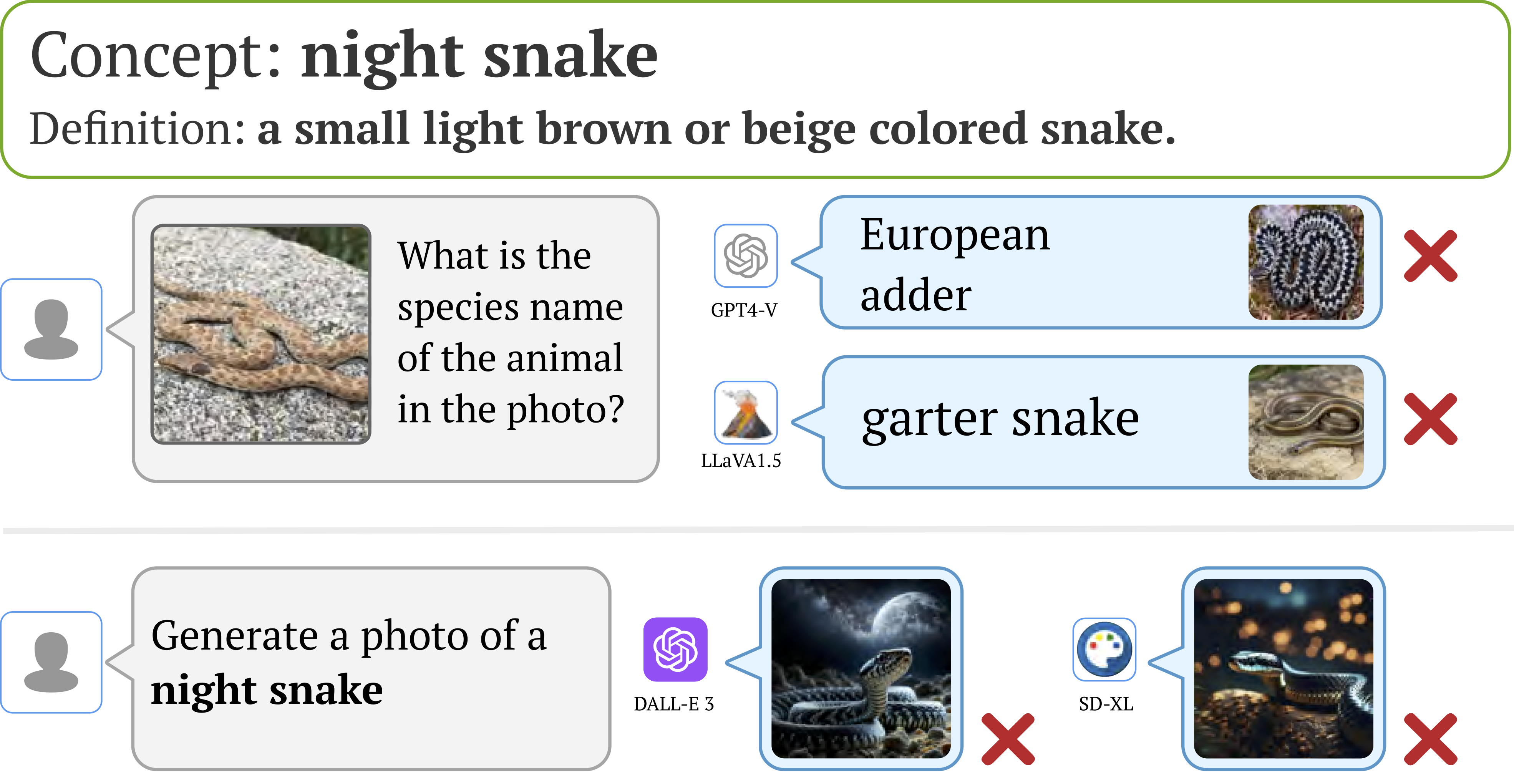}

    \vspace{-1mm}
    \caption{
    \small
    {\bf Vision-language models (VLMs) inherit long tails from their pretraining data. (a)}
    VLMs' pretraining datasets, such as LAION-400M~\cite{laion400m} and LAION-2B~\cite{laion5b}, exhibit long-tailed distributions for visual concepts defined in downstream tasks like ImageNet~\cite{deng2009imagenet}. We sort the 1K ImageNet classes according to their frequency in LAION-400M calculated with our concept frequency estimation method (cf. Fig.~\ref{fig:measure_concept_freq}).
    {\bf (b)}
    For zero-shot recognition, OpenCLIP models~\cite{openclip} trained on LAION-400M and LAION-2B respectively yield per-class accuracies that strongly correlate with the long-tailed concept frequencies (binned on a log-scale). Interestingly, other VLMs such as CLIP~\cite{clip} and MetaCLIP~\cite{xu2023metaclip} (trained on private data) also show similar imbalanced performances, likely because their web-scraped pretraining datasets follow similar long-tailed distributions of the real world. 
    {\bf (c)}
    Our method helps identify rare concepts, such as the {\bf \em night snake}, which is one of the most tailed ImageNet concepts. We show that state-of-the-art multimodal systems, including GPT-4V~\cite{gpt4v}, LLaVA~\cite{liu2023llava}, DALL-E 3~\cite{dalle3}, and SD-XL~\cite{stable_diffusion}, all fail to recognize or generate it. The supplement shows more examples.
    } 
\vspace{-5mm}
\label{fig:splashy-figure}
\end{figure}

Vision-language models (VLMs) such as CLIP~\cite{clip} play a pivotal role in mainstream multimodal systems, including visual chatbots~\cite{liu2023llava, openai2023gpt4} and text-to-image generation~\cite{ramesh2021zeroshot, Betker2023improving}.
Their efficacy largely stems from their web-scale image-text pretraining datasets like LAION~\cite{laion400m, laion5b} that cover a wide range of visual concepts.

{\bf Imbalanced performance of VLMs.}
Despite their strong capabilities in visual understanding,
VLMs often exhibit biased performance in downstream tasks. 
For instance, in zero-shot visual recognition tasks (which do not use training samples), CLIP~\cite{clip} achieves up to 80\% mean accuracy across 1K semantic classes on ImageNet~\cite{deng2009imagenet} but 
less than $<$10\% on specific classes such as {\tt night} {\tt snake} (Fig.~\ref{fig:splashy-figure}b). 
This motivates us to explore {\em why} VLMs are imbalanced, a crucial yet ever-neglected issue.

{\bf Why do VLMs exhibit imbalanced performance?} It is commonly believed that foundational VLMs inherit biases~\cite{mehrabi2021survey} from web-scale pretraining data. However, we find no direct evidence linking VLMs' imbalanced performance to the concept distribution in pretraining data, likely because there is no such tool for measuring {\em concept frequency} in large multimodal datasets like LAION~\cite{laion400m}.

\begin{figure}[t]
    \centering
    \includegraphics[width=1\linewidth]{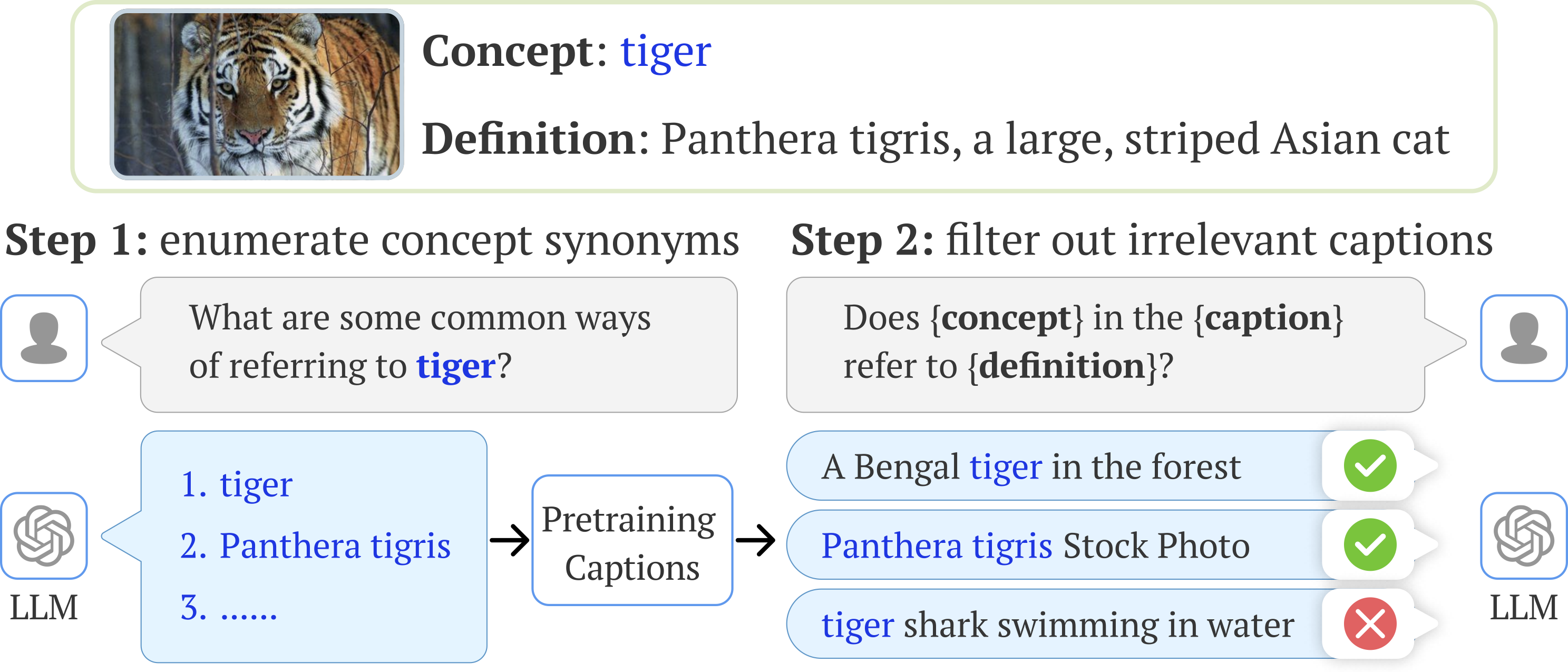}
    \vspace{-4mm}
    \caption{
    {\bf Using large language models (LLMs) to estimate concept frequency in a VLM's pretraining dataset.} We conduct the frequency estimation using publicly available LAION~\cite{laion400m} datasets.
    First, since a visual concept can be expressed in various ways, we ask an LLM (e.g., ChatGPT~\cite{openai2023gpt4}) to enumerate all its synonyms to search for potentially relevant pretraining texts. For example, for {\tt tiger}, we retrieve all captions that contain not only ``{\tt tiger}'' but also its synonyms such as ``{\tt Panthera tigris}''.
    Second, we filter out irrelevant captions that do not refer to the target concept by its definition. 
    For example, although ``{\it tiger shark swimming in water}'' contains ``{\it tiger}'', it actually refers to a type of shark, not the animal {\tt tiger} as defined by ``Panthera tigris, a large, striped Asian cat''. We conduct the filtering process by querying an LLM Llama-2~\cite{touvron2023llama} (cf. Section \ref{sec:analysis}).
    }  
    \label{fig:measure_concept_freq}
    \vspace{-5mm}
\end{figure}

{\bf Concept frequency estimation.}
Estimating class frequency in a typical classification dataset is straightforward, where counting class occurrences using annotated labels is sufficient. However, estimating concept frequency in VLMs' pretraining datasets is more complex because their free-form texts (or captions) contain significant {\em lexical variations}, e.g., a {\tt sneaker} might also be called a {\tt running shoe} or a {\tt trainer}. 
To address this, we leverage off-the-shelf large language models (LLMs).
Fig.~\ref{fig:measure_concept_freq} illustrates our approach, which begins by asking an LLM (such as ChatGPT~\cite{openai2023gpt4})
to enumerate all {\em synonyms} for a specific concept. 
We then use string matching to find all pretraining texts that contain the concept or its synonyms. However, due to {\em linguistic ambiguity}, the initially retrieved texts may contain irrelevant phrases, such as ``{\tt tiger shark in water}'' for the target concept
{\tt tiger} (a mammal). 
We again use an LLM to filter out such irrelevant texts, and conduct human studies to verify the accuracy of our frequency measure.
Our method for calculating concept frequency unveils three key insights: (1) it confirms that VLMs' pretraining data is indeed long-tailed (Fig.~\ref{fig:splashy-figure}a); (2) it shows VLMs perform better on well-represented concepts and worse on under-represented ones (Fig.~\ref{fig:splashy-figure}b);
and (3) it explains why recent multimodal systems (e.g., GPT-4Vision and DALL-E 3 in Fig.~\ref{fig:splashy-figure}c) struggle with rare concepts.
Our analysis also provides technical insights to counteract the bias in VLMs, leading to state-of-the-art zero-shot performance in downstream tasks such as image recognition.

{\bf State-of-the-art zero-shot recognition.} 
Motivated by our frequency estimation, we introduce {\bf RE}trieval-{\bf A}ugmented {\bf L}earning (REAL) to mitigate biased performance of zero-shot VLMs. REAL has two variants. First, observing that some synonyms are more frequent in VLM's pretraining data than the original concept names, we propose {\bf REAL-Prompt}. Specifically, we replace given concept names with their \emph{most frequent synonyms}. For example, {\tt cash} {\tt machine} is replaced with {\tt ATM}, which is ten times more frequent in LAION (Fig.~\ref{fig:REAL-prompt}). 
This minor change already surpasses costly human-engineered~\cite{clip} and LLM-enriched prompts like CuPL~\cite{CUPL} (cf. Table~\ref{tab:comparison_sota}).
Second, inspired by retrieval-augmented strategies~\cite{lewis2020retrieval, guu2020retrieval, liu2023learning, li2023internet, wallingford2023neural}, we introduce {\bf REAL-Linear}, which reuses relevant pretraining data to better adapt VLMs without using data from downstream tasks. The key idea is to retrieve a small, balanced set of images from pretraining data to train a robust linear classifier~\cite{lin2023multimodality}. In contrast to prior arts~\cite{liu2023learning,  wallingford2023neural} that perform costly {\em feature-based} retrieval by running VLMs to compute image or text features, our method implements {\em text-based} retrieval via string matching, achieving a significant boost in efficiency.
As a result, our REAL resoundingly outperforms the recent retrieval-augmented SOTA, REACT~\cite{liu2023learning}, using 400$\times$ less storage and 10,000$\times$ less training time (cf. Table \ref{tab:comparison_sota} and \ref{tab:comparison_sota_compute_cost})!

{\bf Contributions}.
We summarize our major contributions.
\begin{itemize}[topsep=0pt, partopsep=0pt, leftmargin=0.3in]
\item 
    We propose a method for estimating the frequency of visual concepts in VLMs' large-scale pretraining data. Our analysis, for the first time, exposes long-tailed concept distributions in popular datasets like LAION and reveals systematic failures of VLMs~\cite{gpt4v, dalle3, liu2023llava, stable_diffusion} in handling rare concepts.
\item 
    We propose REAL to address the biased performance of zero-shot VLMs. REAL establishes a new state-of-the-art in zero-shot recognition through its efficient prompting and retrieval-augmented training strategies. 
\end{itemize}

\section{Related Works}
\label{sec:related}

{\bf Biases in foundation VLMs.} 
Pretrained on large-scale multimodal datasets~\cite{birhane2021multimodal, wang2022revise, chuang2023debiasing}, VLMs often exhibit biases related to gender, race, and geography~\cite{mehrabi2021survey}, leading to imbalanced predictions in downstream tasks~\cite{agarwal2021evaluating, menon2022task}. Recent studies~\cite{debiasedpl, zhu2023generalized, shao2023investigating} seek to mitigate imbalanced predictions of VLMs by training on additional data from downstream tasks. Despite these efforts, there is no analysis of the imbalances within the pretraining data itself.
Our study presents the first examination of VLMs' pretraining datasets, revealing a long-tailed distribution of concepts that closely correlates with VLMs' imbalanced performance. Our analytical tool also identifies rare concepts that VLMs have insufficiently learned, thereby preventing biases in downstream applications.

{\bf Prompting VLMs for zero-shot recognition.}
VLMs excel in zero-shot recognition tasks, where only the names of target concepts are provided without corresponding training images. CLIP~\cite{clip} shows that putting given concept names in human-engineered prompt templates, such as ``a photo of a \{{\tt class}\}'' and 
``a demonstration of a \{{\tt class}\}'', often enhances zero-shot recognition. LLM-enriched approaches like DCLIP~\cite{menon2022visual} and CuPL~\cite{CUPL} create class-specific prompts by appending rich visual descriptions generated by LLM, for example, ``a tiger, which has sharp claws''. While most works focus on refining prompt templates~\cite{liu2023language, shen2022k, zhou2022learning}, they use the provided class names as is. 
A recent work~\cite{Parashar2023prompting} suggests that prompting with common English names instead of Latin scientific names improves zero-shot recognition of fine-grained species.
Differently, our REAL-Prompt replaces given class names with their most common synonyms found in the pretraining texts.
This simple change outperforms existing methods with much less ChatGPT querying costs.
Moreover, our approach can be combined with existing prompt templates to further improve performance.

{\bf Retrieval-augmented strategy.}
Introduced in the natural language processing (NLP) literature, this strategy addresses challenging tasks such as knowledge-based question-answering~\cite{guu2020retrieval,lewis2020retrieval} by retrieving relevant facts from an external knowledge source (e.g., the Internet or a pretraining dataset) to ground LLMs on the most accurate, up-to-date information.
To improve zero-shot visual recognition, 
recent works~\cite{liu2023learning, li2023internet, wallingford2023neural} finetune VLMs on images retrieved from VLMs' pretraining datasets.
While methods like REACT~\cite{liu2023learning} are effective, they demand significant computing resources, including hundreds of GPU hours and extensive memory for large-batch contrastive finetuning. 
In contrast, our REAL-Linear uses fast string matching to retrieve data based on concept synonyms, thus avoiding costly VLM-based feature extraction. 
Moreover, we train a parameter-efficient linear classifier atop the frozen VLM~\cite{lin2023multimodality}, which significantly enhances efficiency.
Our method not only sets a new state-of-the-art in zero-shot recognition, but also opens avenues for retrieval-augmented research within a modest computational budget.

\section{The Long-Tailed Concept Distribution}
\label{sec:analysis}

This section outlines our approach for estimating concept frequency in VLM's pretraining data and presents our key findings from the analysis.

\subsection{Concept Frequency Estimation}

The foremost challenge in estimating concept frequency is the sheer size of a VLM's pretraining dataset.
For example, 
the popular open-source LAION-400M~\cite{laion5b} dataset (used for training OpenCLIP~\cite{openclip}) takes 
$\sim$10TB of physical storage space. Instead, we estimate concept frequency directly from pretraining {\em texts}, eliminating the need to download images. This allows us to download only the text metadata, requiring only $\sim$60GB space for LAION-400M. Next, we proceed with the following two steps (cf. Fig.~\ref{fig:measure_concept_freq}).

{\bf Step 1: Deriving synonyms for target concepts.} 
A well-known issue in NLP is \emph{lexical variation},
meaning that a concept can be expressed in multiple ways. 
For example, ``{\tt sneaker}'' can be referred to as ``{\tt running shoes}'' or ``{\tt athletic footwear}'', and ``{\tt tiger}'' may also be called ``{\tt Panthera tigris}''.
To account for lexical variation, 
we first derive a list of synonyms\footnote{For simplicity, we use the term ``synonyms'' in a broader sense to encompass all forms of lexical variation, including but not limited to traditional synonyms, idiomatic expressions, and different phrasings that convey the same or similar meanings.} for a given visual concept.
To do so, we turn to an off-the-shelf LLM (e.g., ChatGPT~\cite{openai2023gpt4}), by querying a simple question ``{\em What are some common ways of referring to \{{\bf concept}\}}?''. Then, we use string matching to retrieve all pretraining texts containing these synonyms. This matching process is remarkably efficient --- it takes only 5 hours to retrieve 400M pretraining texts from the LAION-2B dataset for ImageNet's 1K concepts. Importantly, these derived concept synonyms can also be used to improve downstream zero-shot recognition, which we discuss in Sec.~\ref{sec:retrieval_finetuning}.

\begin{figure*}[t]
    \centering
    \includegraphics[width=1.0\linewidth, clip=true,trim = 0mm 0mm 0mm 0mm]{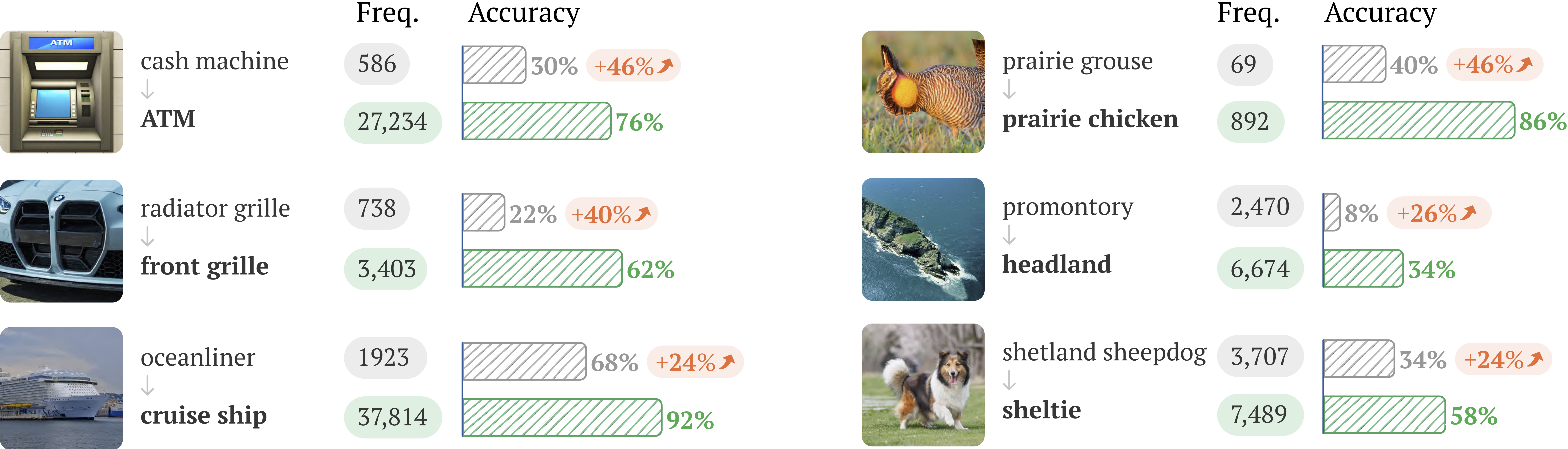} 
    \vspace{-3mm}
    \caption{
    {\bf Demonstration of REAL-Prompt}, 
    which replaces the given concept names (e.g., ``cash machine'') with their most frequent synonyms (e.g., ``ATM'') in the prompt template, e.g., ``\emph{a photo of \{concept\}}''.
    REAL-Prompt uses an LLM (ChatGPT) to obtain a list of synonyms for a given concept, followed by string matching to identify the most frequently occurring ones in pretraining texts. We demonstrate REAL-Prompt on some ImageNet concepts with their most frequent synonyms, frequencies (in LAION-400M), and per-class accuracies (OpenCLIP ViT-B/32). The simple name change in prompts significantly improves zero-shot recognition. We detail the procedure for REAL-Prompt in Sec.~\ref{sec:REAL-Prompt} and compare against prior works in Table~\ref{tab:comparison_sota}.
    }
    \vspace{-3mm}
    \label{fig:REAL-prompt}
\end{figure*}

{\bf Step 2: Filtering out irrelevant pretraining texts.} 
Using simple string matching may be inaccurate because it could retrieve irrelevant captions
due to {\em linguistic ambiguities}. 
For instance, 
the concept ``{\tt tiger}'', defined as ``{\em Panthera tigris, a large, striped Asian cat}'', might appear in irrelevant contexts in the retrieved captions such as ``{\it tiger shark swimming in water}'' and ``{\it Tiger Woods, a famous golf player}''. In these retrieved texts, the ``tiger'' actually refers to a shark species and a celebrity, respectively.
To tackle these ambiguities at an affordable cost, we utilize the state-of-the-art open-source LLM Llama-2~\cite{touvron2023llama}. For each retrieved text, we ask: 
\begin{verse}\small
Does \{{\tt concept}\} in the \{{\bf caption}\} \\
refer to \{{\bf definition}\}? 
\end{verse}
In real-world applications, concepts are typically defined in the labeling policies of downstream tasks~\cite{nuscenes2019, fong2021panoptic}.
In this work, to align with standard benchmarks, we adopt definitions from Elevater~\cite{li2022elevater}. Finally, retrieved captions that are identified as irrelevant to the target concept by the LLM are excluded. We only count the remaining retrieved text to estimate concept frequency.

\subsection{Discussions and Remarks}

{\bf Human-in-the-loop validation.}
As VLMs' pretraining data does not contain ground-truth concept labels, 
we perform manual validation to ensure LLM performs well in filtering the irrelevant texts.
To do so, we first construct a small validation set by downloading a balanced set of pretraining data (32 image-text pairs per concept).
Then, for each concept, we tune the concept definitions for Llama-2 till reaching $>$85\% retrieval precision on the validation set. In particular, since \cite{li2022elevater} releases multiple definitions per concept, we select the best ones that lead to the highest precision over the validation set.
For example, the class {\tt samoyed} in ImageNet refers to a dog breed; 
we find the definition ``{\it a breed of large white herding dog with a thick coat, native to the Ural Mountains}'' to be more precise than others, e.g., ``{\it a member of a people inhabiting northwestern Siberia}''.
To facilitate future research, we will open-source our code for LLM-based analysis and release all concept synonyms and filtered captions.

{\bf The prevalent long tails in VLMs.}
Our analysis reveals an ever-neglected long-tailed distribution of visual concepts (from standard benchmarks like ImageNet~\cite{deng2009imagenet}) within widely-used pretraining datasets like LAION-400M and LAION-2B (cf. Fig.~\ref{fig:splashy-figure}a). 
Additionally, we plot per-class zero-shot accuracies of OpenCLIP models (pretrained on LAION), establishing a strong correlation between the long-tailed distribution of concepts and the imbalanced performance of VLMs (cf. Fig.~\ref{fig:splashy-figure}b).
We also plot the per-class accuracies of CLIP~\cite{clip} and MetaCLIP~\cite{xu2023metaclip}, which are trained on private datasets.
Interestingly, they show similar imbalanced performance across ImageNet's concepts, likely because Internet data follows a similar long-tailed distribution.
We observe the same trend across eight more benchmarks (cf. Table~\ref{tab:more_freq_acc}), e.g., Flowers~\cite{flowers} and Pets~\cite{pets}.
Notably, our frequency estimation method helps find rare (tailed) concepts that challenge popular multimodal systems including the state-of-the-art GPT-4Vision~\cite{openai2023gpt4} and DALL-E 3~\cite{dalle3} (cf. Fig.~\ref{fig:sup_tail_concepts_failure_part1} and \ref{fig:sup_tail_concepts_failure_part2}). 
In sum, our analysis shows that long-tailed issues are prevalent in VLMs.

{\bf Are long tails inevitable in large-scale data curation?} 
Despite the imbalanced performance of CLIP~\cite{clip} and MetaCLIP~\cite{xu2023metaclip}, their pretraining datasets are actually created using a ``balanced'' sampling strategy. Specifically, they define 500K word phrases as web search queries to collect approximately equal numbers of image-text samples for each. However, our analysis indicates that these datasets (which are not fully disclosed to the public) might not be as balanced as intended. We identify key insights into the observed imbalances by examining \cite{xu2023metaclip}'s query statistics:
\begin{itemize}[topsep=0pt, partopsep=0pt, leftmargin=0.3in]
    \item {\bf Internet data are naturally long-tailed:} Despite balanced sampling in \cite{xu2023metaclip}, the resulting distribution is still long-tailed. For example, while each search query is capped at 20K images, the average is around $\sim$1.2K images per query. In fact, we find that over half of the queries, such as ``{\tt tailed frog}'' (a frog), ``{\tt gyromitra}'' (a fungus), and ``{\tt poke bonnet}'' (a traditional hat), have less than 50 samples, likely because these concepts are rare on the web.
    \item {\bf Limitations of query-based balancing:} Balancing per query does not guarantee a balanced distribution of concepts. For example, \cite{xu2023metaclip} inadvertently include overlapping queries such as ``{\tt sneaker}'' and ``{\tt running shoes}'', which can lead to the overrepresentation of certain concepts. Moreover, samples retrieved for a single query often contain other concepts. For instance, samples featuring ``{\tt keyboard}'' may also frequently include ``{\tt mouse}'' and ``{\tt computer}''.
\end{itemize}
\vspace{1mm}
As curating a perfectly balanced pretraining dataset is challenging, we recommend that researchers acknowledge the presence of long tails and prioritize addressing VLM's imbalanced performances in downstream applications. 

\section{Retrieval-Augmented Learning}
\label{sec:retrieval_finetuning}

To address the biased performance of VLMs in zero-shot recognition, we propose {\bf RE}trieval-{\bf A}ugmented {\bf L}earning (REAL), which improves performance without using any data from downstream tasks by retrieving pretraining data relevant to the target concepts. REAL has two variants: REAL-Prompt and REAL-Linear. REAL-Prompt is a novel prompting strategy that replaces the original concept name with its most frequent synonym found in pretraining texts. REAL-Linear retrieves images relevant to the concepts from the pretraining data to form a more balanced subset for training a robust linear classifier. Below we elaborate on these two methods.

\subsection{REAL-Prompt for Zero-Shot Prompting}
\label{sec:REAL-Prompt}

In our analysis, we discover that some synonyms for a concept might appear more frequently in pretraining texts than the concept itself. Therefore, we propose using the most frequent synonym of a concept to construct prompts. Specifically, we utilize an LLM (ChatGPT~\cite{openai2023gpt4}) to enumerate all synonyms for each concept. Next, we count their individual frequencies in the pretraining texts by string matching. We use the most frequent synonym of each concept in the prompt to construct an off-the-shelf classifier $W_{zs}$ for zero-shot recognition following CLIP~\cite{clip}.  
This simple change leads to significantly better zero-shot accuracy than using the original concept names (cf. Fig.~\ref{fig:REAL-prompt}) released by \cite{clip}, which are hand-crafted over one year. As shown in Table~\ref{tab:comparison_sota}, our approach also outperforms recent LLM-based prompting methods that use additional visual descriptors along with the given concept names (e.g., DCLIP~\cite{menon2022visual} and CuPL~\cite{CUPL}).

{\bf Synonym filtering using OpenCLIP's text encoder.} ChatGPT sometimes generates noisy synonyms. For example, for ``{\tt tiger}'', it lists ``{\tt big} {\tt cat}'' as a synonym, which could be easily confused with another ImageNet class ``{\tt tabby} {\tt cat}''. To address this, we use OpenCLIP's text encoder to filter out synonyms that might be confused with other downstream concepts. We retain only those synonyms that have the highest cosine similarity scores with their original class names. This filtering step is fully automated, ensuring a fair comparison with \cite{menon2022visual, CUPL} that perform LLM-based prompting without human input. We show that this step is crucial to REAL-Prompt's performance in Table~\ref{tab:t2t-filter}.

\begin{figure}[t]
    \centering
    \includegraphics[width=1\linewidth]{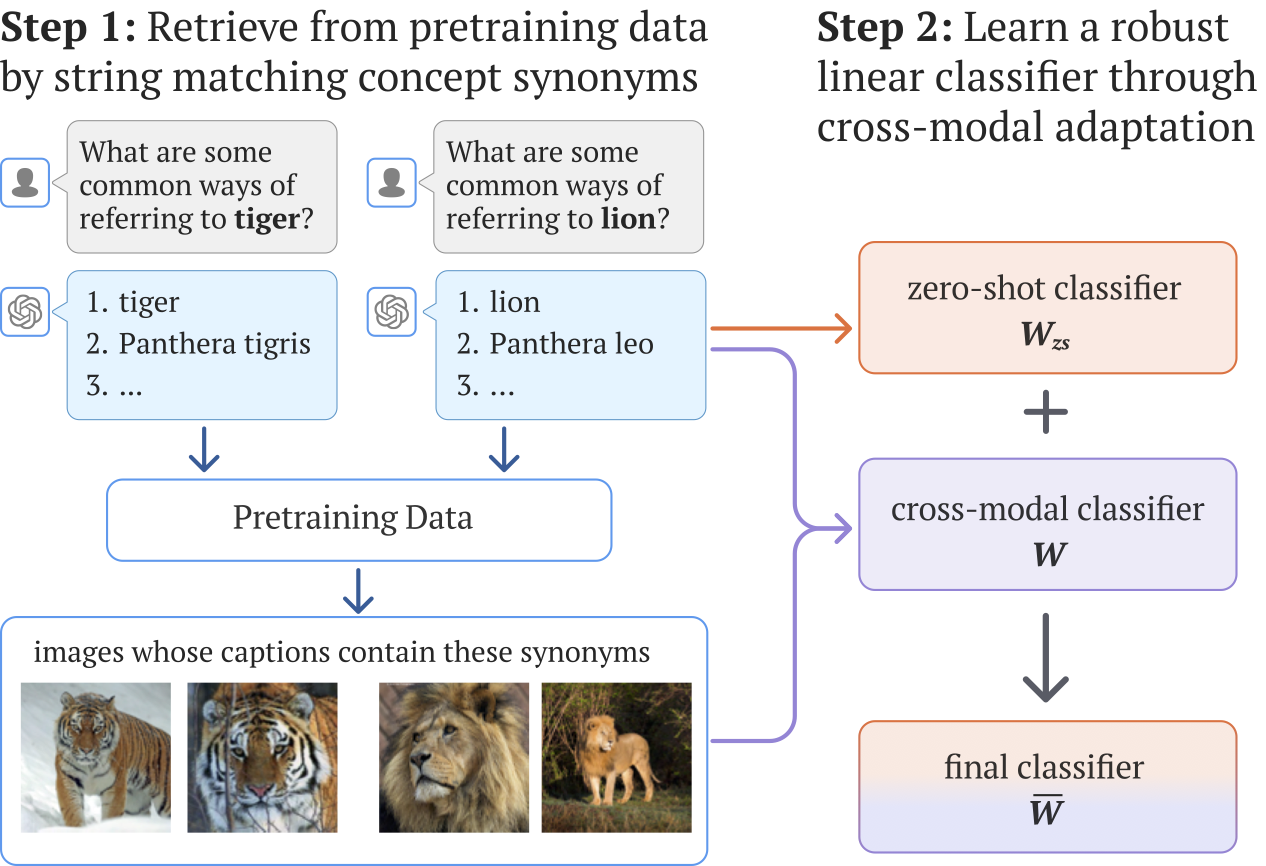}
    \vspace{-4mm}
    \caption{
    {\bf Flowchart of REAL-Linear.}
    First, it uses all synonyms of the given concepts to retrieve a class-balanced subset of pretraining images (e.g., 500 images per class from the dataset LAION-400M).
    Next, it learns a linear classifier $W$ atop the frozen VLM using cross-modal adaptation~\cite{lin2023multimodality}, and then ensembles it with the off-the-shelf classifier $W_{zs}$, whose weights are text prompt embeddings based on the most frequent synonyms. }
    \vspace{-3mm}
    \label{fig:REAL-linear}
\end{figure}

\subsection{REAL-Linear for Linear Classifier Training}
To further improve performance, REAL-Linear finetunes on images retrieved from pretraining data that are relevant to target concepts, as illustrated in Fig.~\ref{fig:REAL-linear}. 

{\bf Step 1: Retrieving data using concept synonyms.} For each concept, we retrieve pretraining data (LAION) whose textual captions contain any concept synonyms from REAL-Prompt. Then, we sort the retrieved data by the cosine similarity between their captions and the averaged text features (generated by OpenCLIP's text encoder using all concept synonyms). We select an equal number of the top-ranked images per concept (e.g., 500) to ensure a class-balanced dataset. 

{\bf Step 2: Training a robust linear classifier.}
To address the potential domain gap between the pretraining data and the downstream task, we construct a robust linear classifier $W$ using cross-modal adaptation~\cite{lin2023multimodality}. Concretely, we learn a linear classifier $W$ atop VLM's embeddings of the retrieved images and concept names, and then ensemble it with the zero-shot classifier $W_{zs}$ (i.e., REAL-Prompt): $\bar W = W + W_{zs}$.

{
\setlength{\tabcolsep}{0.775em}
\begin{table*}[t]
\small
\centering
\caption{\small
{\bf REAL outperforms existing methods on standard zero-shot recognition datasets.}
Within the zero-shot prompting paradigm, 
our REAL-Prompt that prompts with the {\em most frequent synonyms} of visual concepts (using OpenAI's templates~\cite{clip}) outperforms existing prompting approaches that adopt the original concept names, such as DCLIP~\cite{menon2022visual} and CuPL~\cite{CUPL}. 
Within the retrieval-augmented learning paradigm (without using any data from downstream tasks),
our REAL-Linear retrieves a class-balanced subset of pretraining data (500 examples per concept from LAION-400M), 
learns a linear classifier ensembled with the zero-shot classifier used by REAL-Prompt.
REAL-Linear rivals the recent method REACT~\cite{liu2023learning} (which retrieves 10K examples per concept), and importantly, uses 5\% of REACT's retrieved images and 1\% of its compute as detailed in Table~\ref{tab:comparison_sota_compute_cost}. 
We highlight the \textbf{best accuracy} in bold and underline the \underline{second best} numbers.
}
\vspace{-2mm}
\scalebox{0.95}{
\begin{tabular}{ll c c c c c c c c c}
    \toprule
    & Method & ImageNet & Flowers & Cars & Aircraft & Pets & Food & DTD & EuroSAT & Avg \\
    
    \midrule
    
    \multirow{7}{*}{\makecell{Zero-Shot\\Prompting}} & prompt template  \\
    & \quad \textit{``\{concept\}''} & 60.7  & 63.8  & 78.1  & 12.6  &83.3  &80.1  &48.8 &28.6 &57.0 \\
    
    &\quad \textit{``a photo of \{concept\}''} & 62.5 & 66.5 &77.2 &15.8  &84.0  &80.3  &52.8 &36.6&59.5  \\
    
    &\quad OpenAI templates~\cite{clip} & 62.9  & \underline{68.0}  &79.2  & 16.7  &86.7  &\underline{80.9}  &54.5 &\underline{51.5} & 62.6 \\
    
    & DCLIP~\cite{menon2022visual}  & 62.1 &--- &--- &--- & 84.6 & 80.1 & 51.9 & 36.8 &--- \\
    
    & CuPL~\cite{CUPL} & \bf{63.7} & 65.8 & \underline{80.0} & \underline{17.8} & \underline{87.4} & 79.5 & \underline{59.1} &--- &---  \\
    
    & \cellcolor{gray!15}REAL-Prompt & \cellcolor{gray!15}\underline{63.6} & \cellcolor{gray!15}\bf{76.6} &\cellcolor{gray!15}\bf{82.7} &\cellcolor{gray!15} \bf{18.0} & \cellcolor{gray!15}\bf{88.8} & \cellcolor{gray!15}\bf{81.0} &\cellcolor{gray!15}\bf{59.9} &\cellcolor{gray!15}{\bf 57.5}&\cellcolor{gray!15}{\bf 66.0} \\

    \midrule
    \multirow{4}{*}{\makecell{Retrieval\\Augmented}} & REACT (10K)~\cite{liu2023learning} \\
    & \quad Locked-Text & \underline{65.7}  & \underline{73.1}  & {\bf 88.5} & 24.5  & 89.2 & 81.8 & 49.8 &\underline{51.1} & \underline{65.5} \\
    
    & \quad Gated-Image & 64.2 & 72.3  & \underline{88.1} & 
    \underline{24.8}  & \underline{89.5} & {\bf 83.0} & 
    \underline{51.4} & 45.4 & 64.8\\
    
    &\cellcolor{gray!15}REAL-Linear (500) &\cellcolor{gray!15}\textbf{65.9} &
    \cellcolor{gray!15}\textbf{78.8} &
    \cellcolor{gray!15}84.4&
    \cellcolor{gray!15}\textbf{29.6}&
    \cellcolor{gray!15}\textbf{89.5}&
    \cellcolor{gray!15}81.4& 
    \cellcolor{gray!15}\textbf{61.5}& 
    \cellcolor{gray!15}\textbf{51.5}
    &\cellcolor{gray!15}\textbf{67.8} \\

    \bottomrule
\end{tabular}}
\vspace{-2.5mm}
\label{tab:comparison_sota}
\end{table*}
}

{\bf REAL-Linear's exceptional efficiency.}
REAL-Linear is significantly more efficient than the state-of-the-art REACT~\cite{liu2023learning} and can be done using only academic-scale computing resources. Unlike REACT, which requires downloading the whole pretraining dataset and running VLMs to extract features for all pretraining images and texts, REAL-Linear processes only pretraining texts via string matching. As a result, it can process all LAION-400M captions in one hour, whereas REACT needs 250 GPU hours just for feature extraction. In addition, REACT's contrastive training demands extensive resources to ensure performance, e.g., large batch size (4,096) and long training (256 GPU hours on 16 V100 GPUs). In contrast, our linear-probing approach trains in minutes on a modest GPU (12GB). Table~\ref{tab:comparison_sota_compute_cost} compares the efficiency between REACT and REAL.

\section{Experimental Results}
\label{sec:exp_result}


We show the state-of-the-art performance of REAL for zero-shot recognition, outperforming existing prompting and retrieval-augmented methods across standard benchmarks. We ablate the design choices of REAL, revealing technical insights that contribute to its superior performance. We show that REAL can be combined with existing methods for even better results. Moreover, we show that REAL-Prompt improves image generation of rare concepts using text-to-image models like DALL-E 3 and SD-XL.

\subsection{Experimental setup} 
{\bf Datasets and metric.}
We report mean per-class accuracy on standard classification benchmarks, including
ImageNet~\cite{deng2009imagenet},
Flowers~\cite{flowers},
Cars~\cite{cars}, 
Aircraft~\cite{aircraft}, 
Pets~\cite{pets}, 
Food~\cite{food}, 
DTD~\cite{dtd}, 
EuroSAT~\cite{helber2019eurosat},
and CUB~\cite{cub}. 
Moreover, we use variants of ImageNet to study the out-of-distribution (OOD) robustness of our methods, 
including 
ImageNet-V2~\cite{imagenet_v2}, 
ImageNet-Adversarial~\cite{imagenet_adversarial}, 
ImageNet-Rendition~\cite{imagenet_rendention}, and
ImageNet-Sketch~\cite{imagenet_sketch}. 
Table~\ref{tab:datasets} details these datasets. 

{\bf Compared methods.}
We compare against state-of-the-art zero-shot recognition methods for VLMs. We report various prompting strategies that directly use the given concept names, including prompt templates such as {\em ``\{concept\}''}, \emph{``a photo of \{concept\}''}, and {\bf OpenAI}'s hand-engineered templates~\cite{clip}. We also compare with LLM-based prompting methods, {\bf DCLIP}~\cite{menon2022visual} and {\bf CuPL}~\cite{CUPL}, which uses GPT to generate visual descriptions for constructing prompts. Finally, we compare our method with the retrieval-augmented SOTA {\bf REACT}~\cite{liu2023learning} that retrieves data using VLMs' features and performs contrastive finetuning. We report two variants of REACT: Locked-Text and Gated-Image.

{
\setlength{\tabcolsep}{0.33em}
\begin{table}[t]
\small
\centering
\caption{{\bf REAL boosts both head and tail performance.} We show that REAL-Prompt and REAL-Linear (500 retrieved examples per concept) achieve consistent improvement across all classes over the baseline using OpenAI templates~\cite{clip}.
On each dataset, we define the tail as the 20\% least frequent classes and the rest as the head, and report the averaged per-class accuracy over nine standard zero-shot recognition datasets (including CUB~\cite{cub}).
We report detailed improvements on each dataset in Table~\ref{tab:REAL_head_tail}.
}
\vspace{-2mm}
\scalebox{0.9}{    
\begin{tabular}{llllll}
            \toprule
             & \multirow{2}{*}{Method} & \multicolumn{2}{c}{ImageNet} & \multicolumn{2}{c}{Avg of 9 datasets} \\
             \cmidrule(r){3-4} \cmidrule(r){5-6}
             & & Head & Tail & Head & Tail \\
            \midrule
            \multirow{3}{*}{\makecell{LAION\\400M}} & OpenAI templates  &64.8& 55.2 & 65.7 & 52.5 \\
        
            & \cellcolor{gray!15}\rpl &\cellcolor{gray!15}65.4$^{\textcolor{Green}{+0.6}}$ &
            \cellcolor{gray!15}56.2$^{\textcolor{Green}{+1.0}}$ & 
            \cellcolor{gray!15}67.8$^{\textcolor{Green}{+2.1}}$ & 
            \cellcolor{gray!15}56.9$^{\textcolor{Green}{+4.4}}$ \\
            
            & \cellcolor{gray!15}\reall{}-Linear & \cellcolor{gray!15}67.8$^{\textcolor{Green}{+3.0}}$ & \cellcolor{gray!15}58.9$^{\textcolor{Green}{+3.7}}$ & 
            \cellcolor{gray!15}72.5$^{\textcolor{Green}{+6.8}}$ & 
            \cellcolor{gray!15}56.0$^{\textcolor{Green}{+3.5}}$ \\

            \midrule                    
            
            \multirow{3}{*}{\makecell{LAION\\2B}} & OpenAI templates & 68.0 & 61.0 & 68.6 & 58.4 \\
            
            & \cellcolor{gray!15}\rpl & 
            \cellcolor{gray!15}68.2$^{\textcolor{Green}{+0.2}}$ & 
            \cellcolor{gray!15}61.6$^{\textcolor{Green}{+0.6}}$ & 
            \cellcolor{gray!15}69.8$^{\textcolor{Green}{+1.2}}$ & 
            \cellcolor{gray!15}61.8$^{\textcolor{Green}{+3.4}}$\\
            
            &\cellcolor{gray!15}REAL-Linear & \cellcolor{gray!15}69.8$^{\textcolor{Green}{+1.8}}$ & \cellcolor{gray!15}64.8$^{\textcolor{Green}{+3.8}}$ & 
            \cellcolor{gray!15}76.2$^{\textcolor{Green}{+7.6}}$ &
            \cellcolor{gray!15}63.6$^{\textcolor{Green}{+5.2}}$ \\
            \bottomrule
\end{tabular}}
\vspace{-4mm}
\label{tab:comparison_sota_head_tail}
\end{table}
}

{\bf Implementation details.}
In this work, 
we ablate on a series of OpenCLIP VLMs \cite{openclip, cherti2023reproducible},
which are publicly available along with their two pretraining datasets, namely LAION-400M \cite{laion400m} and LAION-2B \cite{laion5b}.
We report the performance of OpenCLIP ViT-B/32 architecture in the main paper and show that REAL generalizes to other architectures in Table~\ref{tab:diff_archs}.
For our REAL-Linear that learns a linear classifier, 
we simply use the hyperparameters provided by prior work~\cite{lin2023multimodality, wiseft}.
We use a single GeForce RTX 2080 Ti (12GB) to train all the models and allocate 50GB of storage to host retrieved data for all the nine benchmark datasets.

\subsection{Results}

{\bf Using frequent synonyms improves zero-shot recognition.}
Table~\ref{tab:comparison_sota} (top half) compares REAL-Prompt (based on OpenAI's prompt templates) against other prompting-based methods like DCLIP~\cite{menon2022visual} and CuPL~\cite{CUPL}, which use GPT to generate visual descriptions. REAL-Prompt significantly outperforms them simply by replacing the original concept names with their most frequent synonyms. This highlights the need to reconsider concept names in prompts. REAL-Prompt is also much cheaper because it queries 
ChatGPT for synonyms, which are shorter than the rich visual descriptions queried by DCLIP and CuPL. For every 1K concepts, DCLIP and CuPL generate 50K ($\$$0.5) and 500K ($\$$5) tokens using ChatGPT, respectively, while REAL-Prompt only requires 10K tokens ($\$$0.1).

{\bf REAL-Linear achieves the state-of-the-art.}
Table~\ref{tab:comparison_sota} (bottom half) compares our REAL-Linear against the retrieval-augmented SOTA REACT~\cite{liu2023learning}.
REAL-Linear achieves $\sim$3\% higher accuracy averaged across eight benchmarks than REACT, while using only 500 retrieved images per concept compared to REACT's 10K. 
Importantly, REAL-Linear is significantly more efficient (cf. Table~\ref{tab:comparison_sota_compute_cost}): it requires $\sim$1\% of REACT's computes, making it more accessible to the research community. It also outperforms another recent method NeuralPriming~\cite{wallingford2023neural} under its experimental setup (cf. Table~\ref{tab:prime_vs_real}).

{\bf REAL improves accuracy on tail classes.} Table~\ref{tab:comparison_sota_head_tail} shows that REAL boosts performance for both tail (least frequent 20\%) and head (the rest 80\%) classes on ImageNet and all nine datasets. For more specific improvements on each dataset, see Table~\ref{tab:REAL_head_tail}.

{\bf REAL benefits existing zero-shot methods.} 
Our methods can be readily applied with existing methods to further improve performance. Table~\ref{tab:additive_real-prompt} shows that REAL-Prompt's most frequent synonyms can be applied on any prompt templates, including LLM-enriched ones like DCLIP~\cite{menon2022visual} and CuPL~\cite{CUPL}. Likewise, REAL-Linear can be applied on top of REACT's finetuned OpenCLIP models, which achieves even better performance as shown in Table~\ref{tab:REAL_robust}.

{\bf Ablation studies for REAL-Linear.} To understand REAL-Linear's superior performance, we conduct several experiments, with key insights summarized below. Table~\ref{tab:synonyms_diversity} shows that using all concept synonyms, as opposed to just the original concept names, can help retrieve more diverse pretraining data, improving the averaged accuracy by 4\%. Table~\ref{tab:ablation_finetune} demonstrates that learning the linear classifier with both text and image features using cross-modal WiSE-FT~\cite{lin2023multimodality, wiseft} leads to a 6.4\% increase compared to linear probing with only image features. Lastly, Table~\ref{tab:ablation_shots} shows that increasing the retrieval size from 100 to 500 per concept yields a modest accuracy improvement of 0.9\%. Based on this, we adopt 500 as the standard retrieval size for our experiments.

\begin{figure*}[ht]
    \centering
    \small
    \begin{tabular}{p{17cm}}
    \includegraphics[width=1.0\linewidth, clip=true,trim = 0mm 0mm 4.5mm 0mm]{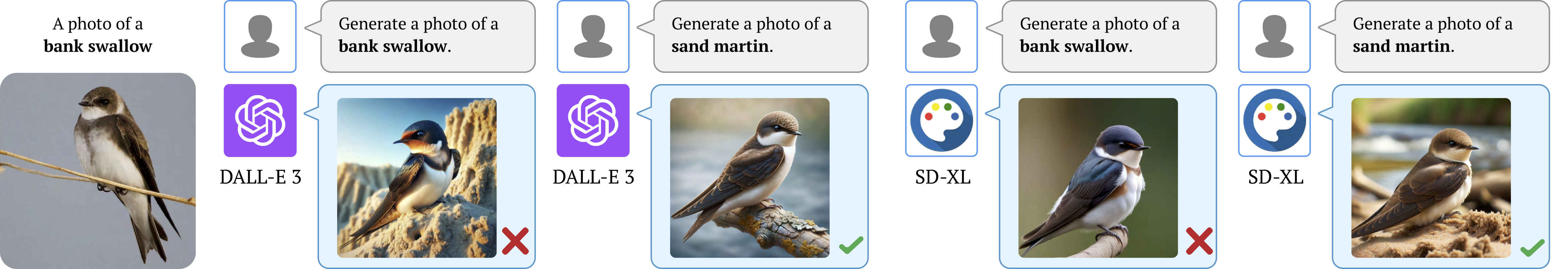}\\
    \midrule
    \includegraphics[width=1.0\linewidth, clip=true,trim = 0.5mm 0mm 4.5mm 0mm]{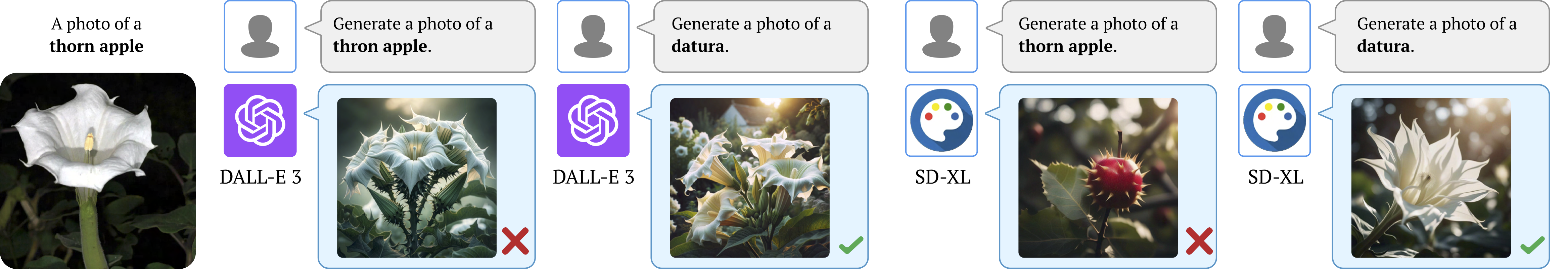}
    \end{tabular}
    \vspace{-4.5mm}
    \caption{\textbf{Improving image generation using REAL-Prompt.}
    We show two rare concepts identified by our frequency estimation where DALL-E 3 and SD-XL struggle to generate accurate images: {\tt bank} {\tt swallow} (top, a bird from the CUB dataset) and {\tt thorn apple} (bottom, a flower from the Flowers dataset). Using their original names, both DALL-E 3 and SD-XL incorrectly render the bird's colors. Additionally, DALL-E 3 erroneously adds thorns to the flower, while SD-XL depicts an apple with literal thorns. Instead, using the most frequent synonyms ({\tt sand martin} for {\tt bank} {\tt swallow}, {\tt datura} for {\tt thorn} {\tt apple}) as found by REAL-Prompt results in both systems generating accurate images. See more examples in Fig.~\ref{fig:sup_improve-dalle3_part1} and \ref{fig:sup_improve-dalle3_part2}. 
    }
\label{fig:improve-dalle3}
\end{figure*}

{\bf Improving image synthesis using REAL-Prompt.}
Fig.~\ref{fig:improve-dalle3} shows that, while state-of-the-art generative models such as DALL-E 3~\cite{dalle3} and SD-XL~\cite{stable_diffusion} 
may fail to generate correct images for some rare concepts (identified by our frequency estimation on LAION-400M), replacing the rare concepts used in prompts with their most frequent synonyms (found by REAL-Prompt) can help generate more accurate images. More qualitative examples can be found in Fig.~\ref{fig:sup_improve-dalle3_part1} and \ref{fig:sup_improve-dalle3_part2}.

{
\setlength{\tabcolsep}{0.65em}
\begin{table}[t]
\small
\caption{
{\bf Compute cost comparison between REACT \cite{liu2023learning} and our REAL-Linear.}
We compare the resources required for each method on the ImageNet experiment. Clearly, REAL-Linear (retrieving 500 pretraining images per concept) uses much less compute than REACT, e.g.,
retrieving 1000$\times$ less images,
using 400$\times$ less storage and 10,000$\times$ less training time.
}
\vspace{-3mm}
\centering
\scalebox{0.84}{    
\begin{tabular}{llccc}
            \toprule
            Stage & Resource & \makecell{REACT\\ \cite{liu2023learning}}
            & \makecell{REAL\\(500)} & {\makecell{Relative Cost}} \\
            \midrule
            \multirow{3}{*}{Retrieval} & retrieved examples & 400M & 0.5M & \textcolor{Green}{0.1\%} 
            \\
            & time & 200 hrs & 6 hrs & \textcolor{Green}{3\%} 
            \\
            & storage & 10 TB& 25 GB & \textcolor{Green}{0.25\%}  
            \\
            \midrule
            \multirow{4}{*}{Learning} & training images & 10M & 0.5M & \textcolor{Green}{5\%} 
            \\
            & time & 256 hrs & 2 mins & \textcolor{Green}{0.01\%}
            \\
            & \# of learned parameters & 87M & 0.5M & \textcolor{Green}{0.6\%}
            \\
            & GPU memory & 256 GB & 2 GB & \textcolor{Green}{0.8\%}
            \\
            \bottomrule
\end{tabular}
}
\label{tab:comparison_sota_compute_cost}
\end{table}
}

{
\setlength{\tabcolsep}{0.7em}
\begin{table*}[t]
\small
\centering
\caption{
{\bf Improvements using REAL-Prompt with existing prompting methods.}
Our REAL-Prompt can be combined with existing prompt templates including OpenAI's hand-engineered templates and LLM-enriched templates~\cite{clip} like DCLIP~\cite{menon2022visual} and CuPL~\cite{CUPL}. On datasets such as Flowers, DTD, and EuroSAT, 
integrating REAL-Prompt results in an accuracy boost of 5$\sim$8\%. }
\vspace{-3.5mm}
\label{tab:additive_real-prompt}
\scalebox{0.96}{
\begin{tabular}{llllllllll}
\toprule
 Prompting Method&
  ImageNet &
  Flowers &
  Cars &
  Aircraft &
  CUB &
  Pets &
  Food &
  DTD &
  EuroSAT \\ \midrule
  
OpenAI templates~\cite{clip} & 62.9 & 68.0 & 79.2 & 16.7 & 63.8 & 86.7 & \underline{80.9} & 54.5 & \underline{51.5} \\
\rowcolor{gray!15}
\quad + \rpl &
  63.6$^{\textcolor{Green}{+0.7}}$ &
  \textbf{76.6}$^{\textcolor{Green}{+8.6}}$ &
  \textbf{82.7}$^{\textcolor{Green}{+3.5}}$ &
  \underline{18.0}$^{\textcolor{Green}{+1.3}}$ &
  64.0$^{\textcolor{Green}{+0.2}}$ &
  \textbf{88.8}$^{\textcolor{Green}{+2.1}}$ &
  \textbf{81.0}$^{\textcolor{Green}{+0.1}}$ &
  \textbf{59.9}$^{\textcolor{Green}{+5.4}}$ &
  \textbf{57.5}$^{\textcolor{Green}{+6.0}}$ \\
  
DCLIP~\cite{menon2022visual} &
  62.1 &
  -- &
  -- &
  -- &
  \underline{64.5} &
  84.6 &
  80.1 &
  51.4 &
  36.8 \\
  \rowcolor{gray!15}
  \quad + \rpl &
  62.9$^{\textcolor{Green}{+0.8}}$ &
  -- &
  -- &
  -- &
  \textbf{64.7}$^{\textcolor{Green}{+0.2}}$ &
  \underline{88.1}$^{\textcolor{Green}{+3.5}}$ &
  80.0$^{\textcolor{Red}{-0.1}}$ &
  55.5$^{\textcolor{Green}{+4.1}}$ &
  36.9$^{\textcolor{Green}{+0.1}}$ \\
  
CuPL~\cite{CUPL} &
  \underline{63.7} &
  65.8 &
  80.0 &
  17.8 &
  -- &
  87.4 &
  79.5 &
  59.1 &
  -- \\
  \rowcolor{gray!15}
 \quad + \rpl &
  \textbf{64.2}$^{\textcolor{Green}{+0.5}}$ &
  72.3$^{\textcolor{Green}{+6.5}}$ &
  \underline{81.7}$^{\textcolor{Green}{+1.7}}$ &
  \textbf{18.3}$^{\textcolor{Green}{+0.5}}$ &
  -- &
  88.0$^{\textcolor{Green}{+0.6}}$ &
  79.5$^{\textcolor{Gray}{+0.0}}$ &
  \underline{59.3}$^{\textcolor{Green}{+0.2}}$ &
  -- \\ \bottomrule
\end{tabular}}
\end{table*}
}

{
\setlength{\tabcolsep}{0.35em}
\begin{table}[t]
\small
\centering
\caption{
{\bf Enhancing REACT's robustness with REAL-Linear.}
Our REAL-Linear (using 500 retrieved images per concept), when applied to REACT~\cite{liu2023learning}'s finetuned OpenCLIP models (ViT-B/32 trained on LAION-400M), improves zero-shot accuracy across various challenging ImageNet variants. These variants, including ImageNet-V2~\cite{imagenet_v2}, 
ImageNet-Adversarial~\cite{imagenet_adversarial}, 
ImageNet-Rendition~\cite{imagenet_rendention}, and
ImageNet-Sketch~\cite{imagenet_sketch}, are specifically designed to assess model robustness against domain shifts.
} 
\vspace{-3mm}
\scalebox{0.82}{
\begin{tabular}{llllll}
\toprule
\multirow{2}{*}{Method} & \multirow{2}{*}{\makecell[c]{ImageNet}} & \multicolumn{4}{c}{$\rightarrow$ ImageNet Variants} \\ 
\cmidrule(l){3-6} & & V2 \cite{imagenet_v2} & A \cite{imagenet_adversarial} & R \cite{imagenet_rendention} & S \cite{imagenet_sketch} \\
\midrule
OpenAI templates~\cite{clip} &62.9 &55.1 &21.7 &73.5 &49.4             \\

\rowcolor{gray!15}
REAL-Linear 
&65.9$^{\textcolor{Green}{+3.0}}$ 
&57.3$^{\textcolor{Green}{+2.2}}$ 
&22.7$^{\textcolor{Green}{+1.0}}$ 
&73.9$^{\textcolor{Green}{+0.4}}$ 
&50.9$^{\textcolor{Green}{+1.5}}$             \\
\midrule

REACT Locked-Text &65.7 &57.2 &20.3 &77.6 &54.8 \\

\rowcolor{gray!15}
\quad + REAL-Linear 
&67.7$^{\textcolor{Green}{+2.0}}$ 
&59.1$^{\textcolor{Green}{+1.9}}$
&21.3$^{\textcolor{Green}{+1.0}}$
&78.1$^{\textcolor{Green}{+0.5}}$ 
&55.9$^{\textcolor{Green}{+1.1}}$ \\ 

\midrule

REACT Gated-Image &64.2 &56.3 &21.1 &75.9 &52.4  \\

\rowcolor{gray!15}
\quad + REAL-Linear
&{66.9$^{\textcolor{Green}{+2.7}}$} 
&{\makecell{59.1$^{\textcolor{Green}{+2.8}}$}} 
&{\makecell{21.7$^{\textcolor{Green}{+0.6}}$}} 
&{\makecell{76.8$^{\textcolor{Green}{+0.9}}$}} 
&{\makecell{54.2$^{\textcolor{Green}{+1.8}}$}} \\
\bottomrule
\end{tabular}
}
\label{tab:REAL_robust}
\vspace{-5mm}
\end{table}
}

\subsection{Discussions}
\label{sec:discussions}

{\bf Broad impacts}.
Our work has positive societal impacts. Our concept frequency estimation method explains {\em why} VLMs are biased (or imbalanced) by confirming the long-tailed distribution of concepts in their pretraining data. By identifying concepts that VLMs have insufficiently learned, we can implement targeted measures to prevent unfair or biased predictions related to these concepts.

{\bf Limitations and future work}.
We acknowledge several limitations in our methods. First, while we offer a method to estimate concept frequency, we cannot accurately evaluate its precision and recall due to the absence of ground-truth annotations of the pretraining data.
Second, our estimation method relies only on the textual captions which may overlook other visual concepts that are present in the images but not in their captions.
Lastly, filtering ambiguous captions using off-the-shelf LLMs for each caption-concept pair is time-consuming.
We expect the future work to address these limitations.

\section{Conclusions}
\label{sec:conclusions}
We investigate the critical yet ever-neglected long-tailed issues of Vision-Language Models (VLMs). We use large language models (LLMs) to estimate concept frequency in VLMs' large-scale multimodal pretraining data, uncovering long-tailed distributions for concepts used in downstream tasks.
Crucially, we demonstrate a strong correlation between the long-tailed concept distributions and VLMs' imbalanced zero-shot performance.
To address this imbalance, we propose retrieval-augmented learning (REAL), with two variants: REAL-Prompt and REAL-Linear. 
REAL-Prompt replaces original class names from downstream tasks with their most common synonyms found in the pretraining texts, outperforming both human-engineered and LLM-enriched prompts. On the other hand, REAL-Linear leverages concepts synonyms to fetch a balanced subset of pretraining data for training a robust linear classifier atop the frozen VLM, surpassing the previous SOTA using $400\times$ less storage and 10,000$\times$ less training time. Finally, we highlight that modern text-to-image generators (e.g., DALL-E 3 and SD-XL) often fail to generate images for the rare concepts identified by our frequency estimation method. 
By applying REAL-Prompt, we demonstrate that using common synonyms helps generate more accurate images.

\section*{Acknowledgements}
{
We thank Tiffany Ling for polishing figures.
This work was partially supported by the University of Macau (SRG2023-00044-FST), and NSFC (No.62206256).
Shubham Parashar acknowledges the support from the CSE Department at Texas A\&M University.
}

\newpage
\clearpage
{
    \small
    \bibliographystyle{ieeenat_fullname}
    \bibliography{main}
}

\newpage
\clearpage

{
   \newpage
       \twocolumn[
        \centering
        \Large
        \textbf{\thetitle} \\ \vspace{0.5em} {Supplementary Material}\\
        \large
        \vspace{10mm}
       ] 
   }

\renewcommand{\thesection}{\Alph{section}}
\setcounter{section}{0}

\section*{}
\begin{center}
    \emph{\bf \em \large Outline}
\end{center}
{This document supplements the main paper with comprehensive analyses and ablations. Below outlines the document.
\begin{itemize}[topsep=-3pt, partopsep=3pt, leftmargin=0.3in]
\item {\bf Section \ref{sec:dataset_details}.} We provide details of the nine benchmark datasets and the four ImageNet variants.

\item {\bf Section \ref{sec:sup_concept_freq}.} We report our estimated concept frequency on the other eight benchmark datasets.

\item {\bf Section \ref{sec:sup_benefits_real}.} We report REAL performance on head and tail classes across nine benchmark datasets.

\item {\bf Section \ref{sec:sup_real}.} We attach all implementation details of REAL for reproducibility.

\item {\bf Section \ref{sec:sup_ablations}.} We present further ablations of REAL-Linear to highlight the importance of synonym-based retrieval and cross-modal adaptation. 

\item {\bf Section \ref{sec:sup_generalization}.} We show that the performance gain of REAL can generalize across different architectures, pretraining datasets, and prompt templates.

\item {\bf Section \ref{sec:sup_failure}.} We show more failures of state-of-the-art multimodal systems (visual chatbots and text-to-image generative models) on diverse tailed concepts.

\item {\bf Section \ref{sec:real_prompt_generative}.} We qualitatively show that REAL-Prompt can help generate images featuring rare concepts.

\end{itemize}
}

\section{Dataset Details}
\label{sec:dataset_details}
Table~\ref{tab:datasets} shows the details of the nine benchmarks, including the number of classes and the size of testset.
These datasets are widely used in the research community of zero-shot recognition.

\begin{table}[ht]
\centering
\caption{\small \textbf{Details of thirteen benchmark datasets.} }
\vspace{-3mm}
\renewcommand{\arraystretch}{1.2}
\resizebox{\linewidth}{!}{
\begin{tabular}{lrrl}
\toprule 
Dataset & \#Classes & \#Testing data & Remark \\
\midrule
Flowers~\cite{flowers} & 102 & 2,463 & flower classification \\
Cars~\cite{cars} & 196 & 8,041 & car (brand and year) classification \\
Aircraft~\cite{aircraft} & 100 & 3,333 & aircraft classification \\
Pets~\cite{pets} & 37 & 3,669 & domestic pet classification \\
Food~\cite{food} & 101 & 30,300 & food classification \\
DTD~\cite{dtd} & 47 & 1,692 & texture classification \\
EuroSAT~\cite{helber2019eurosat} & 10 & 8,100 & satellite imagery classification \\
CUB~\cite{cub} & 200 & 5,794 & bird classification \\
\hline
ImageNet~\cite{deng2009imagenet} & 1,000 & 50,000 & wordnet categories classification \\
ImageNet-V2~\cite{imagenet_v2} & 1,000 & 30,000 & an ImageNet variant of temporal shift \\
ImageNet-A~\cite{imagenet_adversarial} & 200 & 7,500 & an ImageNet variant of adversarial samples  \\
ImageNet-R~\cite{imagenet_rendention} & 200 & 30,000 & an ImageNet variant of artistic renditions \\
ImageNet-S~\cite{imagenet_sketch} & 1,000 & 50,000 & an ImageNet variant of sketches \\
    \bottomrule
    \end{tabular}
}
\label{tab:datasets}
\vspace{-3mm}
\end{table}

\section{Results of Concept Frequency Estimation}
\label{sec:sup_concept_freq}

\begin{figure}[ht]
\centering
\small
\scalebox{0.87}{
\begin{tabular}{cc}
\includegraphics[width=0.25\linewidth]{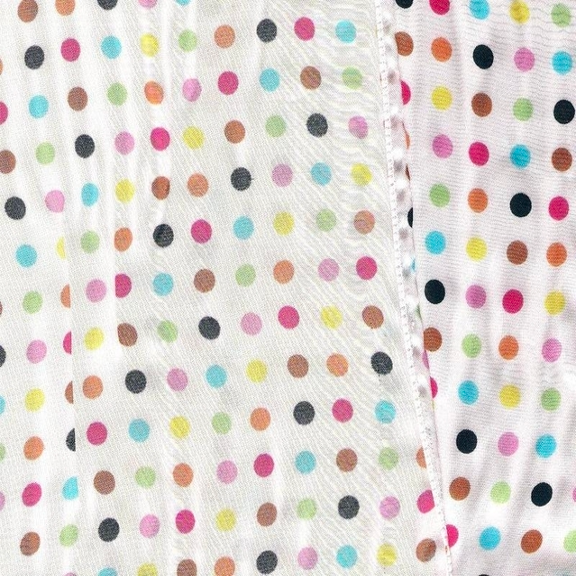}
\includegraphics[width=0.25\linewidth]{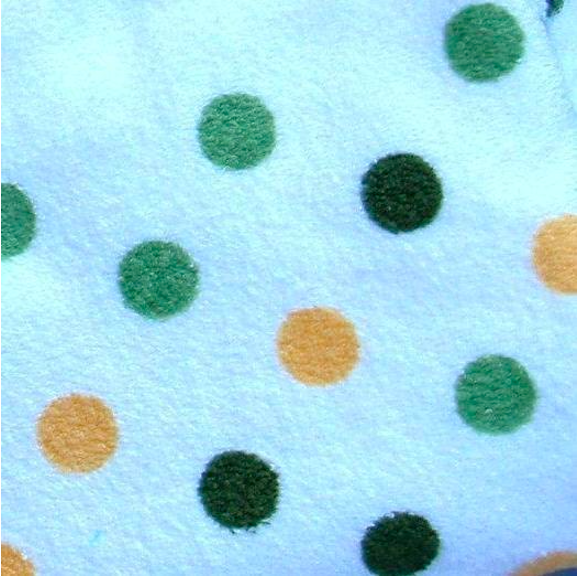} &
\includegraphics[width=0.25\linewidth]{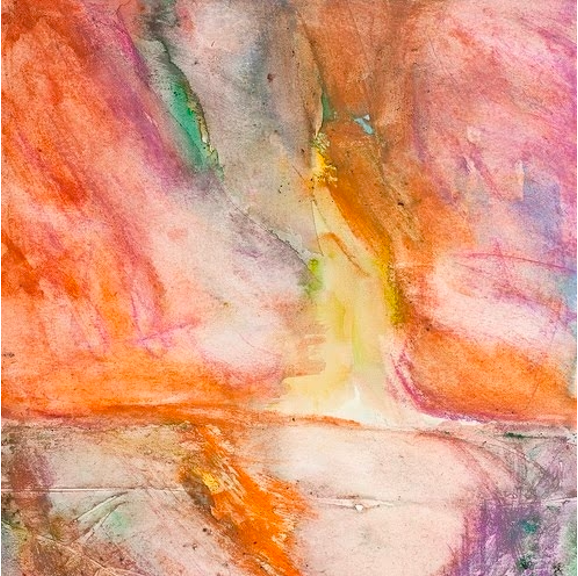}
\includegraphics[width=0.25\linewidth]{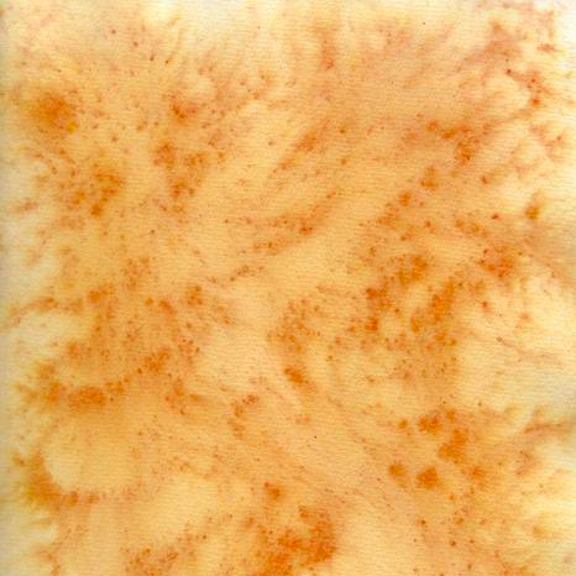}
\\
(a) {\tt dotted} vs. {\tt polka dotted} & (b) {\tt smeared} vs. {\tt stained}\\
\end{tabular}
}
\vspace{-3mm}
\caption{\small {\bf Classes in the DTD~\cite{dtd} dataset can be semantically ambiguous.} The texture class {\tt dotted} is a super-set of another class {\tt polka dotted}. For another case, people use the class name {\tt smeared} and {\tt stained} interchangeably.   
}
\vspace{-3mm}
\label{fig:dtd}
\end{figure}

In Table~\ref{tab:more_freq_acc}, we plot the concept frequency calculated using our proposed method for the other eight benchmark datasets. Surprisingly, we find that all of them follow an imbalanced distribution (as measured in LAION). Moreover, we plot the per-class zero-shot accuracies grouped by concept frequency and confirm a strong correlation between concept frequency and zero-shot accuracy in the majority of the datasets except for the DTD dataset. For DTD, we find that certain classes can overlap with others. For example, the {\tt dotted} and {\tt polka dotted}, {\tt smeared} and {\tt stained} (see Figure~\ref{fig:dtd}). Such ambiguous labeling makes DTD an outlier for our frequency analysis. 

\begin{table*}[h!]
\small
\centering
\caption{\small {\bf Vision-language models (VLMs) inherit long tails from their pretraining data.} We show that concepts from the other eight benchmark datasets all follow a long-tailed distribution in the pretraining datasets (e.g. LAION-400M~\cite{laion400m}, LAION-2B~\cite{laion5b}). The strong correlation between concept frequency and accuracy prevalently exists among the datasets. For DTD, the trend deteriorates because of the ambiguous labeling of class names (see Figure~\ref{fig:dtd}).  
}
\vspace{-3mm}
\scalebox{0.9}{
\begin{tabular}{cc|cc}
\toprule
\textbf{concept frequency} & \textbf{freq. vs. zero-shot acc.} & \textbf{concept frequency} & \textbf{freq. vs. zero-shot acc.}  \\
\hline
\multicolumn{2}{c|}{(a) CUB} & \multicolumn{2}{c}{(b) Food} \\
\includegraphics[width=0.25\linewidth]{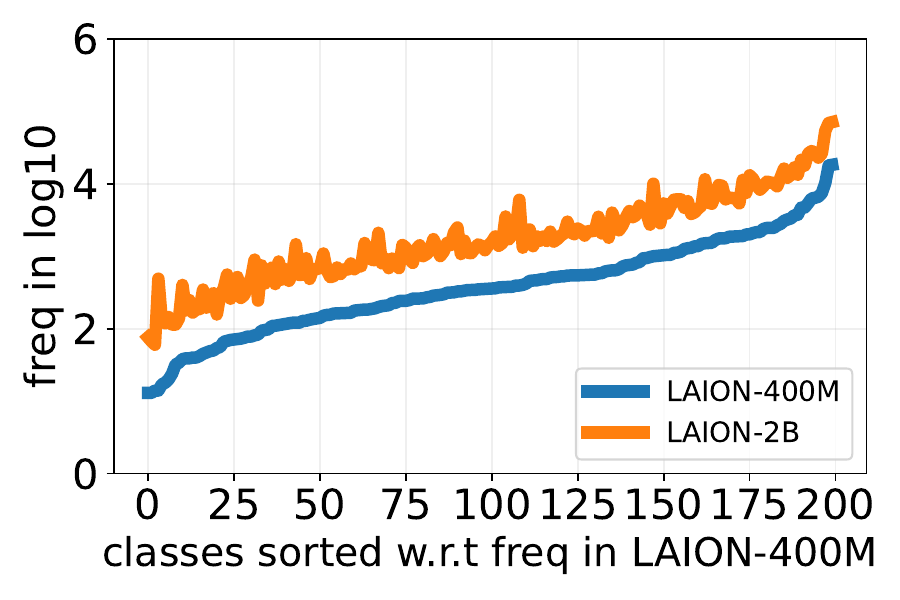} & \includegraphics[width=0.25\linewidth]{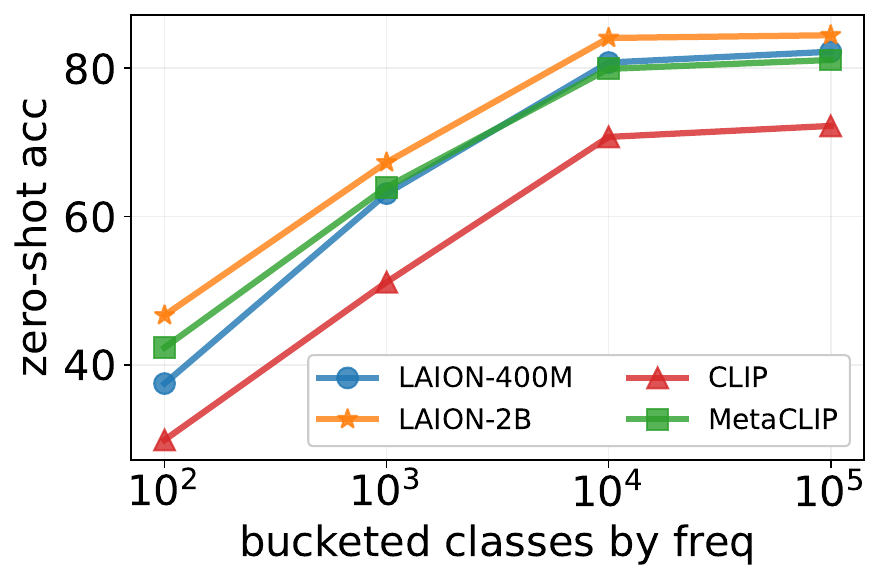} & \includegraphics[width=0.25\textwidth]{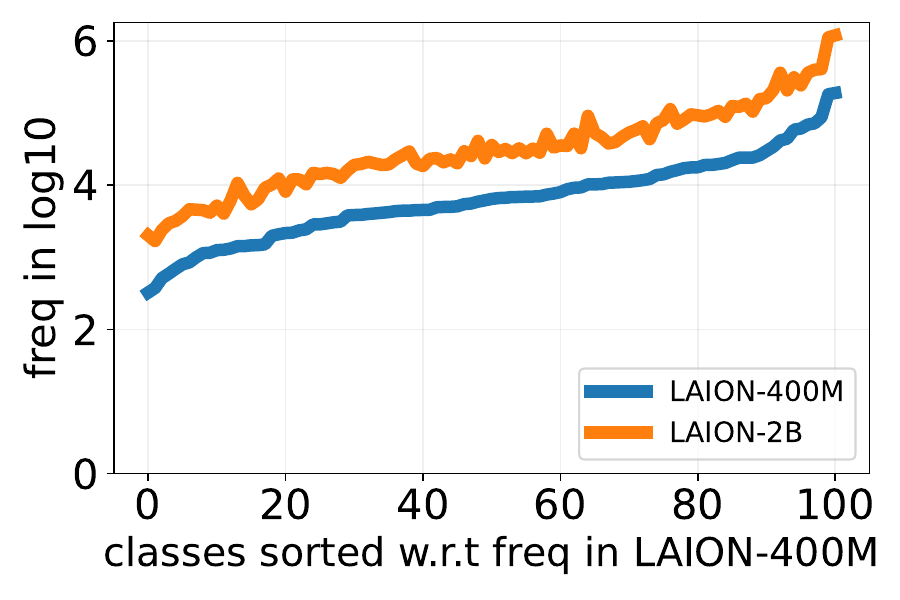} & \includegraphics[width=0.25\textwidth]{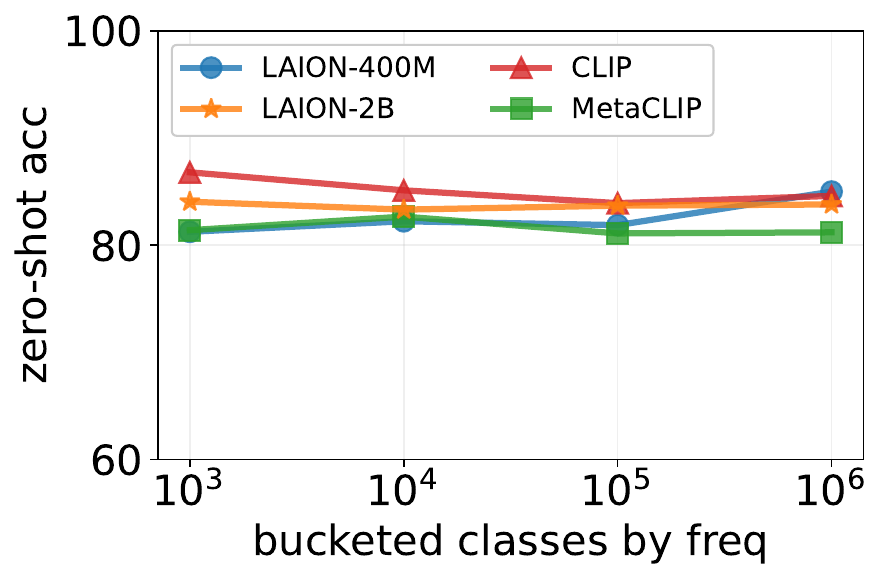} \\
\multicolumn{2}{c|}{(c) DTD} & \multicolumn{2}{c}{(d) Flowers} \\
\includegraphics[width=0.25\linewidth]{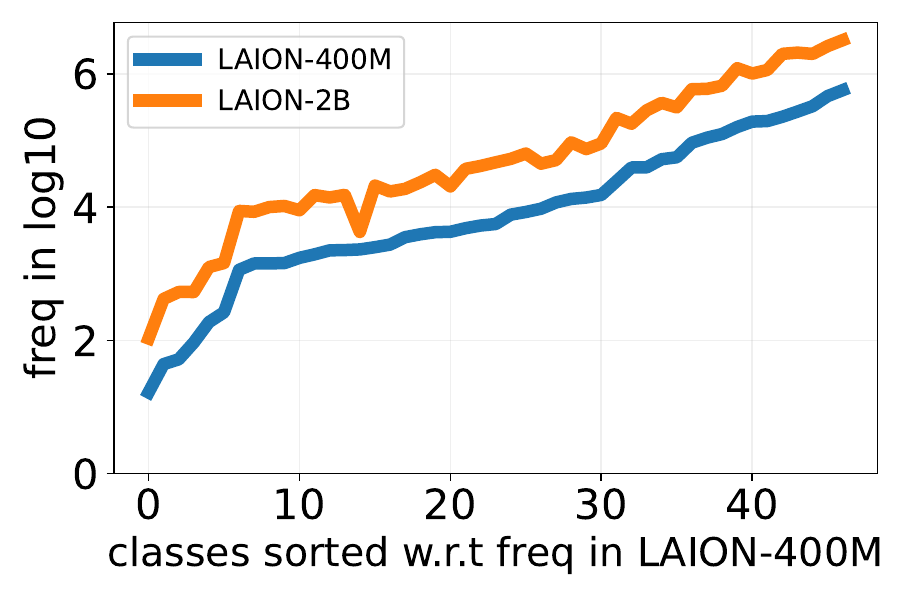} & \includegraphics[width=0.25\linewidth]{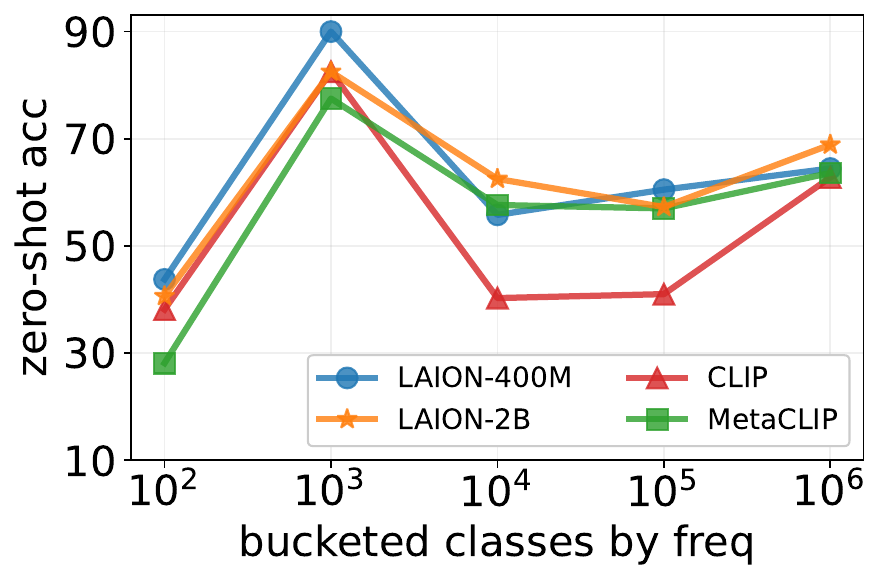} & \includegraphics[width=0.25\textwidth]{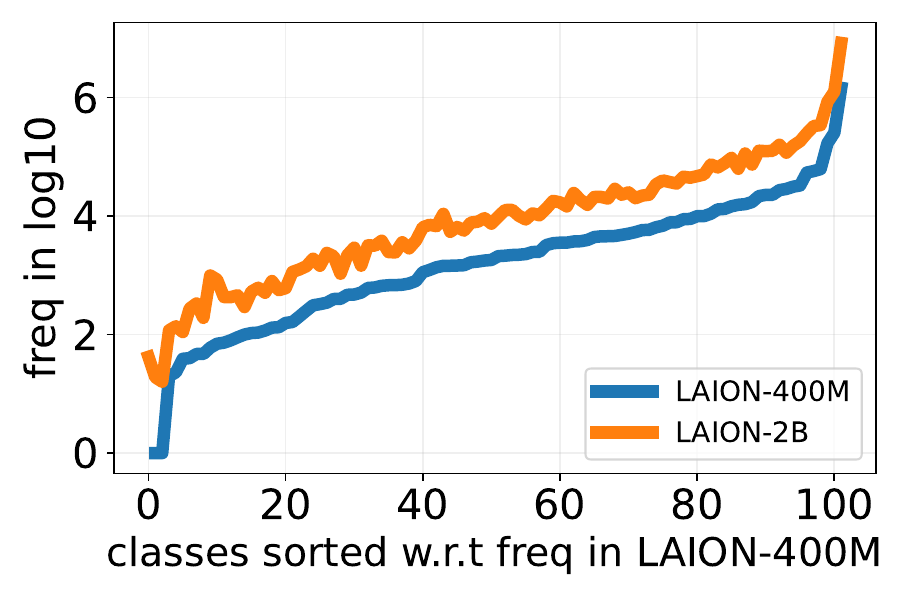} & \includegraphics[width=0.25\textwidth]{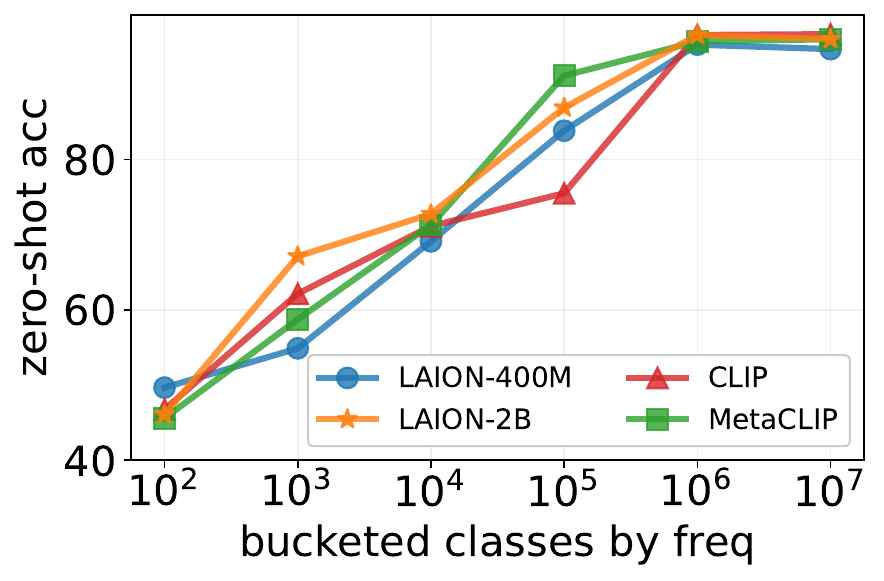} \\
\multicolumn{2}{c|}{(e) EuroSAT} & \multicolumn{2}{c}{(f) Pets} \\
\includegraphics[width=0.25\linewidth]{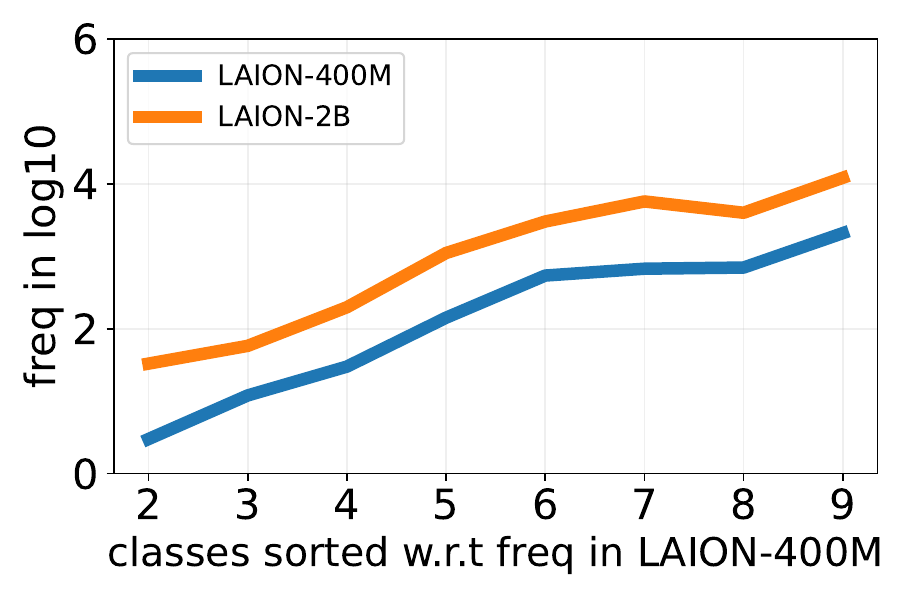} & \includegraphics[width=0.25\linewidth]{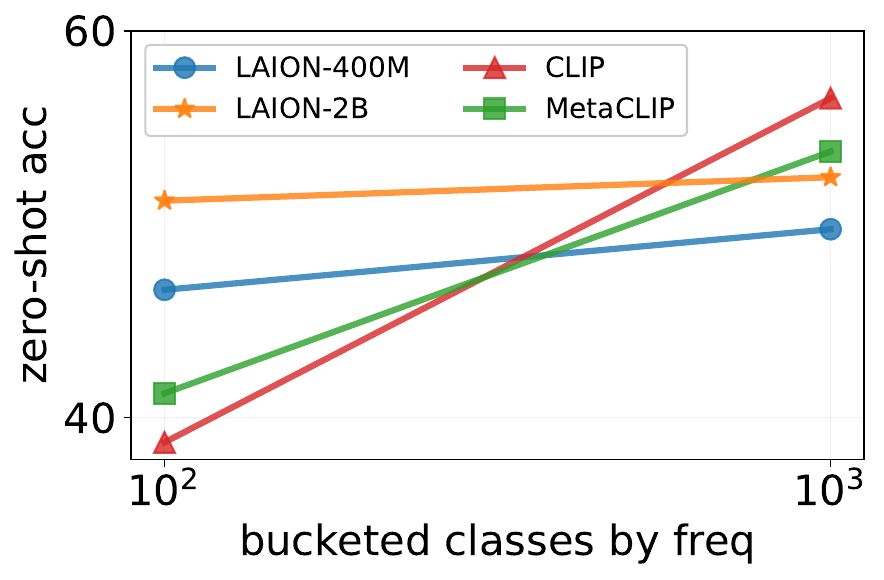} & \includegraphics[width=0.25\textwidth]{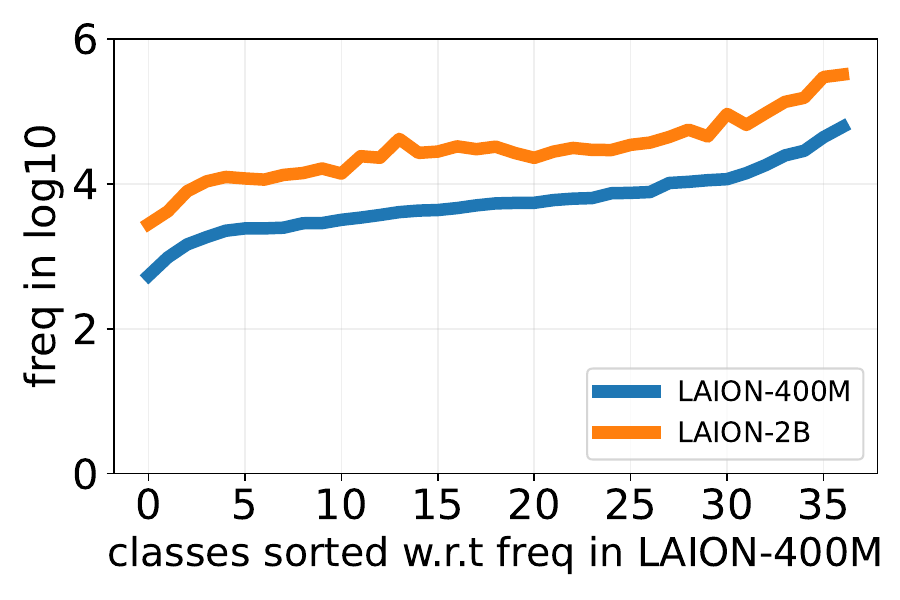} & \includegraphics[width=0.25\textwidth]{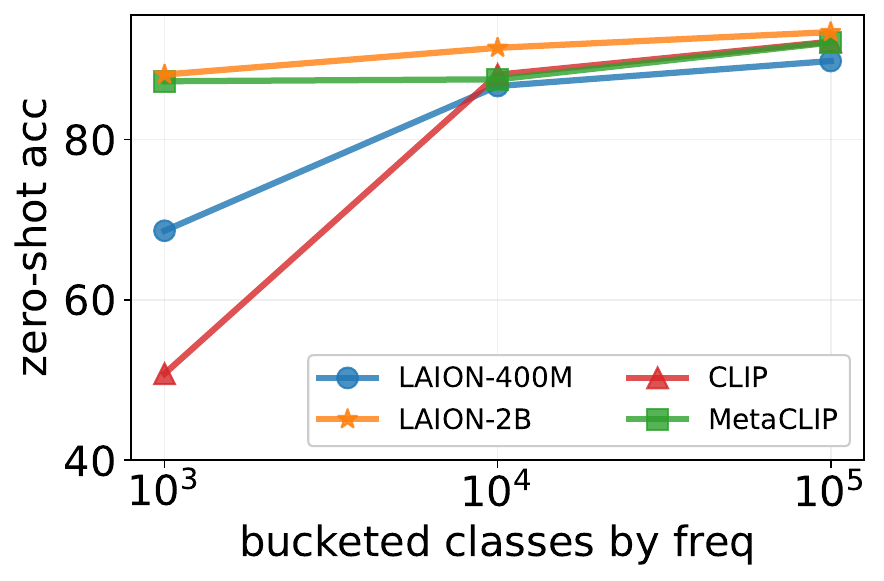} \\
\multicolumn{2}{c|}{(g) Cars} & \multicolumn{2}{c}{(h) Aircraft} \\
\includegraphics[width=0.25\linewidth]{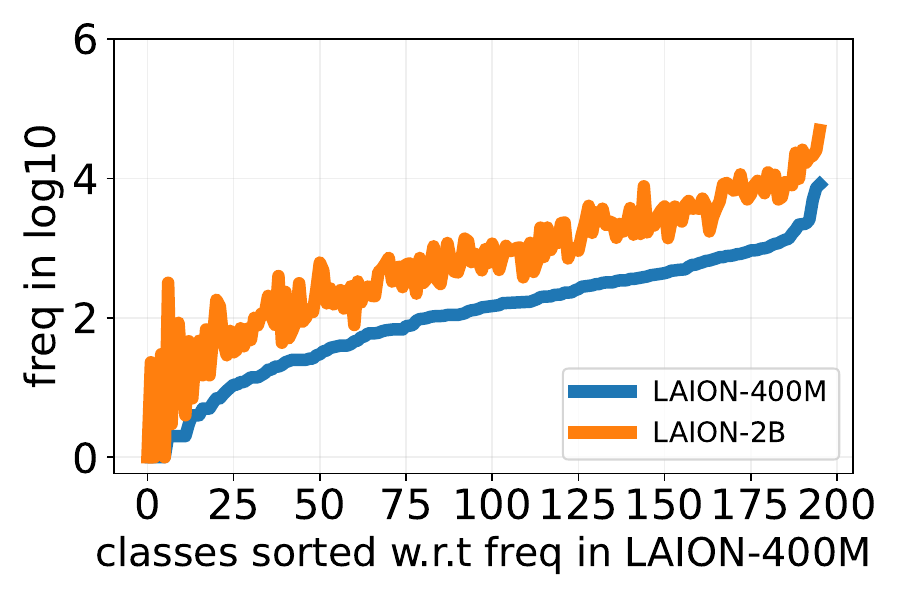} & \includegraphics[width=0.25\linewidth]{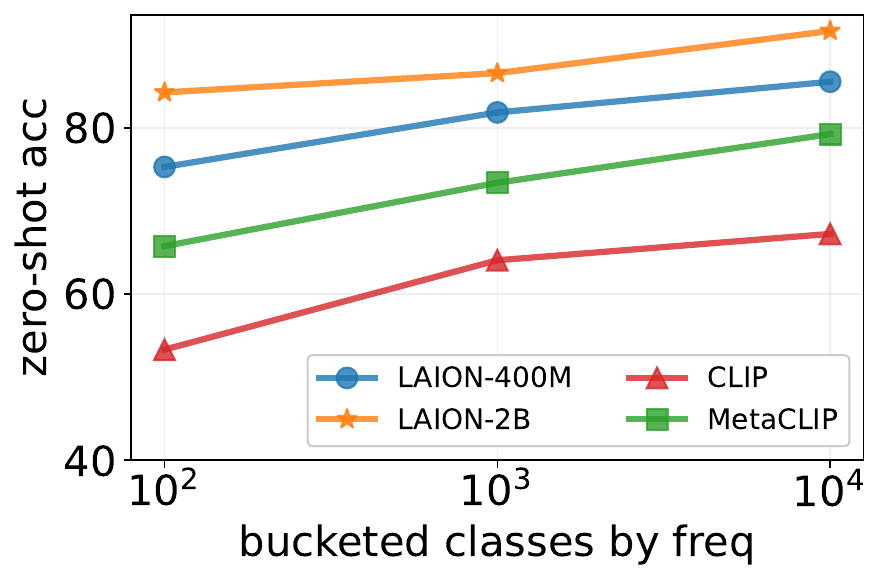} & \includegraphics[width=0.25\textwidth]{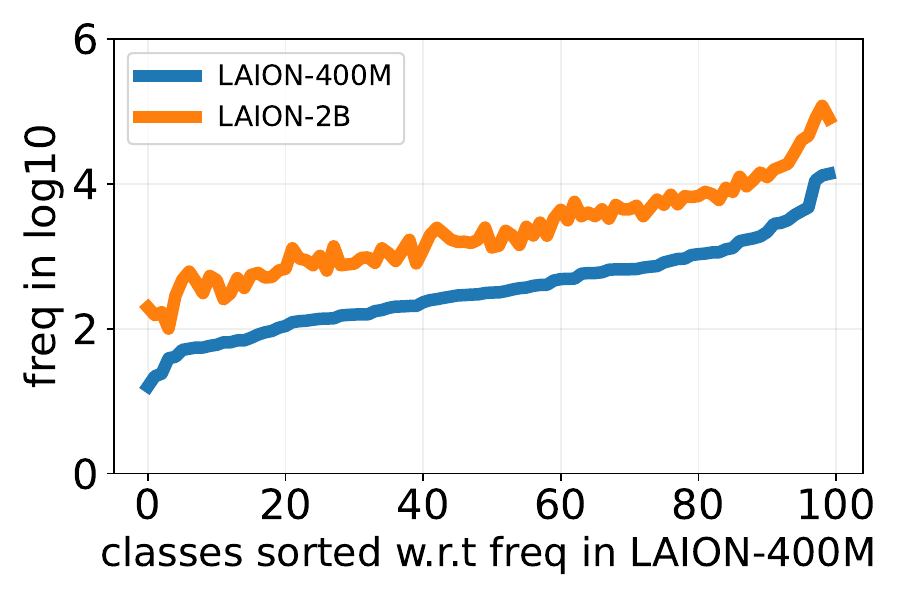} & \includegraphics[width=0.25\textwidth]{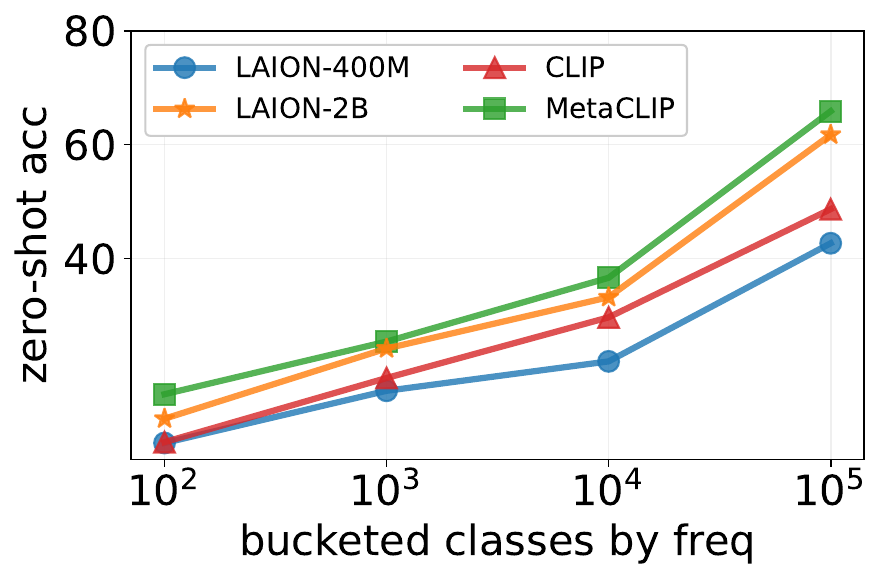} \\
\bottomrule
\end{tabular}
}
\label{tab:more_freq_acc}
\end{table*}

\section{Performance Breakdown of REAL}
\label{sec:sup_benefits_real}

In Table~\ref{tab:REAL_head_tail}, we show the improvement of REAL on the head and tail classes across nine benchmark datasets. We emphasize that REAL can significantly lift both head and tail accuracy on downstream tasks using the original pretraining data. 

{
\setlength{\tabcolsep}{0.85mm}
\begin{table*}[t]
\small
\centering
\caption{{\bf Breakdown improvements of REAL.}
REAL-Prompt and REAL-Linear (500 retrieved examples per concept) can significantly improve upon the baseline performance using the OpenAI templates~\cite{clip} on nine standard zero-shot recognition benchmark datasets.
We define the tail as the 20\% least frequent classes and the rest as the head for each dataset.
REAL significantly lifts both head and tail accuracies on these datasets. 
}
\label{tab:REAL_head_tail}
\vspace{-2mm}
\scalebox{0.85}{
\begin{tabular}{lllccccccccccccccccc|cc}
\toprule
& \multirow{2}{*}{\makecell{Method}}  &
  \multicolumn{2}{c}{ImageNet} &
  \multicolumn{2}{c}{Flowers} &
  \multicolumn{2}{c}{Cars} &
  \multicolumn{2}{c}{Aircraft} &
  \multicolumn{2}{c}{CUB} &
  \multicolumn{2}{c}{Pets} &
  \multicolumn{2}{c}{Food} &
  \multicolumn{2}{c}{DTD} &
  \multicolumn{2}{c}{EuroSAT} &
  \multicolumn{2}{c}{Avg} \\
\cmidrule(l){3-4} \cmidrule(l){5-6}
\cmidrule(l){7-8}   \cmidrule(l){9-10}
\cmidrule(l){11-12}   \cmidrule(l){13-14}
\cmidrule(l){15-16}   \cmidrule(l){17-18}
\cmidrule(l){19-20}   \cmidrule(l){21-22} 
 &
   &
  Head &
  Tail &
  Head &
  Tail &
  Head &
  Tail &
  Head &
  Tail &
  Head &
  Tail &
  Head &
  Tail &
  Head &
  Tail &
  Head &
  Tail &
  Head &
  Tail &
  Head &
  Tail \\
\midrule \multirow{5}{*}{\makecell{LAION\\400M}} & OpenAI templates             
&64.8  &55.2  & 70.0 & 50.6 &81.1  &72.9  &18.9  &8.4  & 69.1 & 40.1 & 87.4 & 83.5 & 80.4 & 82.5 &54.3 &55.2 &65.0 &23.9    &65.7  &52.5  \\
                          &REAL-Prompt &\makecell{65.4\\ \textcolor{Green}{+0.6}}  &\makecell{56.2\\ \textcolor{Green}{+1.0}}  & \makecell{76.8 \\ \textcolor{Green}{+6.8}}  & \makecell{58.8 \\ \textcolor{Green}{+8.2}}  &\makecell{85.2\\ \textcolor{Green}{+4.1}}  &\makecell{73.7\\ \textcolor{Green}{+0.8}}  &\makecell{20.8\\ \textcolor{Green}{+1.9}}  & \makecell{7.3\\ \textcolor{Red}{-1.1}}  & \makecell{69.3\\ \textcolor{Green}{+0.2}}  & \makecell{40.6\\ \textcolor{Green}{+0.5}} & \makecell{88.7\\ \textcolor{Green}{+1.3}} & \makecell{88.6\\ \textcolor{Green}{+5.1}} & \makecell{80.5\\ \textcolor{Green}{+0.1}} &  \makecell{82.3\\ \textcolor{Red}{-0.2}} &\makecell{59.3\\ \textcolor{Green}{+5.0}}  &\makecell{62.0\\ \textcolor{Green}{+6.8}} &\makecell{64.3\\ \textcolor{Red}{-0.7}} &\makecell{41.9\\ \textcolor{Green}{+18.0}}&\makecell{67.8\\ \textcolor{Green}{+2.1}}&\makecell{56.8\\ \textcolor{Green}{+4.3}}  \\
                          
                          &\makecell[l]{REAL-Linear\\(500)}          &\makecell{67.8\\ \textcolor{Green}{+2.9}}  &\makecell{58.9\\ \textcolor{Green}{+3.7}}  & \makecell{82.4\\ \textcolor{Green}{+12.4}} & \makecell{57.2\\ \textcolor{Green}{+6.6}} &\makecell{87.0\\ \textcolor{Green}{+5.9}}  &\makecell{73.2\\ \textcolor{Green}{+0.3}}  &\makecell{34.4\\ \textcolor{Green}{+15.5}}  &\makecell{10.0\\ \textcolor{Green}{+1.6}}  & \makecell{79.3\\ \textcolor{Green}{+10.2}} & \makecell{50.4\\ \textcolor{Green}{+10.3}} & \makecell{89.7\\ \textcolor{Green}{+1.3}}  & \makecell{87.7\\ \textcolor{Green}{+4.2}} & \makecell{80.8\\ \textcolor{Green}{+0.4}}  & \makecell{83.6\\ \textcolor{Green}{+1.1}}  &\makecell{60.8\\ \textcolor{Green}{+6.5}}  &\makecell{63.5\\ \textcolor{Green}{+8.3}}  &\makecell{69.9\\ \textcolor{Green}{+4.9}} &\makecell{19.2\\ \textcolor{Red}{-4.7}}
                          &\makecell{72.5\\ \textcolor{Green}{+6.5}} 
                          &\makecell{56.0\\ \textcolor{Green}{+3.5}}
                          
                          \\
\midrule 
\multirow{5}{*}{\makecell{LAION\\2B}} & OpenAI templates &68.0&61.0&75.6&50.5&87.0&82.5&27.9&11.5&73.0&49.0&90.5&90.6&82.0&85.1&58.0&55.2&54.0&38.1&68.6&58.4    \\
                          &REAL-Prompt &\makecell{68.2\\ \textcolor{Green}{+0.2}}	&\makecell{61.6\\ \textcolor{Green}{+0.6}}	&\makecell{79.4\\ \textcolor{Green}{+3.8}}	&\makecell{55.1\\ \textcolor{Green}{+4.6}}	&\makecell{89.2\\ \textcolor{Green}{+2.2}}	&\makecell{80.8\\ \textcolor{Red}{-1.7}}	&\makecell{29.3\\ \textcolor{Green}{+1.4}}	&\makecell{11.3\\ \textcolor{Red}{-0.2}}	&\makecell{72.8\\ \textcolor{Red}{-0.2}}	&\makecell{47.7\\ \textcolor{Red}{-1.3}}	&\makecell{91.5\\ \textcolor{Green}{+1.5}}	&\makecell{92.8\\ \textcolor{Green}{+2.2}}	&\makecell{82.1\\ \textcolor{Green}{+0.1}}	&\makecell{85.1\\ \textcolor{Gray}{+0.0}}	&\makecell{64.4\\ \textcolor{Green}{+6.4}}	&\makecell{63.5\\ \textcolor{Green}{+8.3}}	&\makecell{51.6\\ \textcolor{Red}{-2.4}}	&\makecell{58.7\\ \textcolor{Green}{+20.6}}
                          &\makecell{69.8\\ \textcolor{Green}{+1.2}}
                          &\makecell{61.8\\ \textcolor{Green}{+3.4}}\\
                          &\makecell[l]{REAL-Linear\\(500)}
                          &\makecell{69.8\\ \textcolor{Green}{+1.8}}
                          &\makecell{64.8\\ \textcolor{Green}{+3.8}}
                          &\makecell{84.1\\ \textcolor{Green}{+8.5}}
                          &\makecell{66.9\\ \textcolor{Green}{+16.4}}
                          &\makecell{90.0\\ \textcolor{Green}{+3.0}}
                          &\makecell{82.3\\ \textcolor{red}{-0.2}}
                          &\makecell{45.4\\ \textcolor{Green}{+17.6}}
                          &\makecell{25.5\\ \textcolor{Green}{+14.0}}
                          &\makecell{82.4\\ \textcolor{Green}{+9.4}}
                          &\makecell{62.2\\ \textcolor{Green}{+13.2}}
                          &\makecell{91.5\\ \textcolor{Green}{+1.0}}  
                          &\makecell{92.6\\ \textcolor{Green}{+2.0}}
                          &\makecell{82.3\\ \textcolor{Green}{+0.3}}
                          &\makecell{86.2\\ \textcolor{Green}{+1.1}}
                          &\makecell{64.5\\ \textcolor{Green}{+6.5}}
                          &\makecell{70.0\\ \textcolor{Green}{+14.8}}
                          &\makecell{76.0\\ \textcolor{Green}{+22.0}}  &\makecell{22.0\\ \textcolor{Red}{-16.1}}
                          &\makecell{76.2\\ \textcolor{Green}{+7.6}}
                          &\makecell{63.6\\ \textcolor{Green}{+5.2}}
                          \\
\bottomrule
\end{tabular}
}
\vspace{+1mm}
\end{table*}
}

\section{Further Details for REAL}
\label{sec:sup_real}
{\bf Synonym filtering in REAL-Prompt.} We use OpenCLIP's text encoder to filter out ChatGPT-generated synonyms that might be confused with other downstream concepts. Specifically, we retain only those synonyms that have the highest cosine similiarity scores with their original class names (not with another downstream concept). This filtering step is critical to REAL-Prompt's performance as shown in Table~\ref{tab:t2t-filter}.

{\bf Linear probing in REAL-Linear.} We follow previous work~\cite{lin2023multimodality, wiseft} and adopt the same procedure and hyperparameters to learn a robust cross-modal classifier. Specifically, we initialize the weights of the cross-modal linear classifier using averaged text features constructed using the most frequent synonyms and OpenAI templates~\cite{clip}. Next, we stick to the reported~\cite{lin2023multimodality} learning rate of 1e-4 with a cosine annealing schedule, weight decay of 1e-2, batch size of 32, and training epochs of 10. Finally, we average the learned cross-modal classifier weights with the zero-shot classifier weights (as shown in Figure~\ref{fig:REAL-linear}). 
We apply the same set of hyperparameters for all datasets and model architectures. We will release our code and retrieved data for reproducibility.

\section{More Ablations of REAL-Linear}
\label{sec:sup_ablations}
In this section, we show that synonym-based retrieval and cross-modal adaptation are crucial for the performance of REAL-Linear. We also explain the lower performance of REAL-Linear on Stanford Cars dataset. Lastly, we ablate the retrieval sizes v.s. zero-shot accuracies.

{\bf Synonyms help retrieve diverse data.} It is crucial to retrieve images whose captions contain any of the concept synonyms instead of just the name predefined by the downstream task. Table~\ref{tab:synonyms_diversity} shows that using all synonyms can retrieve more diverse images for a performance boost of 4\% from 64.2\% to 68.2\% when averaged across eight benchmark datasets, surpassing REACT Locked-Text's 65.5\%. In addition, Table~\ref{tab:prime_vs_real} shows that REAL-Linear outperforms another retrieval-augmented method Neural Priming~\cite{wallingford2023neural}, which does not consider concept synonyms for retrieval. For a fair comparison, we follow \cite{wallingford2023neural} to use 100 retrieved images per class because they do not release the models and hyperparameters.

{
\setlength{\tabcolsep}{0.68em}
\begin{table*}[t]
\vspace{-3mm}
\centering
\small
\caption{\textbf{Using concept synonyms helps retrieve more diverse pretraining images.} 
Retrieving images whose captions contain any of the concept synonyms (instead of just the name predefined by the downstream task) can improve the performance of REAL-Linear (using 500 images per class). We attach the performance of REACT Locked-Text (using 10K images per class) for reference.
}
\vspace{-3mm}
\label{tab:synonyms_diversity}
\scalebox{0.9}{
\begin{tabular}{lccccccccc|c}
\toprule
Method & Images per class &  ImageNet & Flowers & Cars & Aircraft & Pets & Food & DTD & EuroSAT & {\bf Avg} \\ 
\midrule
REACT Locked-Text & 10K & 65.7 & 73.1 & {\bf 88.5} & 24.5 & 89.2 & 81.8 & 49.8 & 51.1 & 65.5 \\
REAL-Linear (without synonyms) & 500 & 64.6 & 72.6 & 71.1 & 27.9  & 88.6 & 81.9 & 56.3 & 50.8 & 64.2 \\
REAL-Linear (with synonyms) & 500 & {\bf 65.9} & {\bf 78.8} & 84.4 & {\bf 29.6} & {\bf 89.5} & {\bf 81.4} & {\bf 61.5} & {\bf 51.5} & {\bf 67.8} \\ 
\bottomrule
\end{tabular}
}
\end{table*}
}

{\bf Learning robust cross-modal classifiers.} We show that cross-modal adaptation~\cite{lin2023multimodality}, which uses both text and image features to learn a linear classifier, is more robust against the distribution shifts between retrieved (pretraining) data and target domains. Concretely, Table~\ref{tab:ablation_finetune} shows that performing naive linear probing using only retrieved images achieves lower accuracy by 6.4\%  averaged across all benchmark datasets, sometimes underperforms the zero-shot classifier~\cite{clip} constructed using OpenAI prompt templates~\cite{clip}. 
This shows that using both images and texts can effectively reduce overfitting to retrieved pretraining data. 

{\bf Remarks on REAL-Linear's performance on the Stanford Cars dataset.} Table~\ref{tab:comparison_sota} shows that the performance of our REAL-Linear on the Cars dataset~\cite{cars} is 4\% lower than that of REACT~\cite{liu2023learning}, despite that we relax the string matching criteria (by matching partial names) to retrieve more relevant images. We attribute the performance gap to the limited images retrieved from LAION-400M~\cite{laion400m}, owing to the fine-grained nature of the class names, e.g. ``Audi S6 Sedan 2011''. Supporting evidence is shown in Table~\ref{tab:synonyms_diversity}, where using synonyms for retrieval increases the accuracy of Cars from 71.1\% to 84.4\%. This suggests future work on better retrieval methods for datasets with specific brand names. 

{\bf Remarks on retrieval sizes.} Retrieving more pretraining examples generally helps REAL-Linear achieve higher accuracies for zero-shot recognition, as shown in Table~\ref{tab:ablation_shots}.
Yet, increasing the retrieval size from 100 to 500 per concept only improves accuracy by 0.9\% (averaged over nine benchmarks). As such, we adopt 500 for our major experiments in this paper.

{
\setlength{\tabcolsep}{1.5em}
\begin{table*}[t]
\small
\centering
\caption{\textbf{REAL-Linear outperforms Neural Priming}. We compare REAL-Linear with Neural Priming~\cite{wallingford2023neural} using the ViT-B/16 model pre-trained on LAION-2B~\cite{laion5b}. REAL-Linear consistently outperforms Neural Priming on all their reported benchmarks, presumably because Neural Priming does not consider synonyms for retrieval.
}
\vspace{-3mm}
\label{tab:prime_vs_real}
\scalebox{0.9}{
\begin{tabular}{lccccccc|c}
\toprule
Method & Images per class & \multicolumn{1}{c}{ImageNet} & Cars & Flowers & Aircraft & Food & Pets & {\bf Avg} \\ 
 \midrule
Neural Priming & 100 & 70.8 & 89.3 & 79.8 & 33.0 & 86.7 & 91.9 & 75.3 \\
REAL-Linear & 100 & {\bf 71.9} & {\bf 90.3} & {\bf 81.9} & {\bf 38.7} & {\bf 86.7} &  {\bf 92.2} & {\bf 77.0} \\ 
\bottomrule
\end{tabular}
}
\end{table*}
}

{
\setlength{\tabcolsep}{0.75em}
\begin{table*}[t]
\small
\caption{
\textbf{Cross-modal adaptation improves the robustness of REAL-Linear.} We highlight that using both images and texts during training can help address the distribution shifts between pretraining data and target domains. Concretely, we adopt cross-modal WiSE-FT~\cite{lin2023multimodality, wiseft}, which first learns a cross-modal linear classifier using both retrieved image features and text features constructed using the most frequent concept synonyms and
OpenAI templates~\cite{clip}. This cross-modal classifier is then ensembled with a zero-shot classifier whose weights are text features of the most frequent synonyms averaged across OpenAI prompt templates. We show that this cross-modal strategy is much more robust against vanilla image-only linear probing that uses only retrieved image features, which overfits to retrieved data and sometimes underperforms the zero-shot classifier.
} 

\vspace{-3mm}
\label{tab:ablation_finetune}
\scalebox{0.82}{
\begin{tabular}{l l l l l l l l l l | l}
    \toprule
    Method & ImageNet & Flowers & Cars & Aircraft & CUB & Pets & Food & DTD & EuroSAT & {\bf Avg} \\
    \midrule

    OpenAI templates~\cite{clip} & 
    62.9  & 
    68.0  &
    79.2  & 
    16.7  &
    63.8  &
    86.7  &
    80.9  &
    54.5  &
    51.5  &
    62.7\\

    REAL-Linear (image-only) &
     62.1$^{\textcolor{Red}{-0.8}}$ &
     78.0$^{\textcolor{Green}{+10.0}}$&
     77.5$^{\textcolor{Red}{-1.7}}$& 
     33.1$^{\textcolor{Green}{+16.4}}$&
     73.1$^{\textcolor{Green}{+9.3}}$& 
     86.1$^{\textcolor{Red}{-0.6}}$& 
     79.5$^{\textcolor{Red}{-1.4}}$&
     53.8$^{\textcolor{Red}{-0.7}}$&
     15.6$^{\textcolor{Red}{-35.9}}$&
     62.4$^{\textcolor{Red}{-0.3}}$\\

    REAL-Linear (cross-modal) &
     65.9$^{\textcolor{Green}{+3.0}}$ &
     78.8$^{\textcolor{Green}{+10.8}}$&
     84.1$^{\textcolor{Green}{+7.9}}$& 
     29.6$^{\textcolor{Green}{+12.9}}$& 
     74.0$^{\textcolor{Green}{+10.2}}$& 
     89.5$^{\textcolor{Green}{+2.8}}$&
     81.4$^{\textcolor{Green}{+0.5}}$&
     61.5$^{\textcolor{Green}{+6.2}}$&
     51.5$^{\textcolor{Green}{+0.0}}$ &
     67.8$^{\textcolor{Green}{+6.1}}$ 
     \\

    \bottomrule
\end{tabular}}
\end{table*}
}

{
\setlength{\tabcolsep}{1em}
\begin{table*}[t]
\small
\centering
\caption{{\bf Zero-shot accuracy vs. retrieval size.} 
We conducted an ablation study the impact of  retrieval size for REAL-Linear, and for comparison, we included results using OpenAI templates and our REAL-Prompt. Notably, even with a smaller retrieval size of 100 images per concept, we achieve strong performance  (only ~1\% lower on avg.), though our best results come with a retrieval size of 500 images per concept.}
\vspace{-3mm}
\scalebox{0.90}{
\begin{tabular}{l c c c c c c c c c | l}
    \toprule
    Number of shots & ImageNet & Flowers & Cars & Aircraft &CUB & Pets & Food & DTD & EuroSAT & {\bf Avg} \\
    \midrule

    OpenAI templates~\cite{clip}  & 
    62.9  & 
    68.0  &
    79.2  & 
    16.7  &
    63.8  &
    86.7  &
    80.9  &
    54.5  &
    51.5  & 
    62.7 \\

    REAL-Prompt  &
    63.6 &
    76.6 &
    82.7 & 
    18.0 &
    64.0 &
    88.8 & 
    81.0 &
    59.9 &
    {57.5} &
    65.8$^{\textcolor{Green}{+3.1}}$ \\
    
    REAL-Linear (100) & 
    65.3 & 
    77.8 &
    84.0 & 
    25.1 &
    72.4 &
    89.3 & 
    81.0 & 
    60.4 & 
    53.3 & 
    67.6$^{\textcolor{Green}{+4.9}}$ \\
    
    REAL-Linear (500) &
    {65.9} &
    {78.8} &
    {84.4} &
    {29.6} &
    {74.0} &
    {89.5} &
    {81.4} &
    {61.5} &
    51.5 &
    {68.5$^{\textcolor{Green}{+5.8}}$} \\
    
    \bottomrule
\end{tabular}}
\vspace{-3mm}
\label{tab:ablation_shots}
\end{table*}
}

{
\setlength{\tabcolsep}{0.7em}
\begin{table*}[ht]

\small
\caption{{\bf REAL-Linear generalizes across different pretraining datasets and architectures.} 
REAL-Linear consistently achieves performance gains with three OpenCLIP architectures (ViT B/32, B/16, and L/14) and pretraining datasets (LAION 400M and 2B). For reference, we attach the performance REACT reported on these benchmarks. Notably, our REAL-Linear (500 examples per class) even outperforms REACT Gated-Image (10K examples per class) when both use a larger visual encoder ViT-L/14. 
We highlight the \textbf{best accuracy} in bold and underline the \underline{second best} numbers for ImageNet.
}
\vspace{-3mm}
\label{tab:diff_archs}
\scalebox{0.88}{
\begin{tabular}{cclccccccccc|l}
\toprule
 & Arch & Method & ImageNet & Flowers & Cars & Aircraft & CUB & Pets & Food & DTD & EuroSAT & Avg\\
 \midrule
\multirow{10}{*}{\makecell{LAION\\400M}} & \multirow{4}{*}{\makecell{ViT\\B/32}} & OpenAI templates~\cite{clip} & 62.9 & 68.0 & 79.2 & 16.7 & 63.8 & 86.7 & 80.9 & 56.1 & 51.5 & 62.6\\
 & & REACT Locked-Text (10K) & \underline{65.7} & 73.1 & 88.5 & 24.5 &-- & 89.2 & 81.8 & 49.8 & 51.1 &--\\
 &  & REACT Gated-Image (10K) & 64.2 & 72.3 & 88.1 & 24.8 &-- & 89.5 & 83.0 & 51.4 & 45.4 &--\\
 &  & \cellcolor{gray!15}REAL-Linear (500) & \cellcolor{gray!15}\textbf{65.9} & \cellcolor{gray!15}78.8 & \cellcolor{gray!15}84.1 & \cellcolor{gray!15}29.6 & \cellcolor{gray!15}74.0 & \cellcolor{gray!15}89.5 & \cellcolor{gray!15}81.4 & \cellcolor{gray!15}61.5 & \cellcolor{gray!15}51.5 &\cellcolor{gray!15}68.5\\ \cline{2-13} 
 
 & \multirow{4}{*}{\makecell{ViT\\B/16}} & OpenAI templates~\cite{clip} & 67.0 & 69.2 & 83.6 & 17.7 & 67.2 & 89.3 & 86.2 & 51.3 & 50.3 & 64.6\\
 &  & REACT Locked-Text (10K) & 69.9 &-- &-- &-- &-- &-- &-- &-- &-- &--\\
 &  & REACT Gated-Image (10K) & \textbf{70.5} &-- &-- &-- &-- &-- &-- &-- &-- &--\\
 &  & \cellcolor{gray!15}REAL-Linear (500) & \cellcolor{gray!15}\underline{69.6} & \cellcolor{gray!15}80.6 & \cellcolor{gray!15}86.5 & \cellcolor{gray!15}31.5 & \cellcolor{gray!15}79.1 & \cellcolor{gray!15}91.3 & \cellcolor{gray!15}86.4 & \cellcolor{gray!15}61.4 & \cellcolor{gray!15}51.9 &\cellcolor{gray!15}71.0\\ \cline{2-13} 
 
 & \multirow{2}{*}{\makecell{ViT\\L/14}} & OpenAI templates~\cite{clip} & \underline{72.7} & 75.4 & 89.5 & 24.9 & 76.4 & 91.8 & 90.0 & 60.2 & 62.3 & 71.5\\
 &  & \cellcolor{gray!15}REAL-Linear (500) & \cellcolor{gray!15}\textbf{74.4} & \cellcolor{gray!15}85.4 & \cellcolor{gray!15}91.0 & \cellcolor{gray!15}40.2 & \cellcolor{gray!15}84.8 & \cellcolor{gray!15}93.4 & \cellcolor{gray!15}90.3 & \cellcolor{gray!15}66.5 & \cellcolor{gray!15}59.8 &\cellcolor{gray!15}76.2\\
 \midrule
\multirow{10}{*}{\makecell{LAION\\2B}} & \multirow{4}{*}{\makecell{ViT\\B/32}} & OpenAI templates~\cite{clip} & 66.6 & 71.8 & 86.0 & 24.5 & 68.5 & 90.6 & 82.7 & 56.1 & 48.0 &66.1\\
 &  & REACT Locked-Text (10K) & 67.5 &-- &-- &-- &-- &-- &-- &-- &-- &--\\
 &  & REACT Gated-Image (10K) & \textbf{69.6} &-- &-- &-- &-- &-- &-- &-- &-- &--\\
 &  & \cellcolor{gray!15}REAL-Linear (500) & \cellcolor{gray!15}\underline{68.8} & \cellcolor{gray!15}80.6 & \cellcolor{gray!15}88.4 & \cellcolor{gray!15}41.3 &\cellcolor{gray!15}78.5 & \cellcolor{gray!15}91.7 & \cellcolor{gray!15}83.1 & \cellcolor{gray!15}65.6 & \cellcolor{gray!15}51.9 &\cellcolor{gray!15}72.2\\ \cline{2-13} 
 
 & \multirow{2}{*}{\makecell{ViT\\B/16}} & OpenAI templates~\cite{clip} & \underline{70.2} & 71.4 & 88.2 & 26.9 & 72.7 & 90.5 & 86.5 & 56.3 & 53.4&68.5 \\
 &  & \cellcolor{gray!15}REAL-Linear (500) & \cellcolor{gray!15}\textbf{72.4} & \cellcolor{gray!15}83.4 & \cellcolor{gray!15}90.3 & \cellcolor{gray!15}45.6 & \cellcolor{gray!15}83.6 & \cellcolor{gray!15}92.2 & \cellcolor{gray!15}87.1 & \cellcolor{gray!15}66.0 & \cellcolor{gray!15}46.9 &\cellcolor{gray!15}74.2\\ \cline{2-13} 
 & \multirow{4}{*}{\makecell{ViT\\L/14}} & OpenAI templates~\cite{clip} & 75.3 & 75.2 & 91.9 & 36.6 & 78.5 & 93.2 & 91.0 & 62.8 & 64.6 & 74.3\\
 & & REACT Gated-Image (10K) & \underline{76.5} &-- &-- &-- &-- &-- &-- &-- &-- &--\\
 &  & \cellcolor{gray!15}REAL-Linear (500) & \cellcolor{gray!15}\textbf{76.9} & \cellcolor{gray!15}86.5 & \cellcolor{gray!15}92.6 & \cellcolor{gray!15}55.3 & \cellcolor{gray!15}87.5 & \cellcolor{gray!15}94.7 & \cellcolor{gray!15}91.2 & \cellcolor{gray!15}69.4 & \cellcolor{gray!15}57.9&\cellcolor{gray!15}79.1\\
 \bottomrule
\end{tabular}
}
\end{table*}
}

\section{Generalization Performance of REAL}
\label{sec:sup_generalization}
In this section, we show that REAL generalizes across model architectures, datasets, and prompt templates. 

\begin{figure}[t]
\centering
\small
{\scriptsize LAION \hspace{7mm} EuroSAT \hspace{15mm} LAION \hspace{7mm} EuroSAT } \\
\scalebox{0.85}{
\begin{tabular}{cc}
\includegraphics[width=0.25\linewidth]{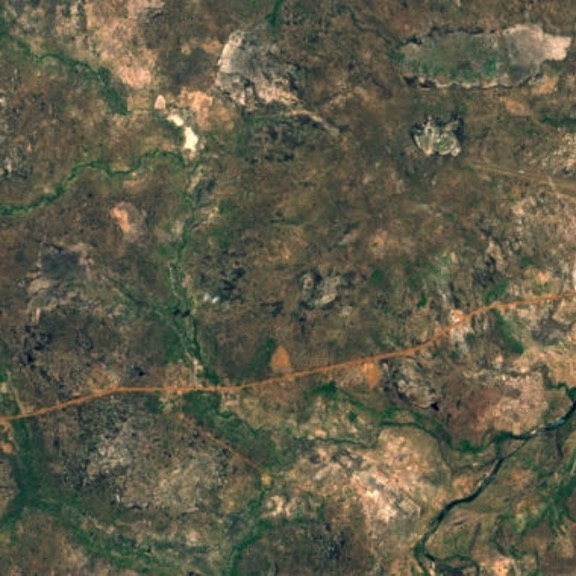}
\includegraphics[width=0.25\linewidth]{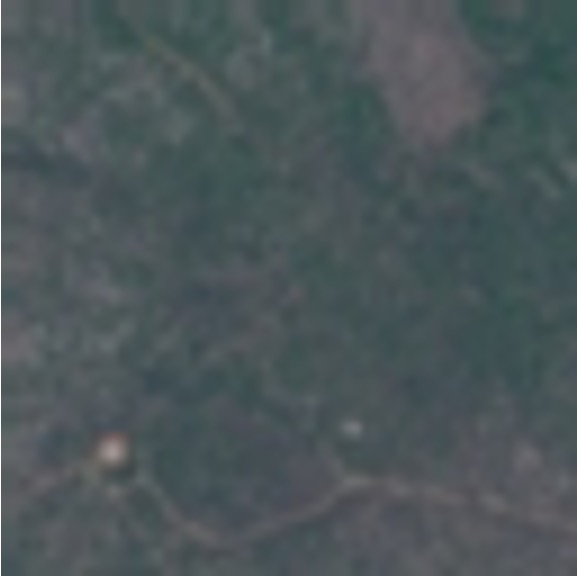} &
\includegraphics[width=0.25\linewidth]{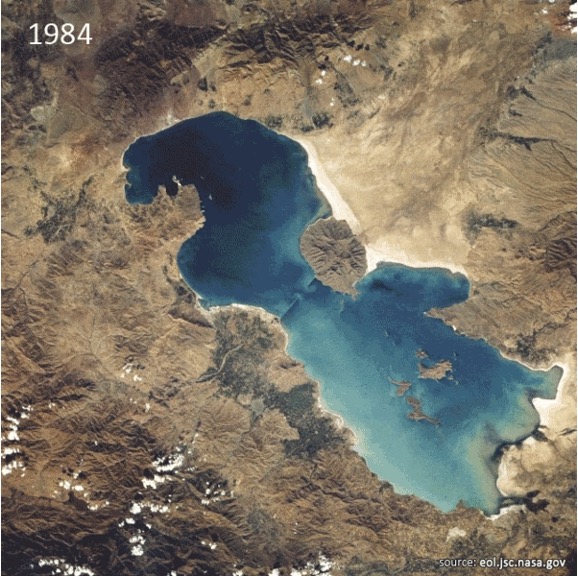}
\includegraphics[width=0.25\linewidth]{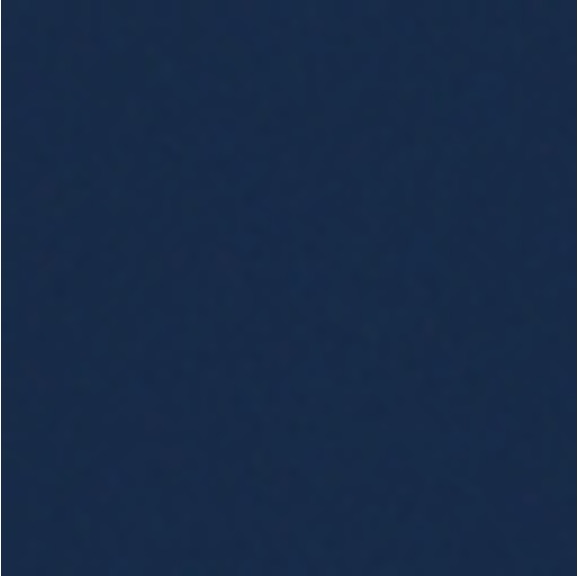}
\\
\makecell{(a) {\tt Herbaceous Vegetation}} & \makecell{(b) {\tt Sea Lake}}\\
\end{tabular}
}
\vspace{-3mm}
\caption{\small {\bf Large distribution shifts between LAION and EuroSAT.} We compare two randomly sampled satellite images from LAION and EuroSAT, for the class {\tt Herbaceous Vegetation} and {\tt Sea Lake}, respectively. Images from LAION present higher resolution and more distinct features while the EuroSAT images are blurry and lack informative features.   
}
\vspace{-3mm}
\label{fig:eurosat}
\end{figure}

{\bf Generalizing across architectures and datasets.} Table~\ref{tab:diff_archs} shows that REAL-Linear consistently improves the zero-shot performance of OpenCLIP across different ViT architectures (B/32, B/16, and L/14) and LAION datasets (400M and 2B).
Yet, both REAL-Linear and REACT~\cite{liu2023learning} fail to improve on the EuroSAT dataset~\cite{helber2019eurosat}, presumably because satellite imagery is very rare in LAION (e.g., we can retrieve at most one image for {\tt Annual Crop} and {\tt Herbaceous Vegetation}). In addition, the few retrieved satellite images in LAION are drastically different from EuroSAT testset images due to sensor shifts, as shown in Figure~\ref{fig:eurosat}.

{\bf Generalizing across prompt templates.} Table~\ref{tab:HFL_prompting} shows that REAL-Prompt is effective regardless of the prompt templates (OpenAI~\cite{clip}, DCLIP~\cite{menon2022visual}, and CuPL~\cite{CUPL}). 

{
\setlength{\tabcolsep}{0.75em}
\begin{table*}[t!]
    \centering
    \small
    \caption{\textbf{REAL-Prompt generalizes across prompt templates.} We show that REAL-Prompt (using the most frequent synonyms) can improve upon both OpenAI prompt templates~\cite{clip} and LLM-enriched templates such as DCLIP~\cite{menon2022visual} and CuPL~\cite{CUPL}.
    }
    \vspace{-3mm}
    \scalebox{.95}{
    \begin{tabular}{cclllllllll}
    \toprule
         & Arch & Method &  ImageNet
         &  Flowers 
         &  Cars 
         &  Aircraft 
         &  CUB 
         &  Pets 
         &  Food 
         &  DTD \\
         \hline
\multirow{18}{*}{\makecell{LAION-400M}} & \multirow{6}{*}{\makecell{ViT-B/32}} &    OpenAI templates ~\cite{clip}& 62.9 & 68.0 & 79.2 & 16.7&63.8 & 86.7 & 80.9 & 54.5 \\ 
&& \cellcolor{gray!15}\quad + \rpl     & \cellcolor{gray!15}63.6                 & \cellcolor{gray!15}76.6                 & \cellcolor{gray!15}82.7                 & \cellcolor{gray!15}18.0                 & \cellcolor{gray!15}64.0                 & \cellcolor{gray!15}88.8                 & \cellcolor{gray!15}81.0                 & \cellcolor{gray!15}59.9   \\             
&& DCLIP ~\cite{menon2022visual} & 62.1 &  --                    &       --               &    --                  & 64.5 & 84.6 & 80.1 & 51.4 \\   
&& \cellcolor{gray!15}\quad + \rpl& \cellcolor{gray!15}62.9                 & \cellcolor{gray!15}--                    &\cellcolor{gray!15}--                     &\cellcolor{gray!15}--                      & \cellcolor{gray!15}64.7                 & \cellcolor{gray!15}88.1                 & \cellcolor{gray!15}80.0                 & \cellcolor{gray!15}55.5   \\              
&& CuPL ~\cite{CUPL}    & 63.7                & 65.8                 & 80.0                 & 17.8                 &--                     & 87.4                 & 79.5                 & 59.1                                                            \\
&& \cellcolor{gray!15}\quad + \rpl& \cellcolor{gray!15}64.2                 & \cellcolor{gray!15}72.3                 & \cellcolor{gray!15}81.7                 & \cellcolor{gray!15}18.3                 & \cellcolor{gray!15} --                    & \cellcolor{gray!15}88.0                 & \cellcolor{gray!15}79.5                 & \cellcolor{gray!15}59.3                                                             \\
\cline{2-11}

& \multirow{6}{*}{\makecell{ViT-B/16}} &OpenAI templates ~\cite{clip} & 67.0 & 69.2 & 83.6 & 17.7 & 67.2 & 89.3 & 86.2 & 51.0\\
&&  \cellcolor{gray!15}\quad + \rpl      & \cellcolor{gray!15}67.6                 & \cellcolor{gray!15}77.1                 & \cellcolor{gray!15}84.4               & \cellcolor{gray!15}19.5                 & \cellcolor{gray!15}67.3                 & \cellcolor{gray!15}91.0                 & \cellcolor{gray!15}86.3                 & \cellcolor{gray!15}58.1  \\              
&&DCLIP ~\cite{menon2022visual} &65.8&--&--&--&68.6&86.2&85.2&51.1\\ 

&&\cellcolor{gray!15}\quad+ \rpl& \cellcolor{gray!15}66.2                 &\cellcolor{gray!15}--                      &\cellcolor{gray!15}--                     &\cellcolor{gray!15}--                     & \cellcolor{gray!15}68.6                 & \cellcolor{gray!15}89.8                 & \cellcolor{gray!15}85.2                 & \cellcolor{gray!15}57.1  \\               
&&CuPL ~\cite{CUPL} & 67.8&67.9&83.4&18.6&--&89.7&85.2&57.9 \\
&& \cellcolor{gray!15}\quad+ \rpl& \cellcolor{gray!15}68.1                 & \cellcolor{gray!15}73.1                 & \cellcolor{gray!15}84.0                 & \cellcolor{gray!15}18.8                 &  \cellcolor{gray!15}--                   & \cellcolor{gray!15}90.5                 & \cellcolor{gray!15}\cellcolor{gray!15}85.2                 & \cellcolor{gray!15}59.8                                                            \\
\cline{2-11}

& \multirow{6}{*}{\makecell{ViT-L/14}} & OpenAI templates ~\cite{clip}& 72.7 & 75.4 & 89.5 & 24.9 & 76.4 & 91.8 & 90.0 & 60.2 \\ 
&&\cellcolor{gray!15}\quad + \rpl  & \cellcolor{gray!15}72.9                 & \cellcolor{gray!15}82.9                 & \cellcolor{gray!15}89.9                 & \cellcolor{gray!15}26.0                 & \cellcolor{gray!15}76.4                 & \cellcolor{gray!15}93.3                 & \cellcolor{gray!15}90.2                 & \cellcolor{gray!15}63.6  \\               
&&DCLIP ~\cite{menon2022visual} &71.8&--&--&--&77.2&89.2&89.3&57.7 \\

&&\cellcolor{gray!15}\quad+ \rpl& \cellcolor{gray!15}72.3                 &  \cellcolor{gray!15}--                    &  \cellcolor{gray!15}--                    &   \cellcolor{gray!15}--                   & \cellcolor{gray!15}77.3                 & \cellcolor{gray!15}92.1                 & \cellcolor{gray!15}89.4                 & \cellcolor{gray!15}60.5   \\               
&&CuPL ~\cite{CUPL} &73.3&76.9&89.3&27.5&--&92.4&89.4&65.4 \\
&&\cellcolor{gray!15}\quad+ \rpl& \cellcolor{gray!15}73.7                 & \cellcolor{gray!15}82.4                 & \cellcolor{gray!15}89.6                 & \cellcolor{gray!15}28.2                 &  \cellcolor{gray!15}--                    & \cellcolor{gray!15}92.8                 & \cellcolor{gray!15}89.4                 & \cellcolor{gray!15}65.7                                                           \\

\midrule
\multirow{18}{*}{\makecell{LAION-2B}} & \multirow{6}{*}{\makecell{ViT-B/32}} &  OpenAI templates ~\cite{clip} & 66.6 & 71.8 & 86.0 & 24.5 & 68.5 & 91.8 & 82.7 & 57.4 \\ 
&& \cellcolor{gray!15}\quad + \rpl     & \cellcolor{gray!15}66.9  & \cellcolor{gray!15}76.2                 & \cellcolor{gray!15}87.5                 & \cellcolor{gray!15}25.6                 & \cellcolor{gray!15}68.2                 & \cellcolor{gray!15}91.8 & \cellcolor{gray!15}82.7 & \cellcolor{gray!15}64.2 \\ 
&& DCLIP ~\cite{menon2022visual} & 65.7  &-- &-- &-- & 68.5                 & 90.5 & 81.2 & 53.2 \\ 
&& \cellcolor{gray!15}\quad + \rpl& \cellcolor{gray!15}66.0  &\cellcolor{gray!15}-- &\cellcolor{gray!15}-- &\cellcolor{gray!15}-- & \cellcolor{gray!15}68.2                 & \cellcolor{gray!15}90.6 & \cellcolor{gray!15}81.2 & \cellcolor{gray!15}57.7 \\ 
&& CuPL ~\cite{CUPL}    & 67.0  & 69.5                 & 86.5                 & 26.5                 &-- & 91.0 & 81.6 & 62.7                        \\
&& \cellcolor{gray!15}\quad + \rpl& \cellcolor{gray!15}67.3  & \cellcolor{gray!15}74.1                 & \cellcolor{gray!15}87.6                 & \cellcolor{gray!15}27.4                 &\cellcolor{gray!15}-- & \cellcolor{gray!15}91.1 & \cellcolor{gray!15}81.6 & \cellcolor{gray!15}63.8                       \\
\cline{2-11}
& \multirow{6}{*}{\makecell{ViT-B/16}} & OpenAI templates ~\cite{clip} & 70.2  & 71.4                 & 88.2                 & 26.9                 & 72.7                 & 91.6 & 86.5 & 57.9 \\
&&\cellcolor{gray!15}\quad + \rpl & \cellcolor{gray!15}70.3 & \cellcolor{gray!15}78.6 & \cellcolor{gray!15}88.7 & \cellcolor{gray!15}28.7 & \cellcolor{gray!15}72.6 & \cellcolor{gray!15}91.7 & \cellcolor{gray!15}86.6 & \cellcolor{gray!15}64.8 \\
&&DCLIP ~\cite{menon2022visual} & 69.5  &-- &-- &-- & 73.6                 & 91.6 & 86.0 & 58.1 \\ 

&&\cellcolor{gray!15}\quad+ \rpl& \cellcolor{gray!15}69.7  &\cellcolor{gray!15}-- &\cellcolor{gray!15}-- &\cellcolor{gray!15}-- & \cellcolor{gray!15}73.5                 & \cellcolor{gray!15}91.7 & \cellcolor{gray!15}86.0 & \cellcolor{gray!15}62.7 \\  
&&CuPL ~\cite{CUPL} & 70.6  & 70.6                 & 88.6                 & 29.6                 &-- & 91.1 & 86.2 & 63.8 \\
&& \cellcolor{gray!15}\quad+ \rpl& \cellcolor{gray!15}70.8  & \cellcolor{gray!15}76.6                 & \cellcolor{gray!15}89.4                 & \cellcolor{gray!15}30.0                 &\cellcolor{gray!15}-- & \cellcolor{gray!15}91.1 & \cellcolor{gray!15}86.2 & \cellcolor{gray!15}64.9                       \\
\cline{2-11}
& \multirow{6}{*}{\makecell{ViT-L/14}} & OpenAI templates ~\cite{clip}& 75.3  & 75.2                 & 91.9                 & 36.6                 & 78.5                 & 94.1 & 91.0 & 64.1 \\ 
&&\cellcolor{gray!15}\quad + \rpl     & \cellcolor{gray!15}75.4  & \cellcolor{gray!15}83.4                 & \cellcolor{gray!15}92.1                 & \cellcolor{gray!15}37.6                 & \cellcolor{gray!15}78.5                 & \cellcolor{gray!15}94.2 & \cellcolor{gray!15}91.0 & \cellcolor{gray!15}67.8 \\ 
&&DCLIP ~\cite{menon2022visual} & 74.5  &-- &-- &-- & 78.3                 & 93.2 & 90.8 & 63.1\\ 

&&\cellcolor{gray!15}\quad+ \rpl& \cellcolor{gray!15}74.9  &\cellcolor{gray!15}-- &\cellcolor{gray!15}-- &\cellcolor{gray!15}-- & \cellcolor{gray!15}78.2                 & \cellcolor{gray!15}93.2 & \cellcolor{gray!15}90.8 & \cellcolor{gray!15}64.4\\
&&CuPL ~\cite{CUPL} & 75.7  & 75.4                 & 92.6                & 41.2                 &-- & 94.3 & 90.6 & 68.7 \\
&&\cellcolor{gray!15}\quad+\rpl& \cellcolor{gray!15}75.9  & \cellcolor{gray!15}82.0                 & \cellcolor{gray!15}92.1                 & \cellcolor{gray!15}41.4                 &\cellcolor{gray!15}-- & \cellcolor{gray!15}94.2 & \cellcolor{gray!15}90.6 & \cellcolor{gray!15}68.8                       \\
\bottomrule
\end{tabular}
}
 \label{tab:HFL_prompting}
\end{table*}
}

\section{More Failures of Multimodal Systems}
\label{sec:sup_failure}

\begin{figure*}[h]
    \centering
    \small
    \vspace{-3mm}
    \begin{tabular}{p{8.5cm}p{8.5cm}}
    \includegraphics[width=1.0\linewidth, clip=true,trim = 0mm 0mm 0mm 0mm]{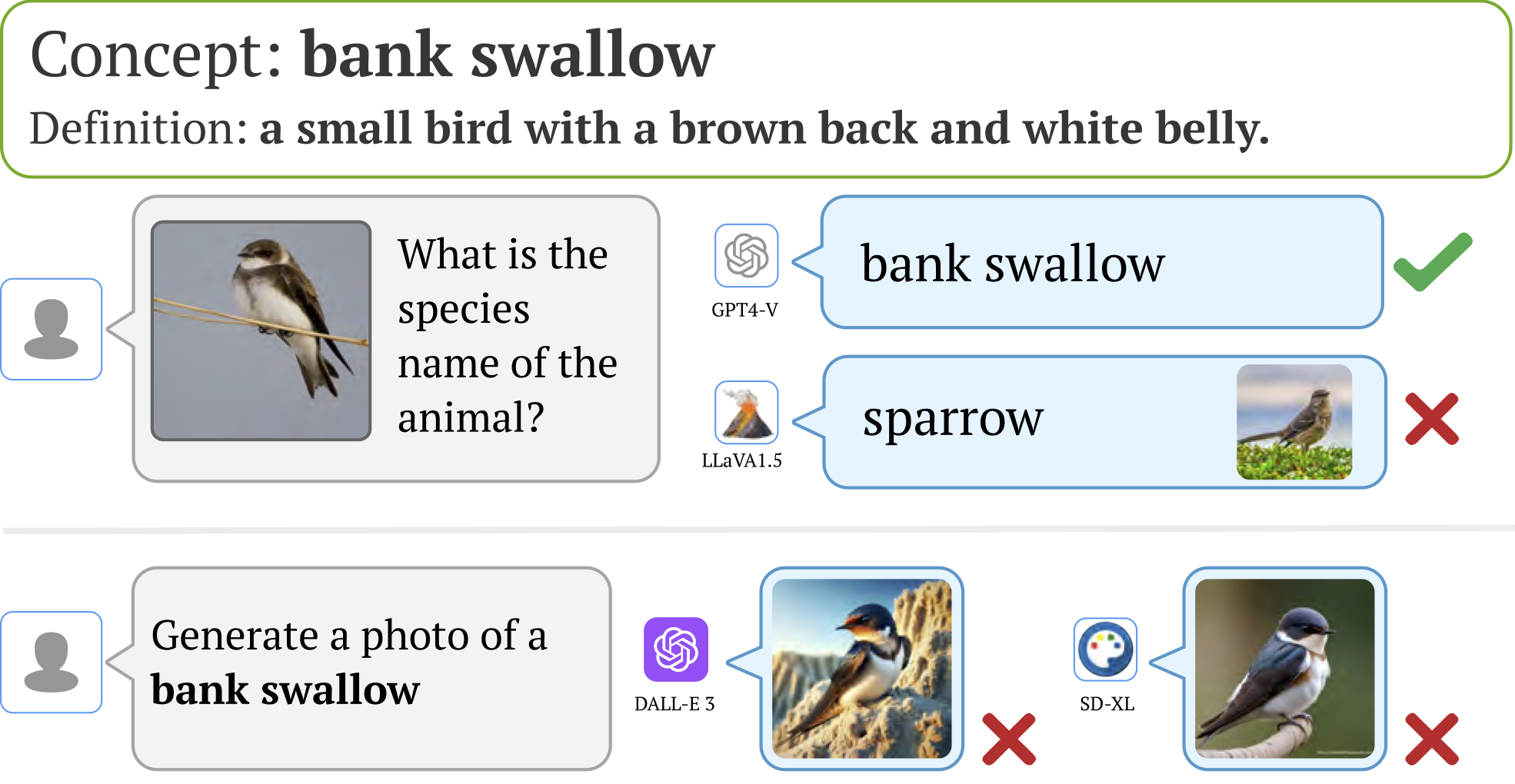}
    &
    \includegraphics[width=1.0\linewidth, clip=true,trim = 0mm 0mm 0mm 0mm]{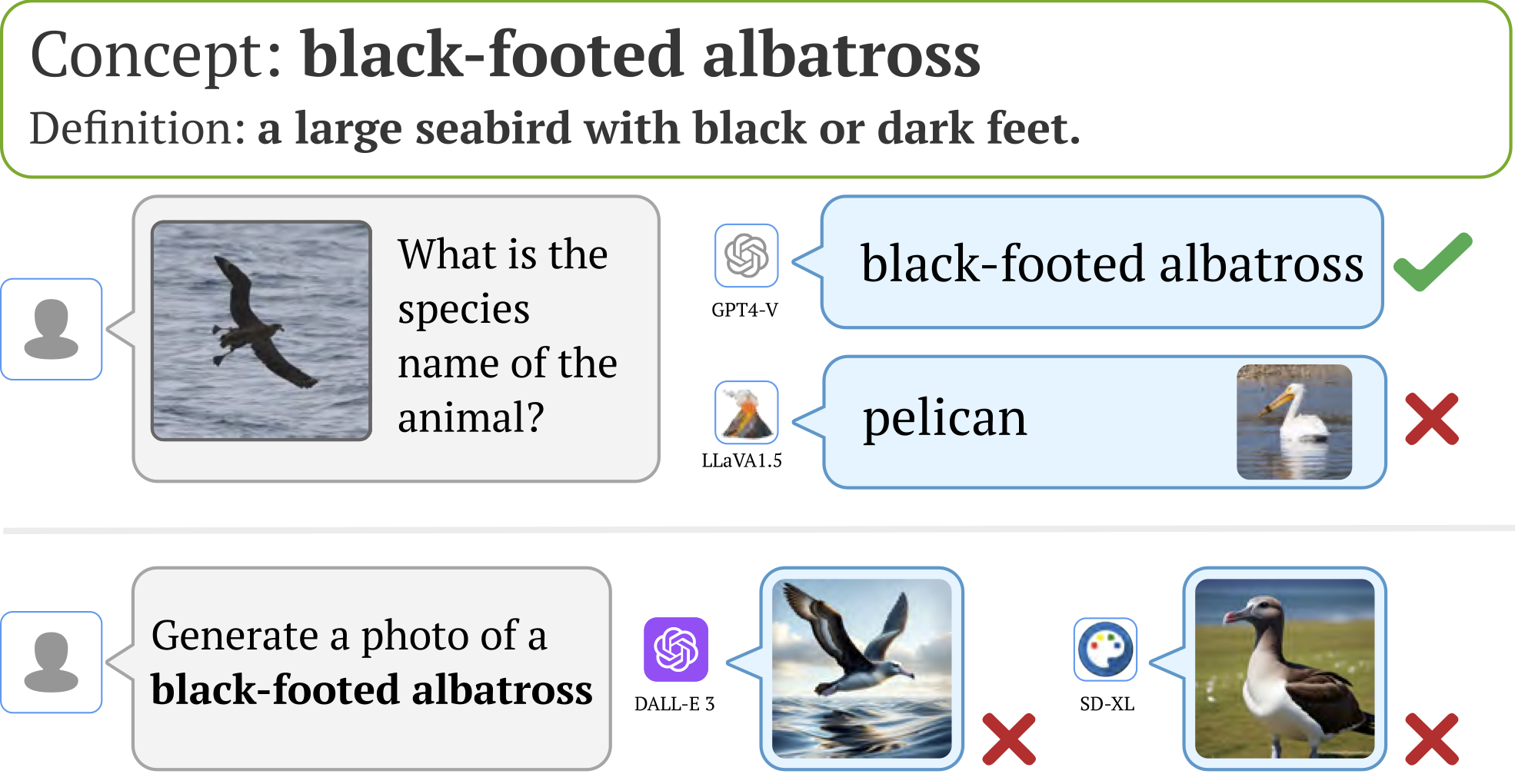} \\
    \vspace{+1mm} \\

    \includegraphics[width=1.0\linewidth, clip=true,trim = 0mm 0mm 0mm 0mm]{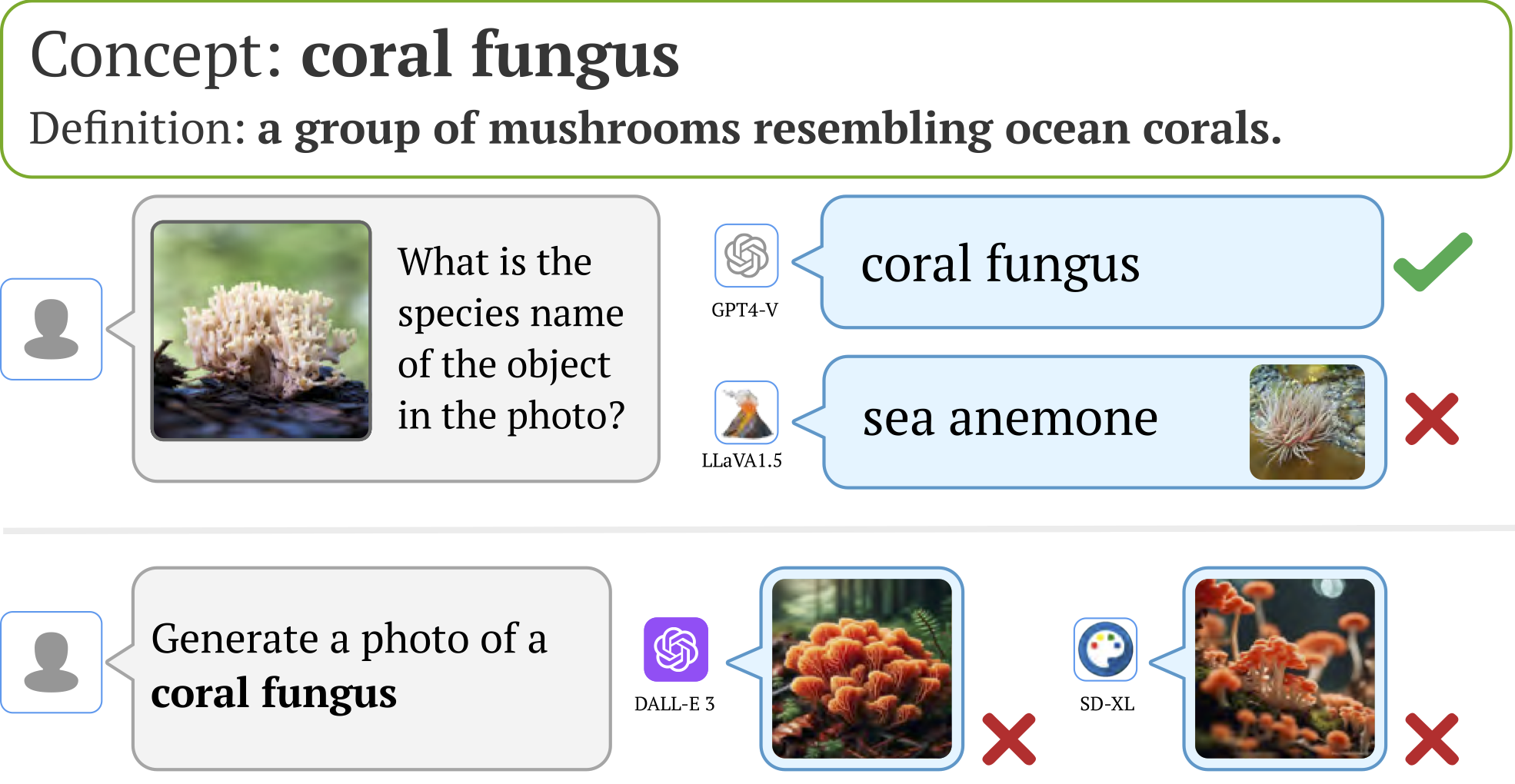}
    &
    \includegraphics[width=1.0\linewidth, clip=true,trim = 0mm 0mm 0mm 0mm]{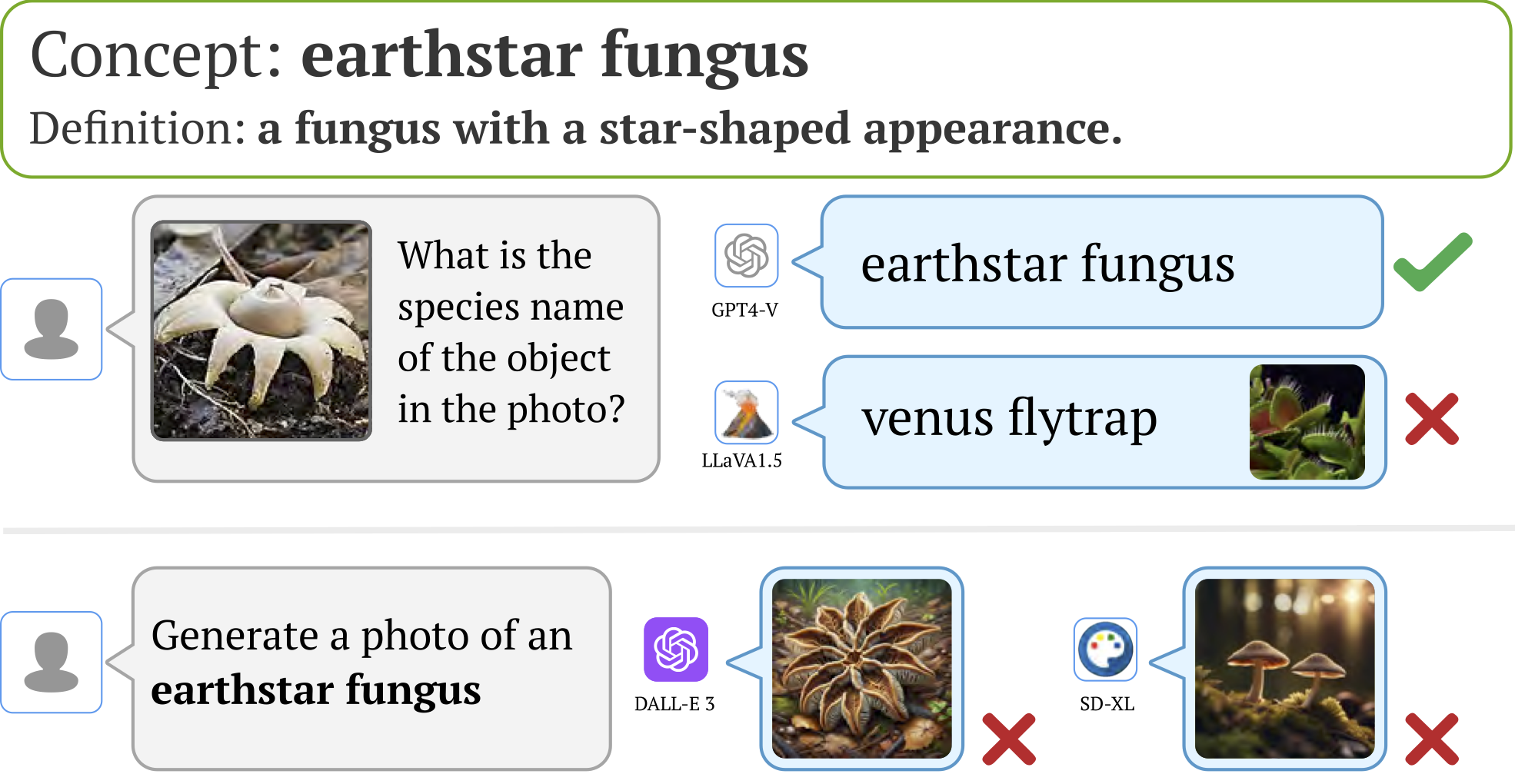} \\
    \vspace{+1mm} \\

    \includegraphics[width=1.0\linewidth, clip=true,trim = 0mm 0mm 0mm 0mm]{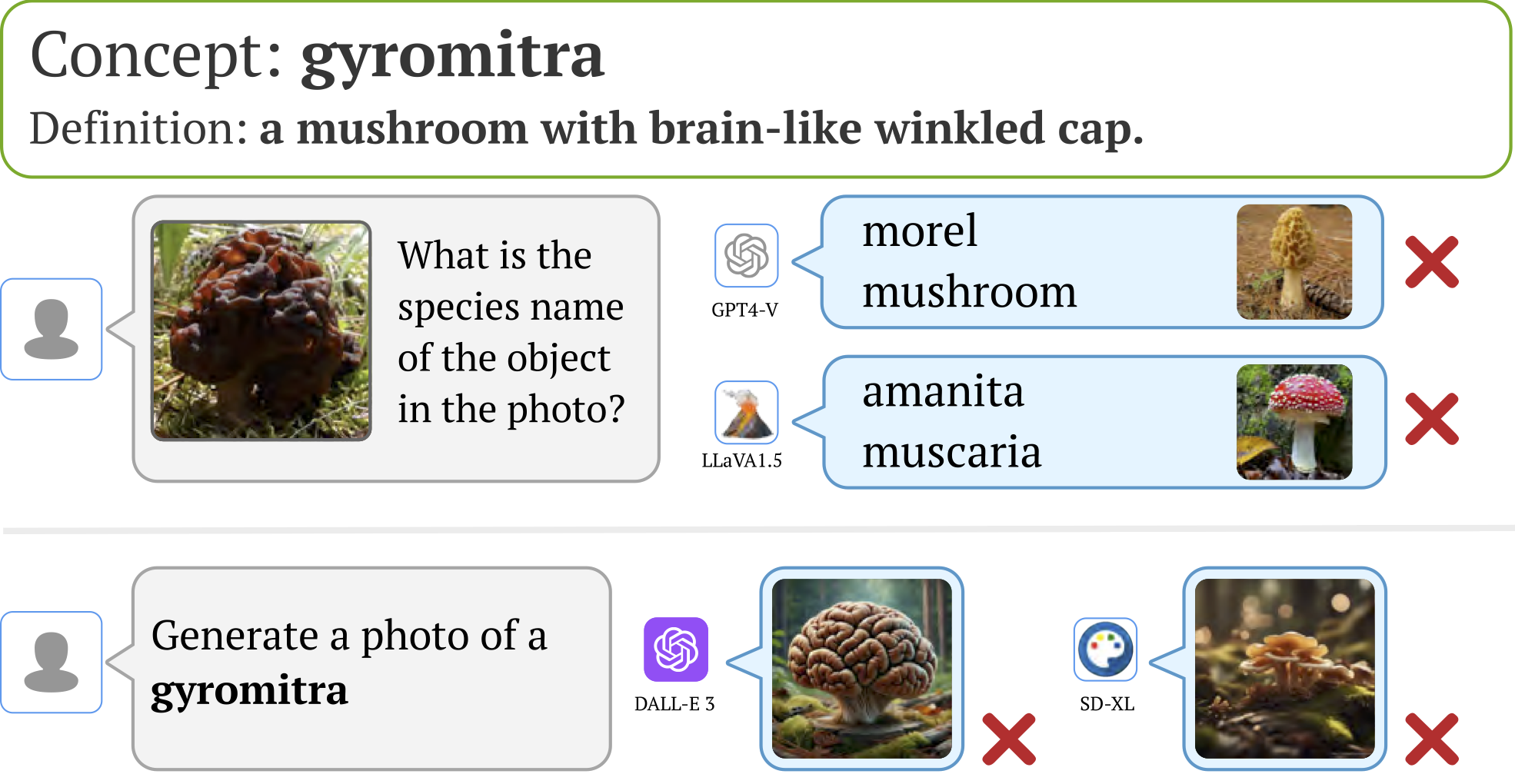}
    &
    \includegraphics[width=1.0\linewidth, clip=true,trim = 0mm 0mm 0mm 0mm]{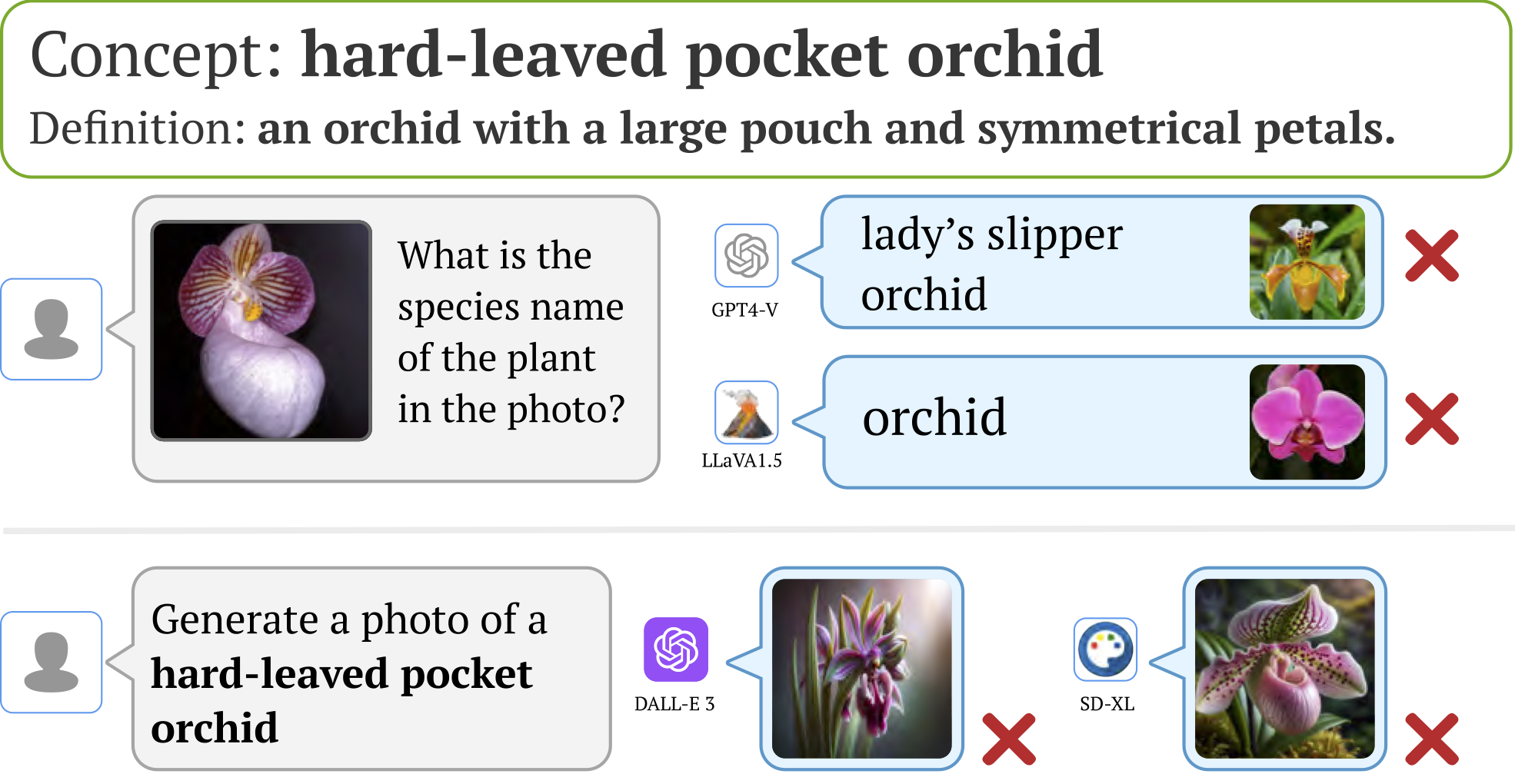} \\ 
    \vspace{+1mm} \\ 

    \includegraphics[width=1.0\linewidth, clip=true,trim = 0mm 0mm 0mm 0mm]{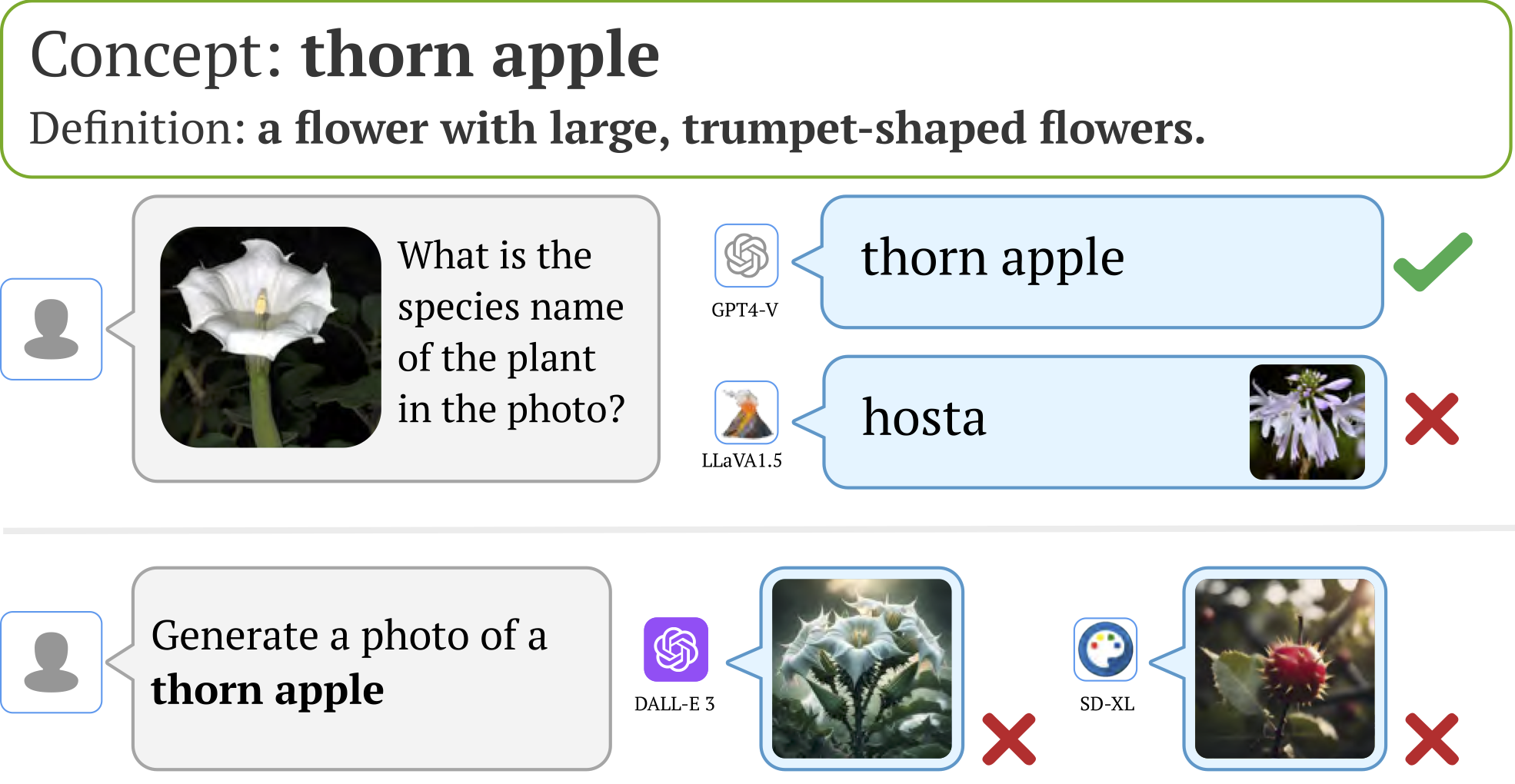}
    &
    \includegraphics[width=1.0\linewidth, clip=true,trim = 0mm 0mm 0mm 0mm]{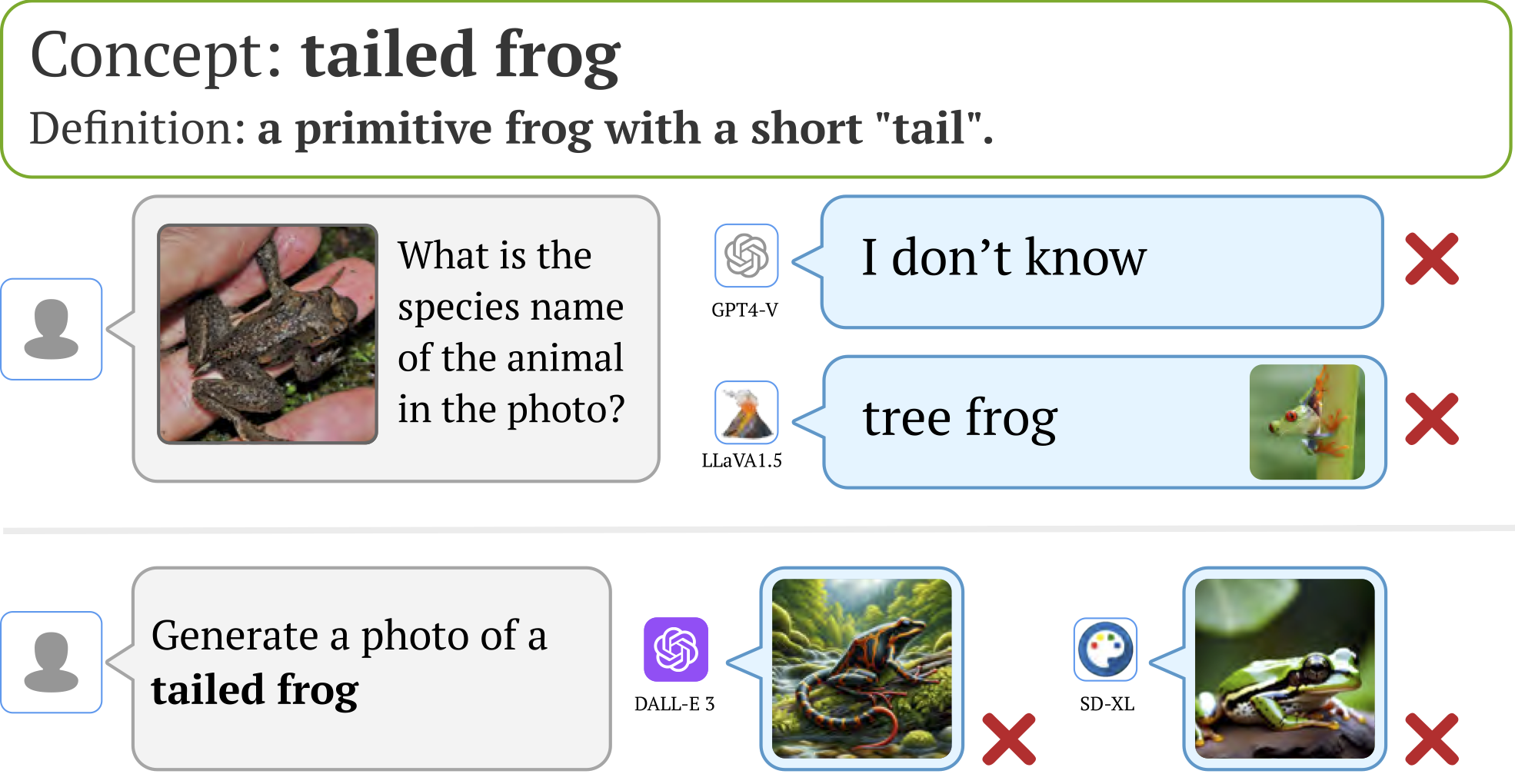} \\
    
    \end{tabular}
    \caption{\textbf{State-of-the-art multimodal systems fail to recognize or generate tailed concepts (part 1).} We show more failure cases of popular multimodal systems (GPT-4V~\cite{gpt4v}, LLaVA1.5~\cite{liu2023llava}, DALL-E 3~\cite{dalle3}, and Stable Diffusion XL~\cite{stable_diffusion}) on tailed concepts sampled from standard benchmark datasets such as ImageNet~\cite{deng2009imagenet}, Flowers~\cite{flowers}, Aircraft~\cite{aircraft}, and etc. For GPT-4V and LLaVA1.5, we include example images of incorrectly predicted classes to show that visual chatbots often misclassify rare concepts as some similar-looking yet more common concepts. We include a definition for each tailed concept to show that DALL-E 3 and Stable Diffusion (SD-XL) can fail to capture the correct colors, shapes, and other characteristics of these concepts.
    }
\vspace{-3mm}
\label{fig:sup_tail_concepts_failure_part1}
\end{figure*}

\begin{figure*}[h]
    \centering
    \small
    \begin{tabular}{p{8.5cm}p{8.5cm}}

    \includegraphics[width=1.0\linewidth, clip=true,trim = 0mm 0mm 0mm 0mm]{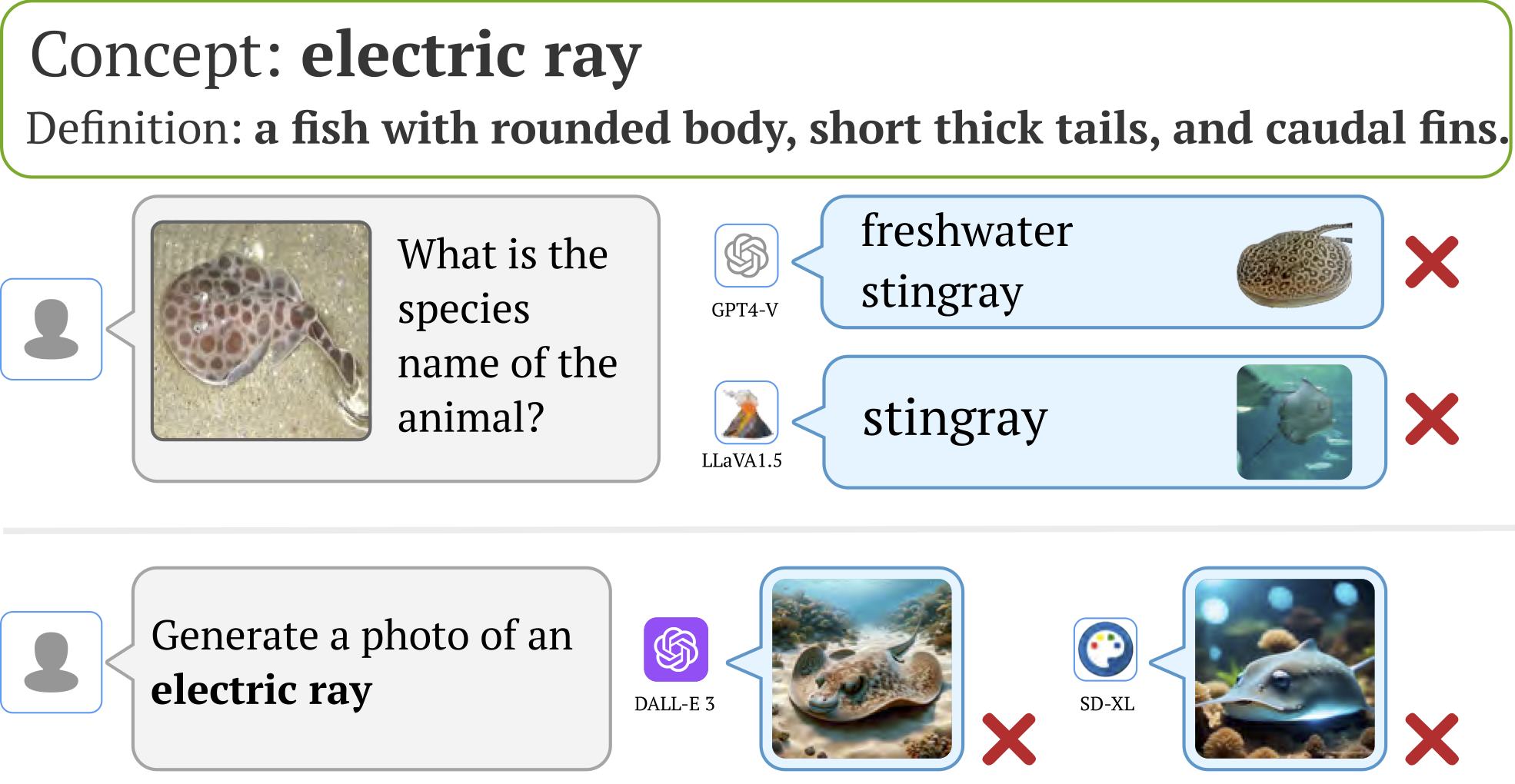}
    &
    \includegraphics[width=1.0\linewidth, clip=true,trim = 0mm 0mm 0mm 0mm]{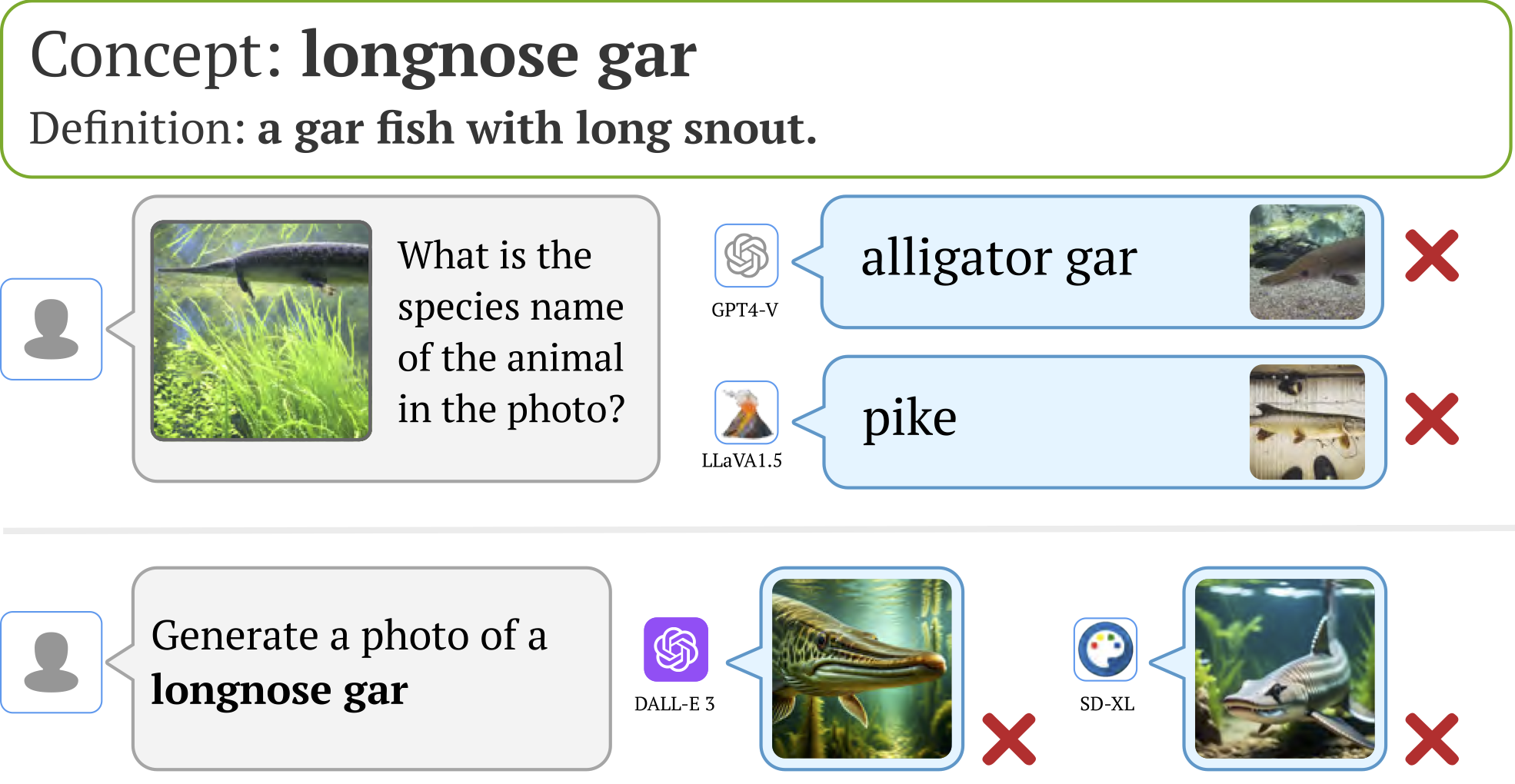} \\
    \vspace{+1mm} \\

    \includegraphics[width=1.0\linewidth, clip=true,trim = 0mm 0mm 0mm 0mm]{figures/failure/night_snake_v2.png}
    &
    \includegraphics[width=1.0\linewidth, clip=true,trim = 0mm 0mm 0mm 0mm]{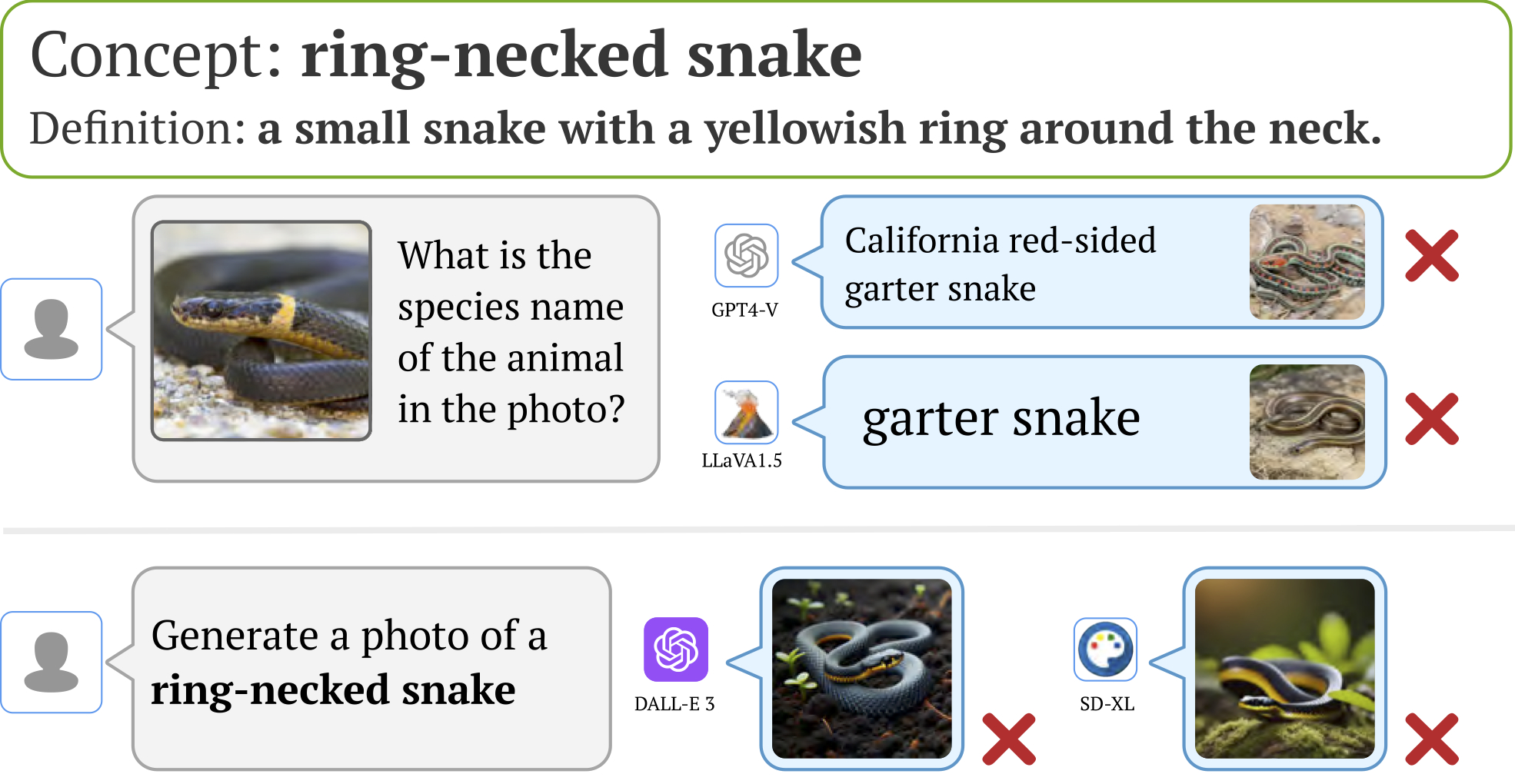} \\
    \vspace{+1mm} \\

    \includegraphics[width=1.0\linewidth, clip=true,trim = 0mm 0mm 0mm 0mm]{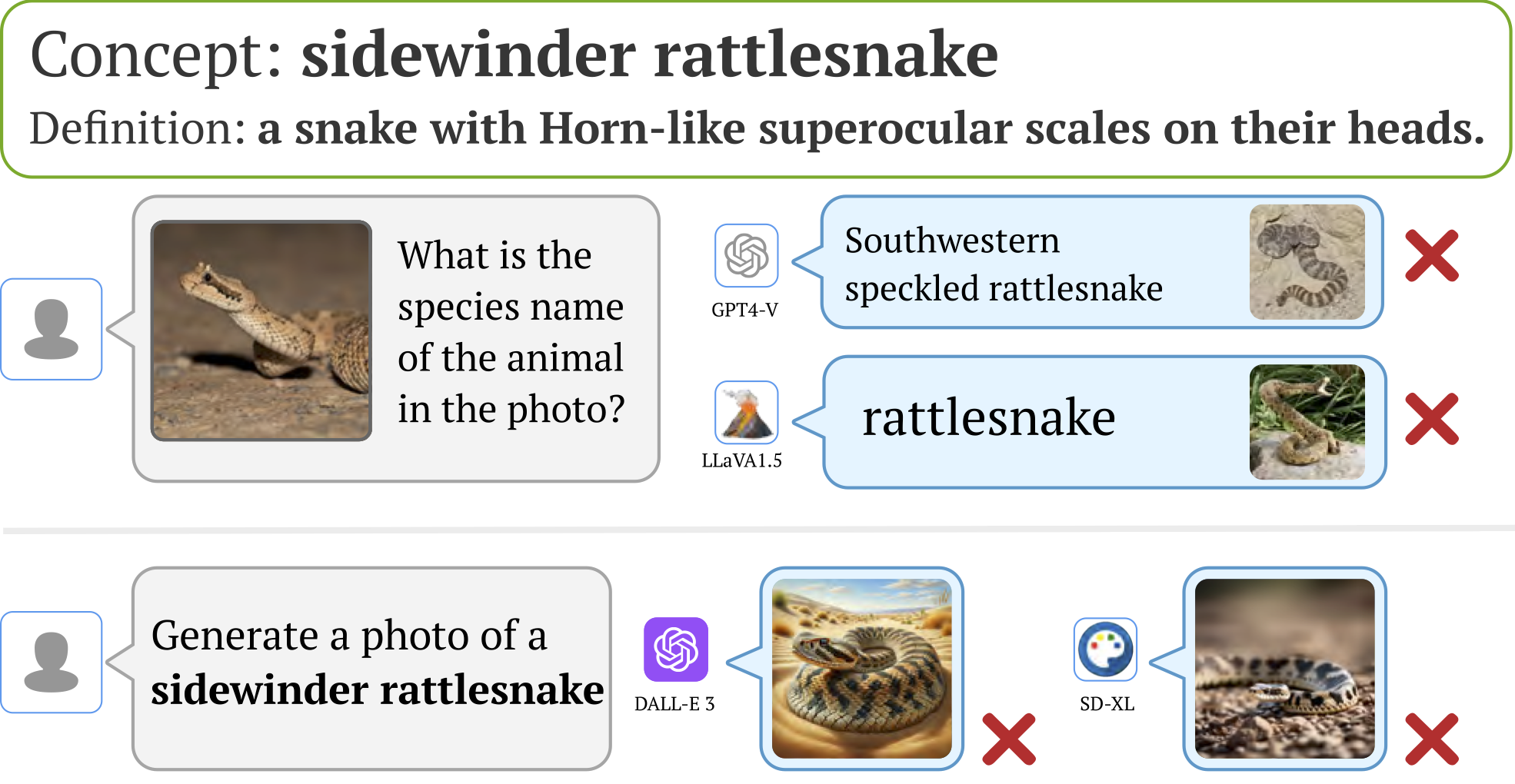}
    &
    \includegraphics[width=1.0\linewidth, clip=true,trim = 0mm 0mm 0mm 0mm]{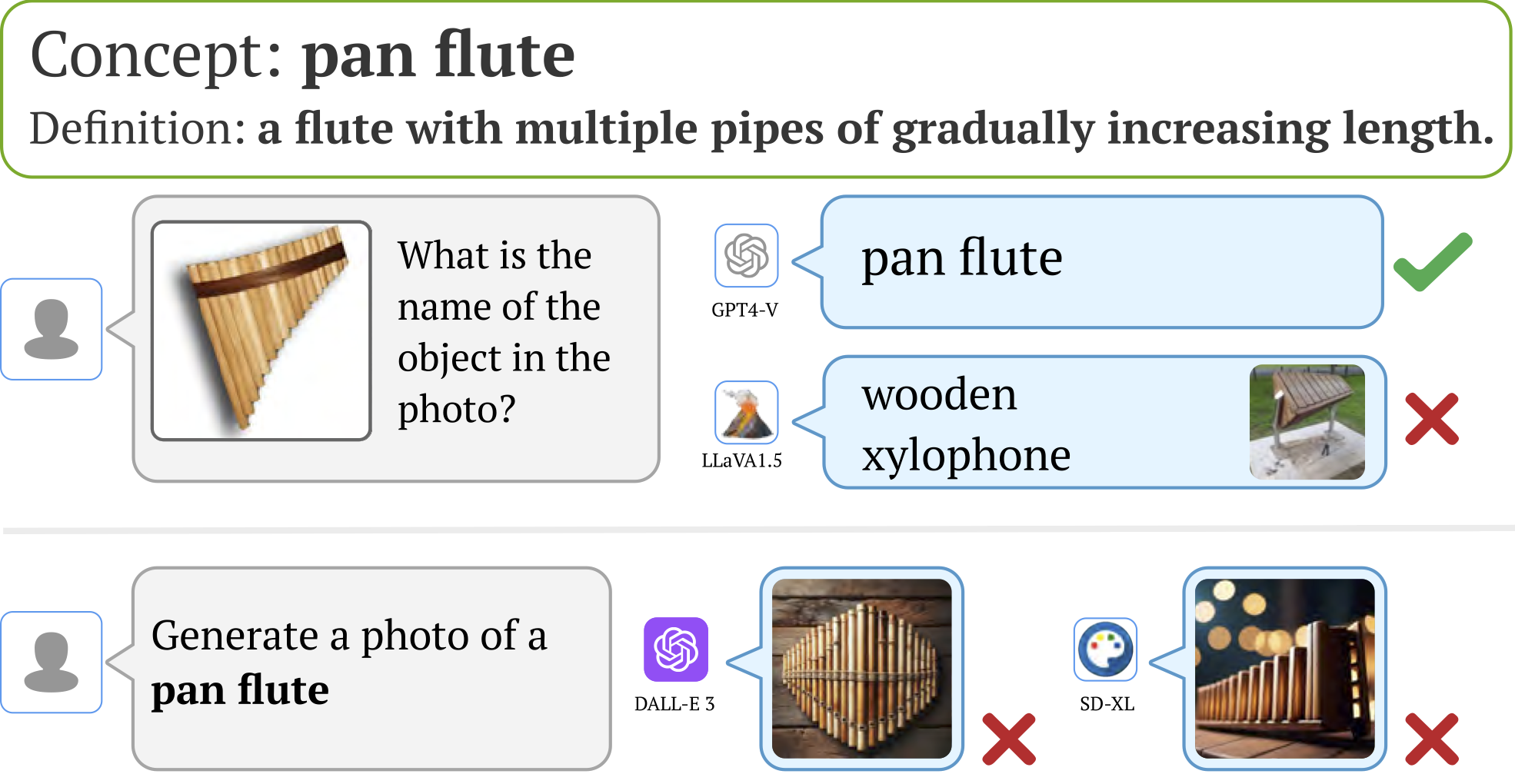} \\
    \vspace{+1mm}\\

    \includegraphics[width=1.0\linewidth, clip=true,trim = 0mm 0mm 0mm 0mm]{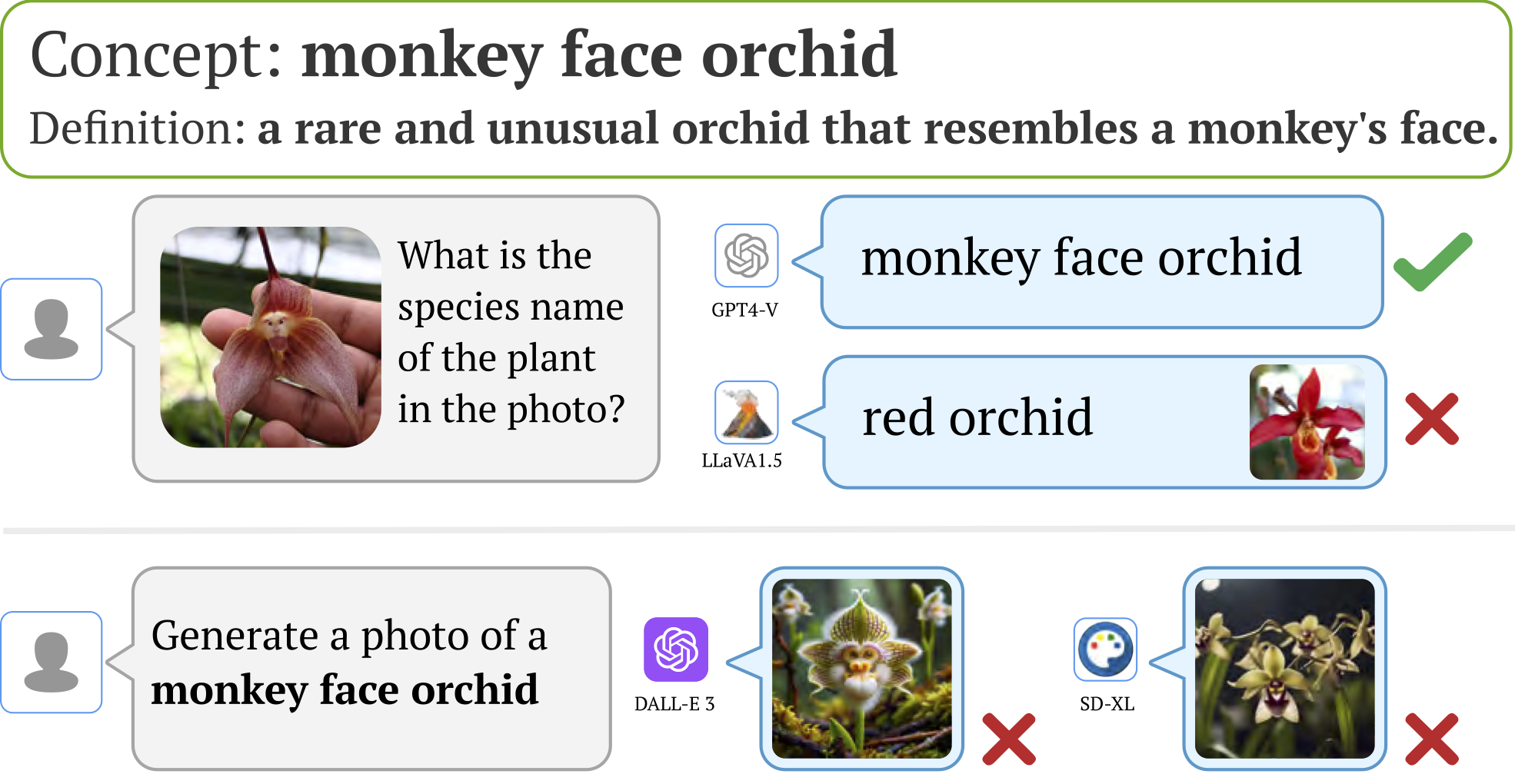}
    &
    \includegraphics[width=1.0\linewidth, clip=true,trim = 0mm 0mm 0mm 0mm]{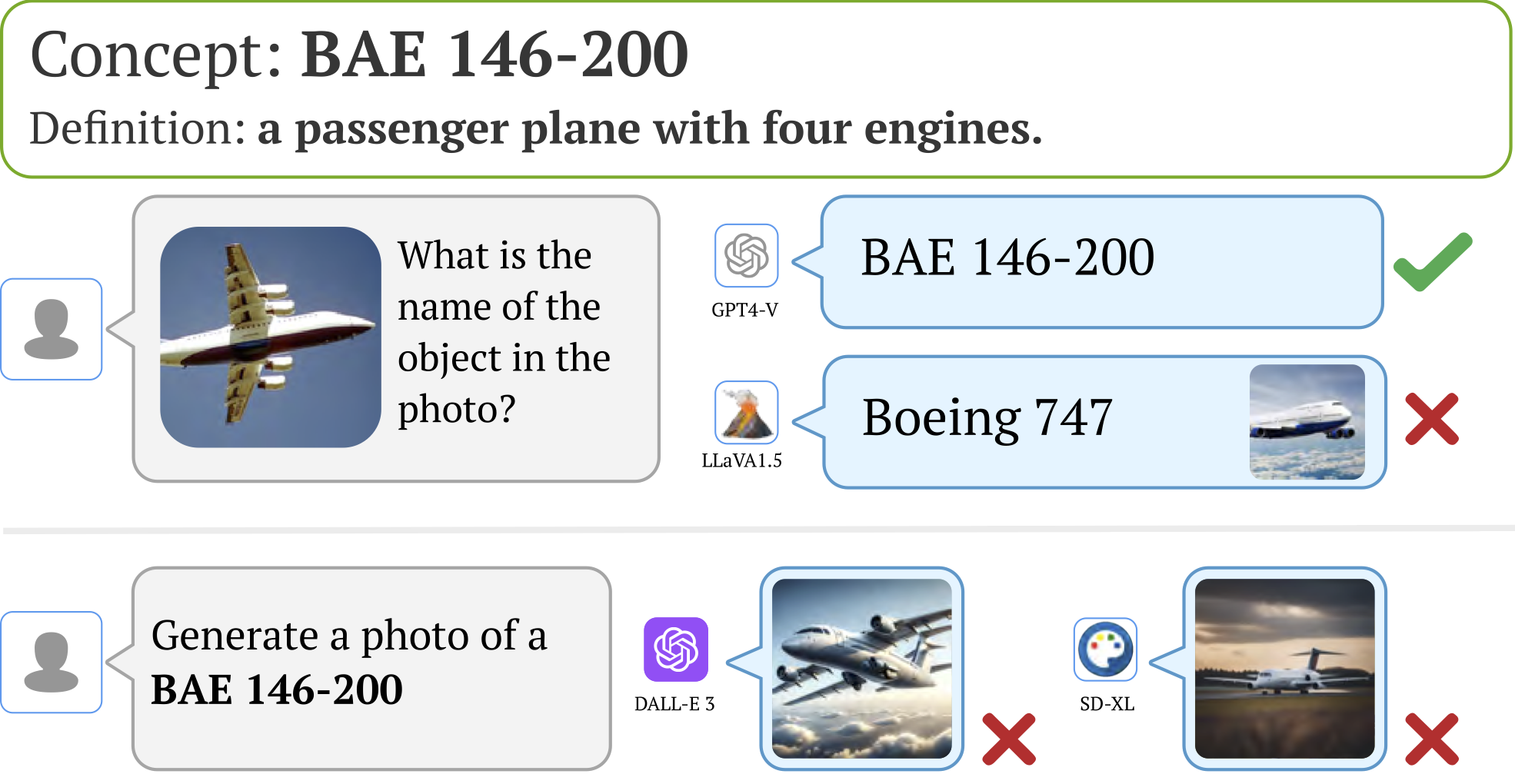} \\
    
    \end{tabular}
    \caption{\textbf{State-of-the-art multimodal systems fail to recognize or generate tailed concepts (part 2).}
    We show more failure cases of popular multimodal systems (GPT-4V~\cite{gpt4v}, LLaVA1.5~\cite{liu2023llava}, DALL-E 3~\cite{dalle3}, and Stable Diffusion XL~\cite{stable_diffusion}) on tailed concepts sampled from standard benchmark datasets such as ImageNet~\cite{deng2009imagenet}, Flowers~\cite{flowers}, Aircraft~\cite{aircraft}, and etc. For GPT-4V and LLaVA1.5, we include example images of incorrectly predicted classes to show that visual chatbots often misclassify rare concepts as some similar-looking yet more common concepts. We include a definition for each tailed concept to show that DALL-E 3 and Stable Diffusion (SD-XL) can fail to capture the correct colors, shapes, and other characteristics of these concepts.
    }
\vspace{-3mm}
\label{fig:sup_tail_concepts_failure_part2}
\end{figure*}

In Figure~\ref{fig:sup_tail_concepts_failure_part1} and~\ref{fig:sup_tail_concepts_failure_part2}, we show more failure cases of state-of-the-art multimodal systems on tailed concepts identified by our frequency estimation method. These tailed concepts are randomly sampled from nine benchmark datasets and span across a variety of domains, including birds, flowers, fungi, snakes, frogs, fish, household items, and more. We qualitatively test the visual recognition abilities of two most popular visual chatbots: GPT-4V~\cite{gpt4v} (trained on proprietary data) and LLaVA1.5~\cite{liu2023llava} (trained on open-source data using a frozen CLIP image encoder). We also test the image generation abilities of two most popular generative models: DALL-E 3~\cite{dalle3} (trained on proprietary data) and Stable Diffusion XL~\cite{stable_diffusion} (trained on open-source data using a frozen CLIP text encoder). We observe that these systems fail to recognize or generate more than half of the tailed concepts we sampled. In particular, LLaVA1.5 and Stable Diffusion XL fail on all these tailed concepts, suggesting a large performance gap between proprietary and open-source multimodal systems.

\section{REAL-Prompt for Generative Models}
\label{sec:real_prompt_generative}
\begin{figure*}[ht]
\centering
\small
    \vspace{-3mm}
    \begin{tabular}{p{17cm}}

    \includegraphics[width=1.0\linewidth, clip=true,trim = 0mm 0mm 4.5mm 0mm]{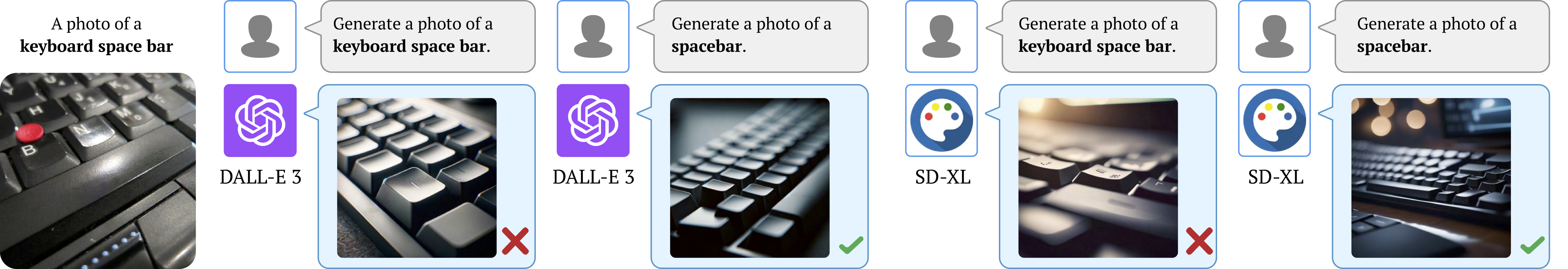}    \\
    {\footnotesize {\bf (a)} {\tt keyboard space bar}, a {\bf long bar} at the {\bf bottom} of a computer keyboard. When prompted with the original concept name ({\tt keyboard space bar}), both DALL-E 3 and SD-XL fail by focusing on generating images of the keyboard. However, when prompted with the most frequent synonyms ({\tt space bar}), both are able to generate correct images.} \\
    \vspace{+1mm}    

    \includegraphics[width=1.0\linewidth, clip=true,trim = 0mm 0mm 4.5mm 0mm]{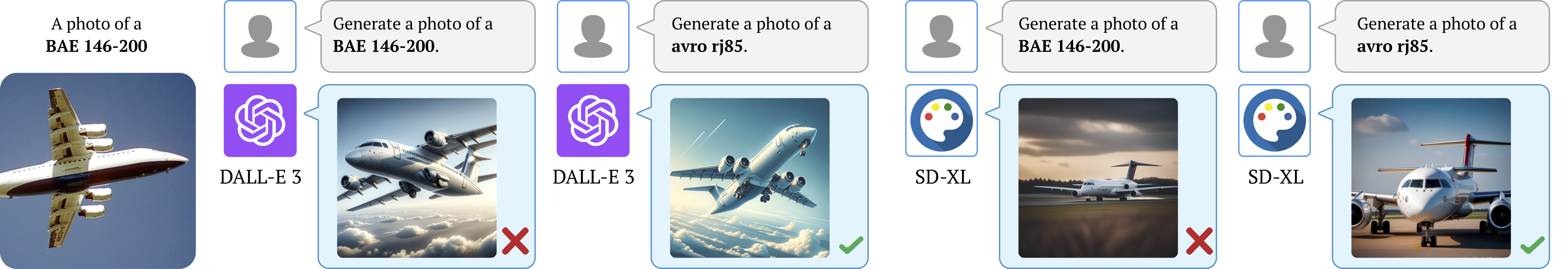}    \\
    {\footnotesize {\bf (b)} {\tt BAE 146-200}, an airplane with {\bf 4 engines}. When prompted with the original concept name ({\tt BAE 146-200}), both DALL-E 3 and SD-XL fail by generating only 2 engines. However, when prompted with the most frequent synonym ({\tt avro rj85}), both are able to generate correct images with 4 engines.} \\
    \vspace{+1mm}        
    
    \includegraphics[width=1.0\linewidth, clip=true,trim = 0mm 0mm 4.5mm 0mm]{figures/dalle3_sd/bank_swallow.png}\\
    \vspace{-3mm}
    {\footnotesize {\bf (c)} {\tt bank swallow}, a small bird with {\bf brown back} and {\bf white belly}. When prompted with the original concept name ({\tt bank swallow}), both DALL-E 3 and SD-XL generate incorrect images of birds with incorrect black backs. However, prompting with the most frequent synonym ({\tt sand martin}) guides both systems to produce correct images.}\\
    \vspace{+1mm}

    \includegraphics[width=1.0\linewidth, clip=true,trim = 0mm 0mm 4.5mm 0mm]{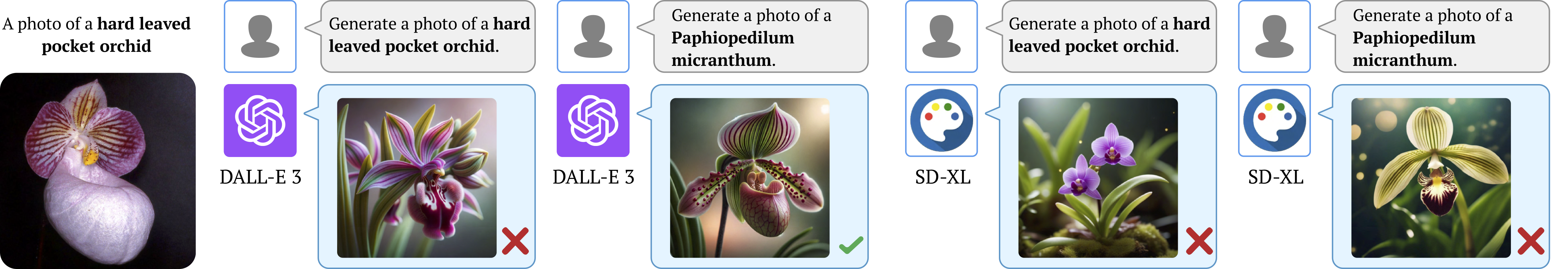}\\
    \vspace{-3mm}
    {\footnotesize {\bf (d)} {\tt hard leaved pocket orchid}, a type of orchid with a {\bf distinctive pouch} and {\bf symmetrical large petals}. When prompted with the original concept name ({\tt hard leaved pocket orchid}), both DALL-E 3 and SD-XL generate incorrect images (note the missing pocket and shape of the petals). However, when prompted with the most frequent synonym ({\tt Paphiopedilum micranthum}), DALL-E 3 produces the correct image. In contrast, SD-XL is able to recover the shape of petals but still misses the pocket.}\\
    
    \end{tabular}
\caption{\textbf{Prompting with the most frequent synonym can help DALL-E 3 and Stable Diffusion generate correct images (part 1).} We show more examples when DALL-E 3~\cite{dalle3} and Stable Diffusion XL (SD-XL)~\cite{stable_diffusion} initially fail to generate correct images for tailed concepts when prompted with their original concept names in standard classification benchmark datasets. We sample diverse tail concepts covering a variety of domains including household items, birds, flowers, insects, reptiles, and etc. We show that REAL-Prompt (prompting with the most frequent synonyms) often helps DALL-E 3 and Stable Diffusion produce correct images. We notice that for the {\tt hard leaved pocket orchid} and {\tt ring-necked snake}, the generated images of SD-XL improve but are still inaccurate. This suggests future work to improve open-source generative models on rare concepts. }
\label{fig:sup_improve-dalle3_part1}
\end{figure*}

\begin{figure*}[ht]
    \centering
    \small
    \begin{tabular}{p{17cm}}

    \includegraphics[width=1.0\linewidth, clip=true,trim = 0mm 0mm 4.5mm 0mm]{figures/dalle3_sd/datura.png}    \\
    \vspace{-3mm}
    {\footnotesize {\bf (e)} {\tt thorn apple}, a plant with {\bf large, white, trumpet-shaped} flowers. When prompted with the original concept name ({\tt thorn apple}), DALL-E 3 generates an image with sharp thorns along its stem. Even worse, SD-XL takes the name superficially and generates an apple with thorns. On the contrary, prompting with the most frequent synonym ({\tt datura}) leads to correct images in both systems.} \\
    \vspace{+1mm}

    \includegraphics[width=1.0\linewidth, clip=true,trim = 0mm 0mm 4.5mm 0mm]{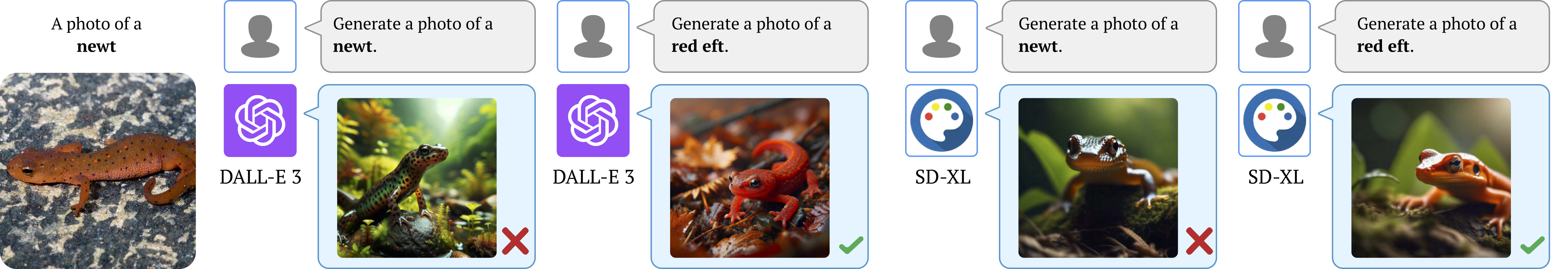} \\
    \vspace{-3mm}
    {\footnotesize {\bf (f)} {\tt newt}, a type of salamander known for its {\bf bright orange to red color} and {\bf scattered darker spots}. When prompted with the original concept name ({\tt newt}), both DALL-E 3 and SD-XL fail by generating a green-colored skin. However, prompting with the most frequent synonym ({\tt red eft}) leads to the correct red-colored body.} \\
    \vspace{+1mm}

    \includegraphics[width=1.0\linewidth, clip=true,trim = 0mm 0mm 4.5mm 0mm]{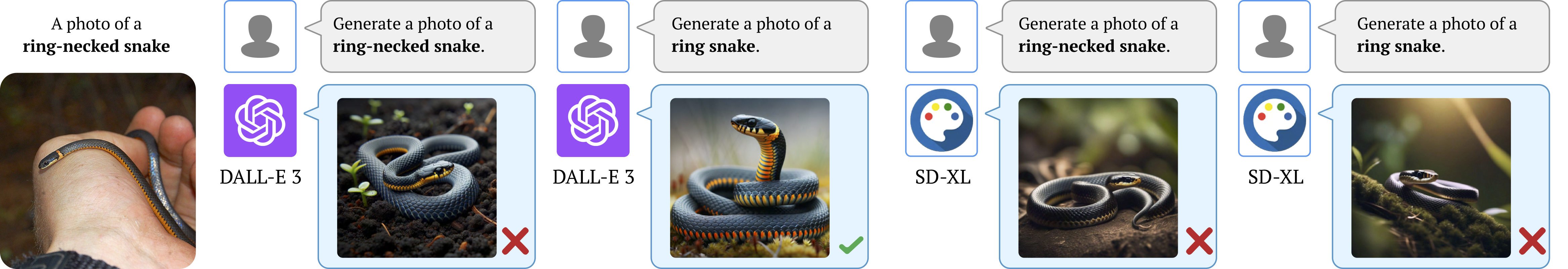} \\
    \vspace{-3mm}
    {\footnotesize {\bf (g)} {\tt ring-necked snake}, a small snake with a {\bf yellowish ring around the neck}. When prompted with the original concept name ({\tt ring-necked snake}), both DALL-E 3 and SD-XL fail by missing the yellow ring around the snake's neck. However, prompting with the most frequent synonym ({\tt ring snake}) helps DALL-E 3 recover the ring. Meanwhile, SD-XL still fails to capture the ring which is likely due to insufficient relevant images in its pretraining data.} \\
    
    \end{tabular}
    \caption{\textbf{Prompting with the most frequent synonym can help DALL-E 3 and Stable Diffusion (SD-XL) generate correct images (part 2).}
    We show more examples when DALL-E 3~\cite{dalle3} and Stable Diffusion XL (SD-XL)~\cite{stable_diffusion} initially fail to generate correct images for tailed concepts when prompted with their original concept names in standard classification benchmark datasets. We sample diverse tail concepts covering a variety of domains including household items, birds, flowers, insects, reptiles, etc. We show that REAL-Prompt (prompting with the most frequent synonyms) often helps DALL-E 3 and Stable Diffusion produce correct images. We notice that for the {\tt hard-leaved pocket orchid} and {\tt ring-necked snake}, the generative images of SD-XL improve but are still inaccurate. This suggests future work to improve open-source generative models on rare concepts.
    }
\label{fig:sup_improve-dalle3_part2}
\end{figure*}

Figure~\ref{fig:sup_improve-dalle3_part1} and~\ref{fig:sup_improve-dalle3_part2} contains qualitative results of REAL-Prompt on state-of-the-art text-to-image generative models including DALL-E 3~\cite{dalle3} and Stable Diffusion XL~\cite{stable_diffusion}. 
This shows that using most frequent synonyms can  help generate correct images for tailed concepts. We also note that our method is more effective on the more capable generative model DALL-E 3, presumably because it is trained with more data than open-source Stable Diffusion XL. This suggests opportunities for future work to improve open-source VLMs on image synthesis for rare concepts.

{
\setlength{\tabcolsep}{0.95em}
\begin{table*}[]
\centering
\small
\caption{\textbf{The importance of synonym-filtering for REAL-Prompt.} After obtaining synonyms from ChatGPT, we use OpenCLIP's text encoder to filter the synonyms that might be confused with other downstream concepts. We show that this filtering step is critical for REAL-Prompt's performance.}
\vspace{-3mm}
\label{tab:t2t-filter}
\begin{tabular}{lcccccccc}
\toprule
                 & ImageNet & Flowers & Cars & Aircraft & CUB  & Pets & Food & DTD  \\
\midrule
REAL-Prompt w/o Synonym Filtering  & 50.5     & 45.0    & 59.3 & 9.5      & 55.6 & 39.9 & 63.5 & 10.9 \\
REAL-Prompt w/ Synonym Filtering & 63.6     & 76.6    & 82.7 & 18.0     & 64.0 & 88.8 & 81.0 & 59.9 \\ 
\bottomrule
\end{tabular}
\vspace{-3mm}
\end{table*}
}

\end{document}